%% file: main.tex
\documentclass[final]{cvpr}

\usepackage[pagebackref,breaklinks,colorlinks]{hyperref}
\usepackage{enumitem}

\usepackage{adjustbox}
\usepackage{listings}
\usepackage{times}
\usepackage{epsfig}
\usepackage{graphicx}
\usepackage{amsmath}
\usepackage{amssymb}
\usepackage{wrapfig}
\usepackage{caption}
\usepackage{subcaption}
\usepackage{float}
\usepackage{array}
\usepackage{color}
\usepackage{multirow}
\usepackage{diagbox}
\usepackage{booktabs}
\usepackage{color}
\usepackage{graphbox}
\usepackage[normalem]{ulem}
\usepackage{multirow}
\usepackage{varwidth}
\usepackage{pifont} %

\definecolor{codegreen}{rgb}{0,0.4,0}
\definecolor{codegray}{rgb}{1.0,0.5,0.5}
\definecolor{codepurple}{rgb}{0.58,0,0}
\definecolor{tealblue}{rgb}{0,0.5,0.5}
\definecolor{codebackcolour}{rgb}{0.95,0.95,0.92}
\definecolor{darkgreen}{RGB}{0,127,0}
\definecolor{darkred}{RGB}{200,0,0}
\definecolor{orange}{rgb}{1,0.5,0}
\def\greencheckmark{\textcolor{darkgreen}{\checkmark}}
\def\redxmark{\textcolor{darkred}{\ding{55}}}  %

\lstdefinestyle{mystyle}{
    backgroundcolor=\color{codebackcolour},
    commentstyle=\bf\color{codegreen},
    keywordstyle=\color{magenta},
    stringstyle=\color{codepurple},
    basicstyle=\ttfamily\tiny,
    breakatwhitespace=false,
    breaklines=true,
    captionpos=b,
    keepspaces=true,
    numbers=none,
    showspaces=false,
    showstringspaces=false,
    showtabs=false,
    tabsize=2
}

\definecolor{grey50}{rgb}{0.5,0.5,0.5}

\newcommand{\zdensity }{\mathcal{Z}^\mathcal{T}}
\newcommand{\zcolor }{\mathcal{Z}^\mathcal{C}}
\newcommand{\fdensity }{F^\mathcal{T}}
\newcommand{\fcolor }{F^\mathcal{C}}
\newcommand{\R}{\mathbb{R}}  %

\newcommand{\bfr}{\ensuremath{\textbf{r}}}

\newcommand\blfootnote[1]{%
  \begingroup
  \renewcommand\thefootnote{}\footnote{#1}%
  \addtocounter{footnote}{-1}%
  \endgroup
}

\def\cvprPaperID{4839} %
\def\confYear{CVPR 2022}

\title{RTMV: A Ray-Traced Multi-View Synthetic Dataset for Novel View Synthesis}
\author{
  {\large Jonathan Tremblay$^{*1}$ \quad Moustafa Meshry$^{*2}$ \quad Alex Evans$^1$ \quad Jan Kautz$^1$ \quad Alexander Keller$^1$} \\
  { Sameh Khamis$^1$ \quad Thomas M\"{u}ller$^1$ \quad Charles Loop$^1$ \quad Nathan Morrical$^1$ \quad Koki Nagano$^1$ } \\
  { Towaki Takikawa$^1$ \quad Stan Birchfield$^1$} \\\\
  $^1$NVIDIA \qquad $^2$University of Maryland, College Park
}

\begin{document}

\twocolumn[{%
\renewcommand\twocolumn[1][]{#1}%
\maketitle
\vspace{-4mm}

\begin{center}
    \centering
    \captionsetup{type=figure}
    \includegraphics[height=7.6cm]{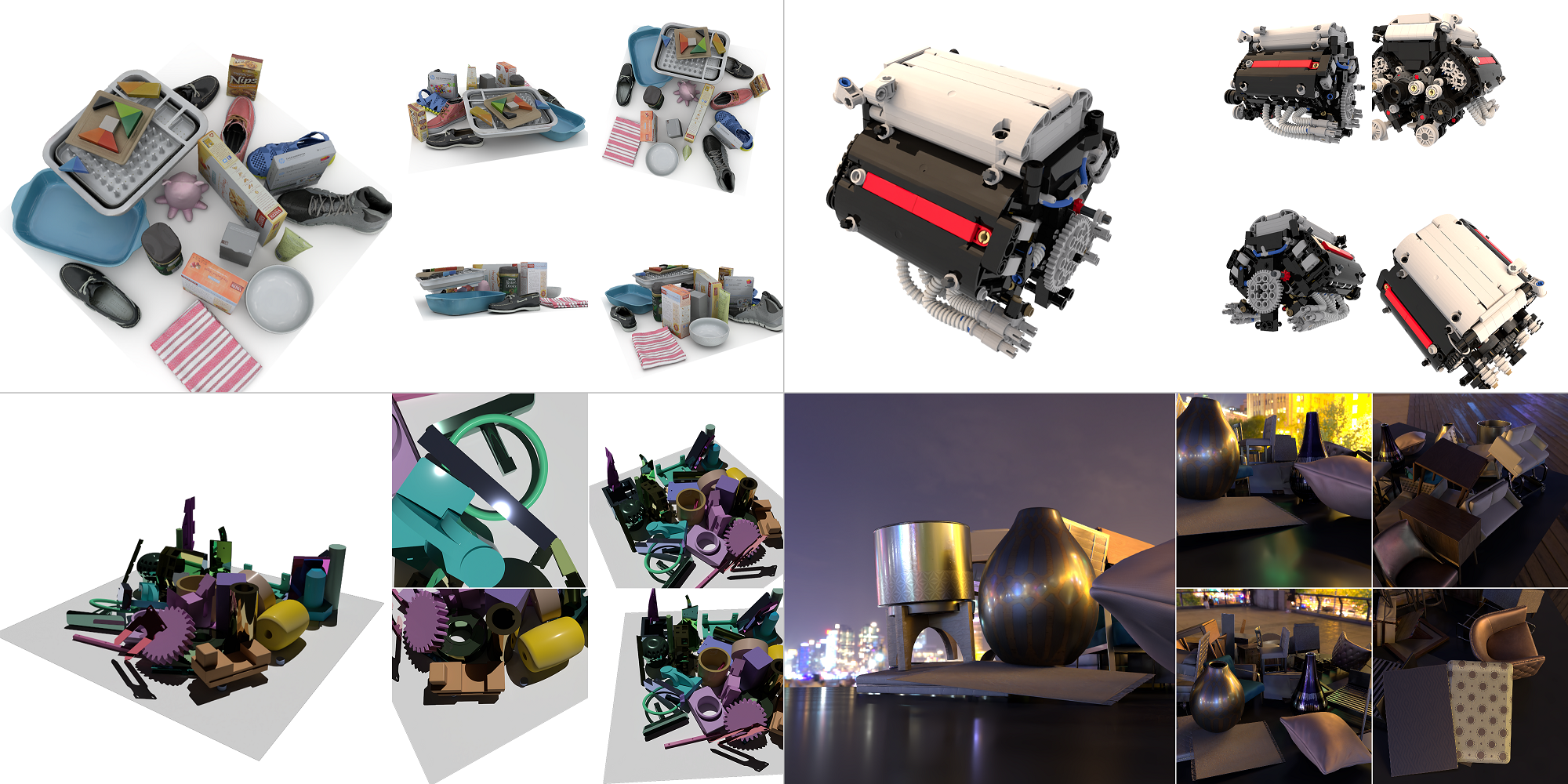}
    \captionof{figure}{We present RTMV, a large-scale high-fidelity ray-traced synthetic dataset for novel view synthesis.
    RTMV is composed of nearly 2000 scenes from 4 different environments exhibiting large varieties in view positions, lighting, object shapes, materials, and textures.
    Each quadrant shows a single scene from each environment, captured from multiple viewpoints.
    \label{fig:header}
    }
    \vspace{-0.2cm}
\end{center}%
}]

\maketitle

\blfootnote{$^*$Equal contribution.  Second author was an NVIDIA intern.  }
\begin{abstract}

\vspace{-3mm}
We present a large-scale synthetic dataset for novel view synthesis consisting of $\sim$300k images rendered from nearly 2000 complex scenes using high-quality ray tracing at high resolution ($1600\times 1600$ pixels).
The dataset is orders of magnitude larger than existing synthetic datasets for novel view synthesis,
thus providing a large unified benchmark for both training and evaluation.
Using 4 distinct sources of high-quality 3D meshes, the scenes of our dataset exhibit challenging variations in camera views, lighting, shape, materials, and textures.
Because our dataset is too large for existing methods to process, we propose Sparse Voxel Light Field (SVLF), an efficient voxel-based light field approach for novel view synthesis that achieves comparable performance to NeRF~\cite{mildenhall2020nerf} on synthetic data, while being an order of magnitude faster to train and two orders of magnitude faster to render.
SVLF achieves this speed by relying on a sparse voxel octree, careful voxel sampling (requiring only a handful of queries per ray), and reduced network structure; as well as ground truth depth maps at training time.\let\thefootnote\relax\footnotetext{Dataset at \small{\url{cs.umd.edu/~mmeshry/projects/rtmv}}}
Our dataset is generated by NViSII~\cite{morrical2021nvisii}, a Python-based ray tracing renderer, which is designed to be simple for non-experts to use and share, flexible and powerful through its use of scripting, and able to create high-quality and physically-based rendered images.
Experiments with a subset of our dataset allow us to compare standard methods like NeRF~\cite{mildenhall2020nerf} and mip-NeRF~\cite{barron2021mip} for single-scene modeling, and pixelNeRF~\cite{yu2020pixelnerf} for category-level modeling, pointing toward the need for future improvements in this area.

\end{abstract}

\section{Introduction}

Since the publication of NeRF~\cite{mildenhall2020nerf},
there has been an explosion of interest in neural volume rendering for novel view synthesis \cite{barron2021mip,martinbrualla2020nerfw,liu2020neural,neff2021donerf,yu2020pixelnerf,park2021nerfies,tewari2021advances}.
This heightened attention of the research community is due to the impressive high-fidelity results that had long remained elusive but which are now possible.
Further, these techniques are able to capture the 3D geometry of the scene with precision \cite{Oechsle2021ICCV,wang2021neus,takikawa2021nglod}, thus opening up an entirely new family of algorithms for 3D scene reconstruction.

Despite this progress, it is not yet clear what are the limitations of such techniques.
Because of the difficulty of acquiring multi-view images of complex scenes with ground truth, only a handful of scenes has been available to date.
As a result, since the existing methods have only been tested on a small number of scenes, it is unclear how they will perform on a wider variety of scenes.  For example, can these methods simultaneously handle multiple objects, diverse materials and textures, challenging lighting conditions, and free camera poses?
Can they generalize to new complex scenes from a few views?
Questions such as these point to the need for a large-scale dataset to evaluate algorithms for novel view synthesis.

\begin{figure}
\centering
\resizebox{0.9\linewidth}{!}{
\begin{tabular}{ccc}
\rotatebox{90}{\footnotesize \ \ \ \ \ \ \ \ \ \ \ \ \ Goog. Scan.} &
\includegraphics[width=0.42\columnwidth]{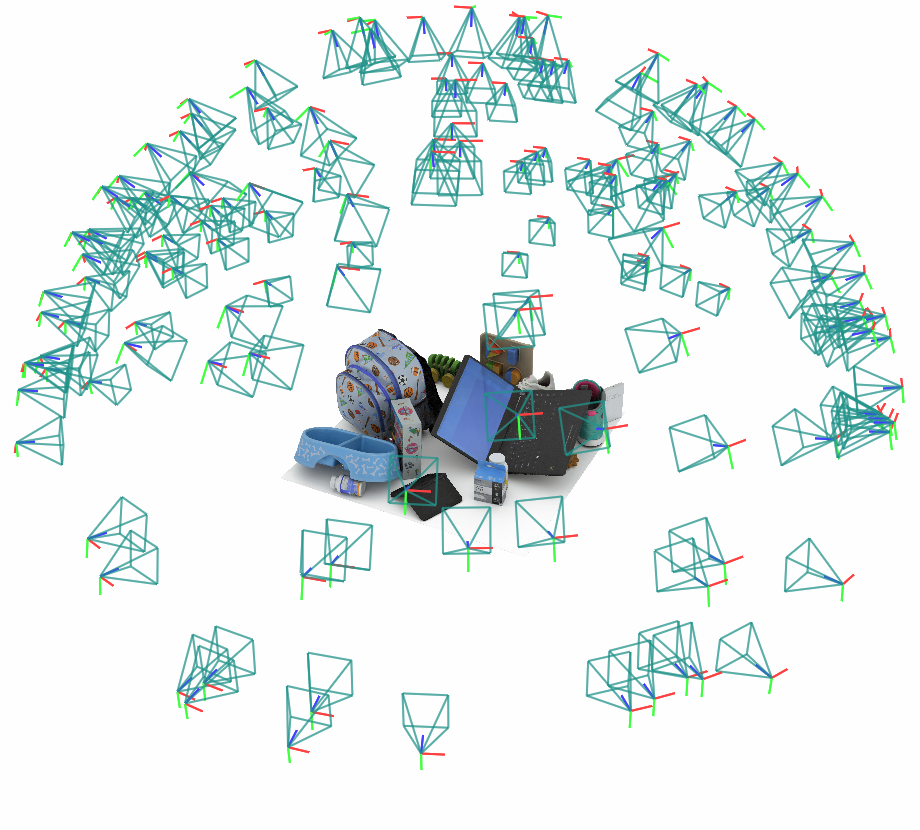} &
\includegraphics[width=0.42\columnwidth]{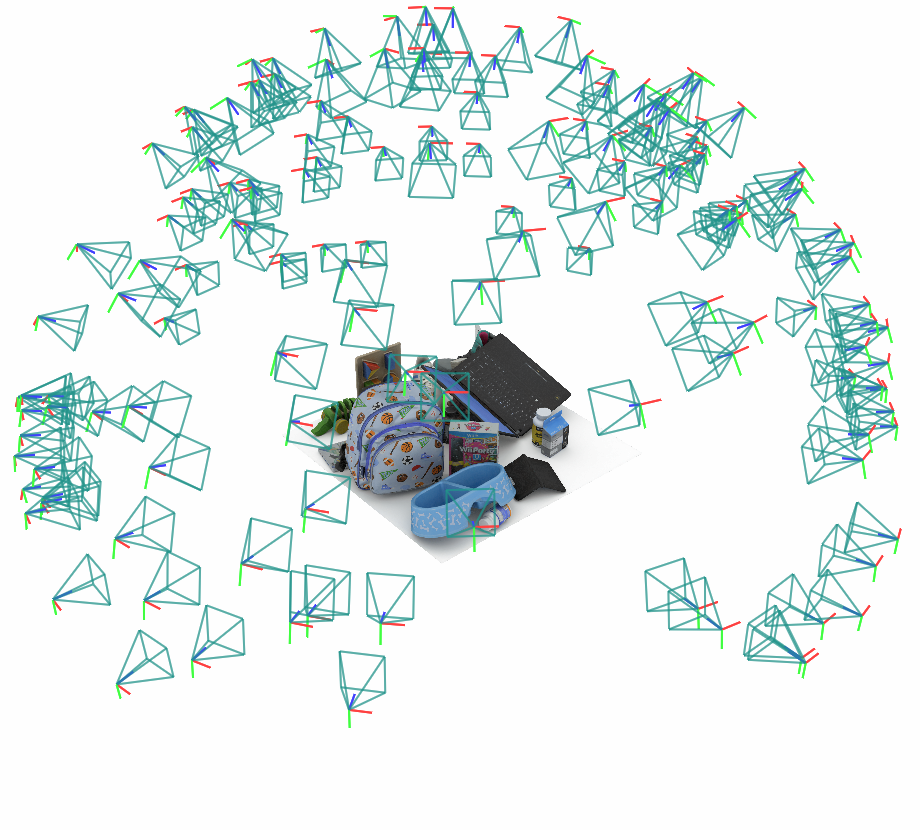} \\
\vspace{-0.15cm}
\rotatebox{90}{\footnotesize \ \ \ \ \ \ \ \ \ \ \ \ \ \ \ \ \ \ \ \ \ \ ABC} &
\includegraphics[width=0.42\columnwidth]{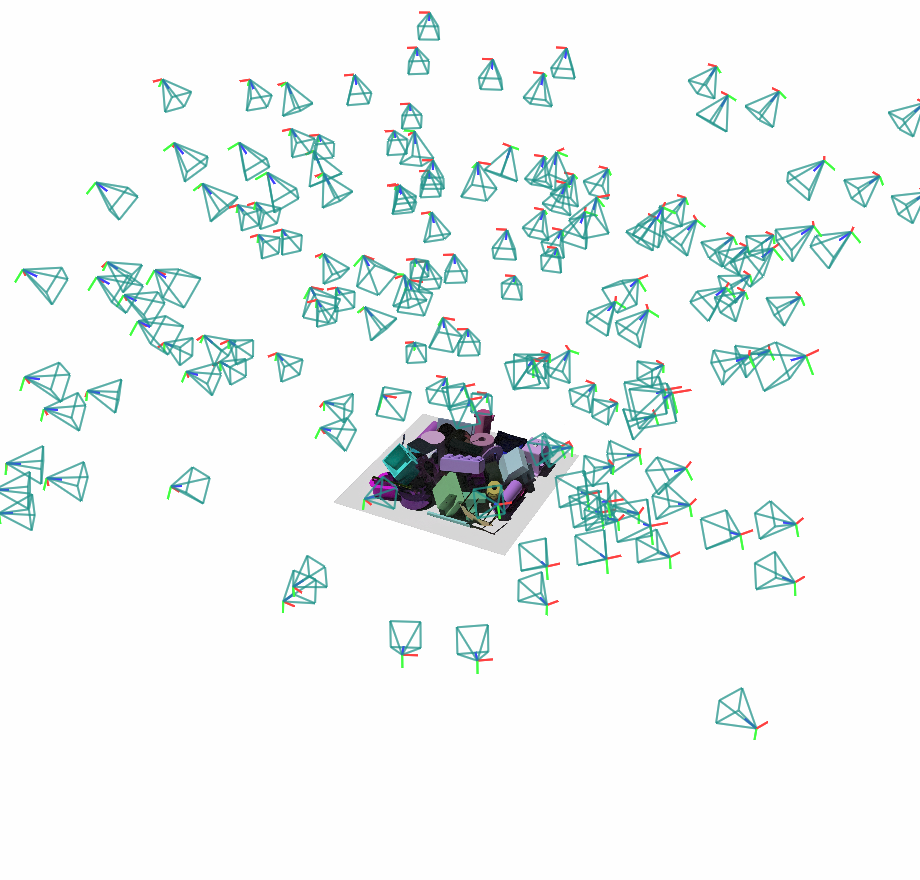} &
\includegraphics[width=0.42\columnwidth]{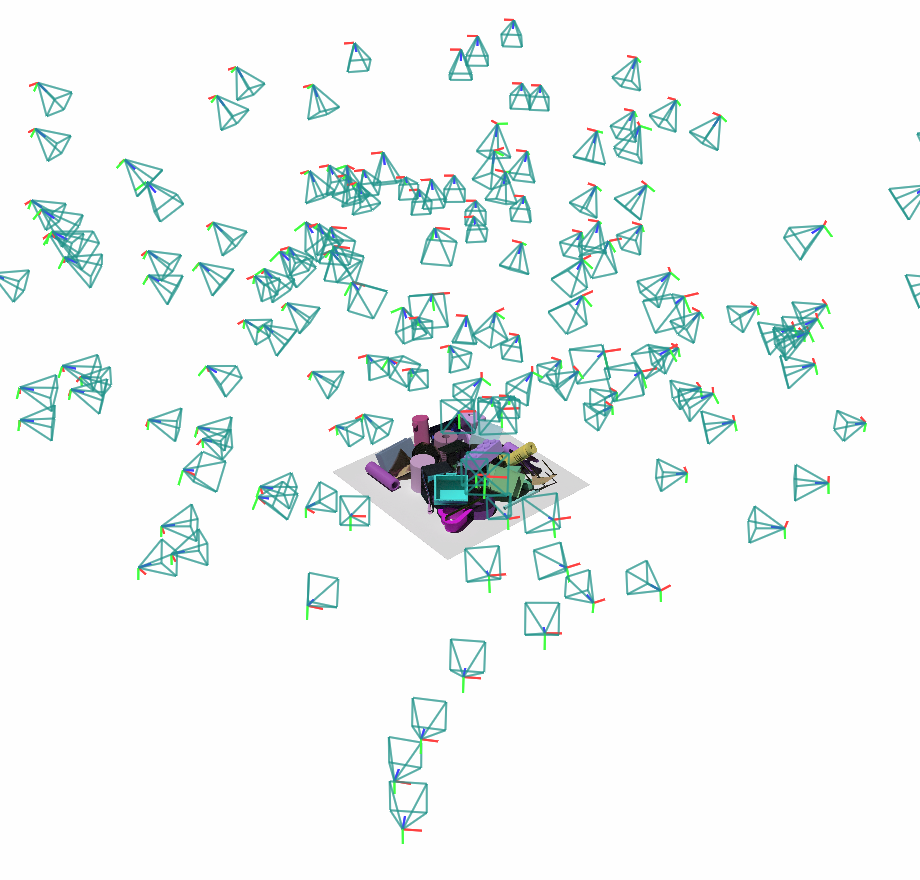} \\
\vspace{-0.15cm}
\rotatebox{90}{\footnotesize \ \ \ \ \ \ \ \ \ \ \ \ \ \ \ \ \ \ \ \ \ Bricks} &
\includegraphics[width=0.42\columnwidth]{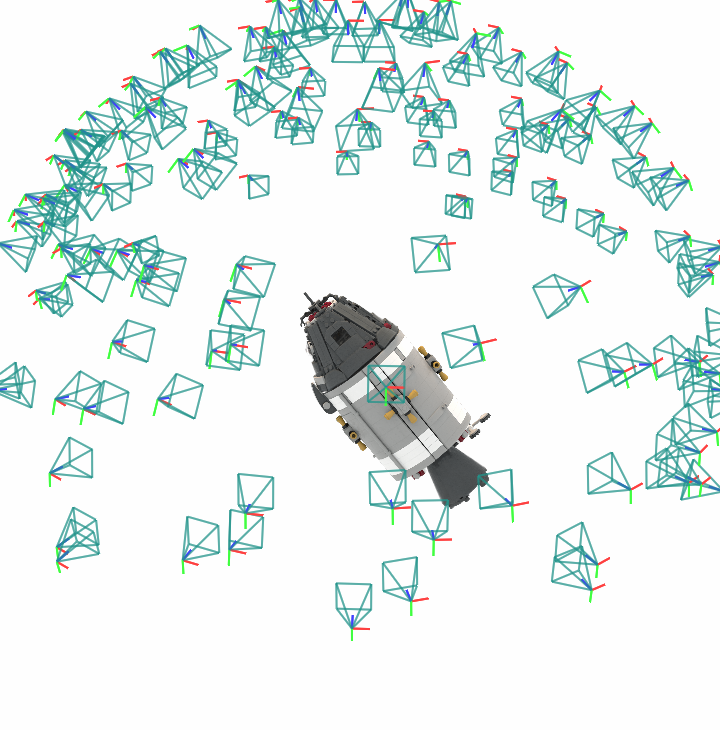} &
\includegraphics[width=0.42\columnwidth]{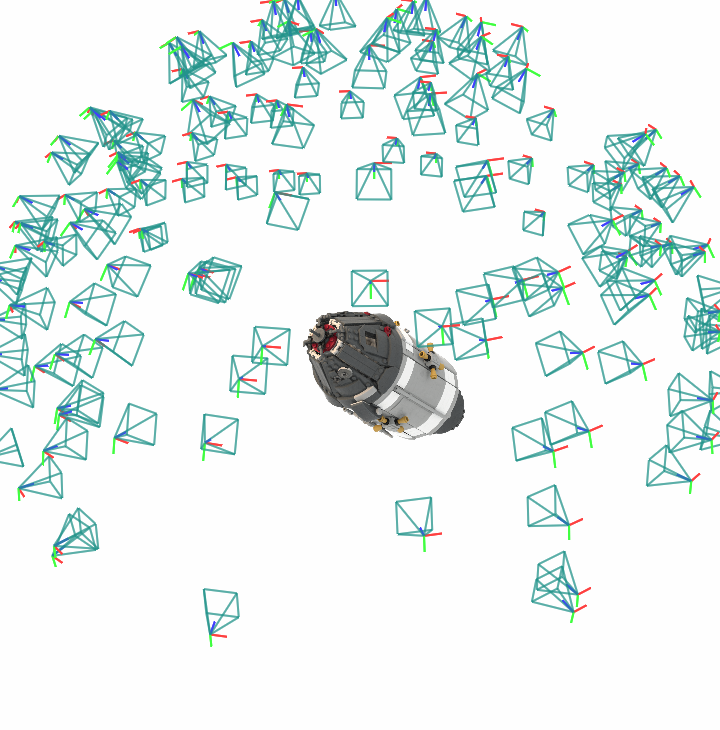} \\
\vspace{-0.15cm}
\rotatebox{90}{\footnotesize \ \ \ \ \ \ \ \ \ \ \ \ \ \ Amz. Ber.} &
\includegraphics[width=0.42\columnwidth]{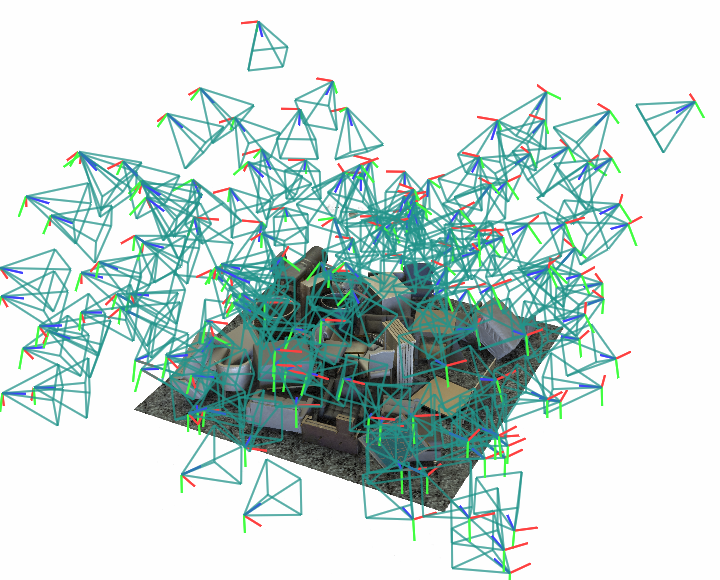} &
\includegraphics[width=0.42\columnwidth]{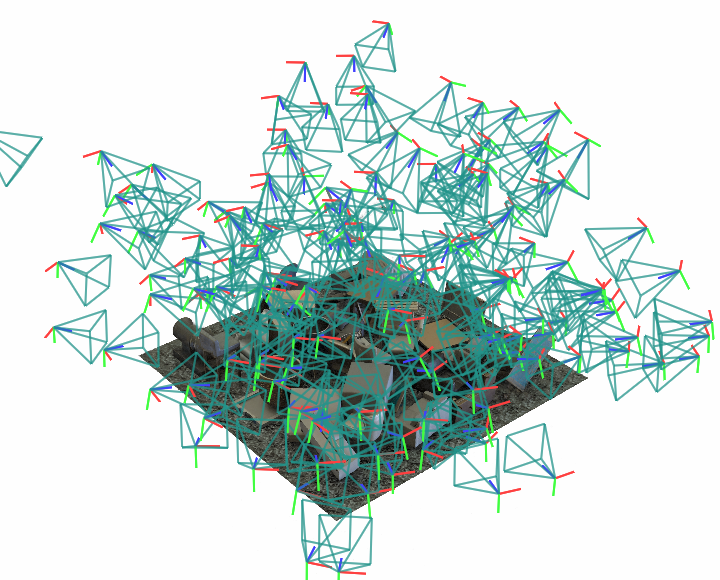} \\
\end{tabular}
}
\caption{
	Camera distributions for a single scene per environment, for each scene we present 2 example view points.
}
\label{fig:cam_single}
\vspace{-0.4cm}
\end{figure}

In this paper we propose to address this problem by sharing a large-scale dataset for novel view synthesis.
Comparison to existing datasets is shown in Table~\ref{tab:related_work}.
Our dataset consists of nearly 2000 scenes composed of multiple objects, with complex materials and textures illuminated by various light sources (see Fig.~\ref{fig:header}).
We selected these objects from 4 distinct sources of large collections, thus providing significant variety to the dataset.
Approximately 300k high-resolution ($1600 \times 1600$) synthetic images were created using ray tracing to ensure high fidelity, with the virtual camera placed either on a hemisphere or freely within the environment (see Fig.~\ref{fig:cam_single}).
Using synthetic data enables the following advantages:  1) noise-free annotations, 2) rich metadata that otherwise would not be possible, and 3) full control over variations.
Moreover, recent efforts at real-time ray tracing have made tremendous progress in reducing the sim-to-real gap for
different computer vision applications, {\em e.g.}, pose estimation~\cite{labbe2020} or facial analysis~\cite{wood2001iccv:fakeit}.
As a result, synthetic data is becoming increasingly important for training and evaluating deep networks~\cite{dosovitskiy2015iccv:flychairs,tremblay2018wad:car,tremblay:corl2018}.

We use this dataset to benchmark several existing algorithms, thus highlighting existing limitations and pointing out the need for future research.
However, we are only able to evaluate available methods on a small subset of our dataset due to their high computational demands.
To reduce the computational cost, we propose a novel algorithm called SVLF (Sparse Voxel Light Field) that is an order of magnitude faster than NeRF to train on synthetic scenes, and nearly two orders of magnitude faster to render.
This speedup is achieved by a novel combination of an octree representation, querying each voxel only once, and using a much smaller network, as well as leveraging depth maps for training.
We show that SVLF achieves results comparable to NeRF while being much more computationally efficient.

Our dataset has many uses beyond evaluation.
First, a large dataset like ours is useful for training neural networks by exposing the network to more variety than is possible with existing datasets.
Second, the size of the dataset allows for investigating few-show view synthesis, where an algorithm trained on a larger set of scenes is then able to process new scenes without training from scratch.
Third, because the synthetic data allows for rich metadata such as ground truth depth, geometry, camera poses, object positions, and so forth, it facilitates research on networks that require these inputs for training (\eg, novel view synthesis algorithms that train on ground truth depth).
Finally, the dataset can be used for problems other than novel view synthesis, such as 3D reconstruction and pose estimation.

\begin{table}
\begin{footnotesize}
    \centering
    \resizebox{\columnwidth}{!}{
    \begin{tabular}{lccccccccc}

        \toprule
         Datasets & \cite{mildenhall2020nerf} & \cite{Knapitsch2017} & \cite{yao2020blendedmvs} & \cite{jensen2014large} & \cite{sitzmann2019srns,yu2020pixelnerf} & \cite{mildenhall2019llff}& \cite{reizenstein2021common}&\textbf{Ours}\\
         \midrule
         scenes & 8 & 15 & 502 & 124 & 3511&24 &20k &2000                  \\
         resolution & 800 & 1080 & 1536 & 1200 & 128&480 &1088& 1600               \\
         multi-object & \redxmark & \redxmark & \greencheckmark & \greencheckmark & \redxmark & \redxmark & \redxmark & \greencheckmark \\
         free camera & \redxmark & \greencheckmark & \greencheckmark & \redxmark & \redxmark & \redxmark & \redxmark & \greencheckmark          \\
         full GT & \greencheckmark & \redxmark & \redxmark & \redxmark & \greencheckmark & \redxmark & \redxmark & \greencheckmark                          \\
         HDR export & \redxmark & \redxmark & \redxmark & \redxmark & \redxmark & \redxmark & \redxmark & \greencheckmark \\
         \bottomrule
    \end{tabular}
    }
    \caption{Comparison of novel view synthesis datasets.
    From top to bottom:  the number of scenes, image resolution (smallest dimension), whether the scene is composed of multiple objects, whether the camera can freely move about the scene (as opposed to front-facing, hemisphere, etc.), whether ground truth (GT) information is available (\eg, depth, lighting, object material), and whether high dynamic range (HDR) is exported.}
    \label{tab:related_work}
    \vspace{-0.3cm}
\end{footnotesize}
\end{table}

\section{Related work}

\medskip \noindent \textbf{Novel view synthesis.} Novel view synthesis is a long-standing problem in computer vision and graphics.
Image-based rendering (IBR) techniques~\cite{gortler1996lumigraph,debevec1996modeling,buehler2001unstructured,sinha2009piecewise,chaurasia2011silhouette,chaurasia2013depth,penner2017soft,riegler2020free} synthesize a novel view from a set of reference images by computing blend weights of nearby reference views.
Another class of work builds a volumetric representation of the scene such as voxel grids~\cite{kar2017learning,penner2017soft,tulsiani2017multi,henzler2020learning,lombardi2019neural} or multi-plane images (MPI)~\cite{choi2019extreme,srinivasan2019pushing,zhou2018stereo,flynn2019deepview,li2020crowdsampling,mildenhall2019llff}.
To render such representations, techniques like alpha compositing or volume rendering are typically used to synthesize novel views.

\medskip \noindent \textbf{Neural implicit representations.}
More recently,
neural implicit representations~\cite{Park_2019_CVPR,Occupancy_Networks,chen2019learning} have shown great potential for novel view synthesis by storing scene properties as the weights of a fully connected network.
Most notably, NeRF~\cite{mildenhall2020nerf} optimizes a 5D radiance field that maps a point and viewing direction to its density and color values, which can then be integrated for rendering.
Mip-NeRF~\cite{barron2021mip} uses integrated positional encoding over a cone frustum to render anti-aliased multi-scale images.
NeRF has opened the door to many exciting research directions~\cite{tewari2021advances,dellaert2020neural,barron2021mip,kaizhang2020,rebain2021derf,tancik2021learned,park2021nerfies,xian2021space,li2021neural,pumarola2020d,du2021nerflow,Gafni_2021_CVPR,Gao-portraitnerf,martinbrualla2020nerfw,Schwarz2020NEURIPS,chanmonteiro2020pi-GAN}.
However, NeRF (and closely-related variants) are slow to train and render, as rendering a single pixel requires evaluating hundreds of samples along a ray, each constituting an expensive neural network query.
To increase speed, some works~\cite{yu2021plenoctrees,liu2020neural} utilize an octree to prune empty space and only sample points close to the surface.
Another alternative, DONeRF~\cite{neff2021donerf}, utilizes depth images to train a depth prediction network, thus achieving high-quality synthesis by evaluating only a few point samples around the estimated depth.  Instant-NGP~\cite{mueller2022instant}, which was developed concurrently with this work, is yet another alternative that uses multiple tricks to achieve orders of magnitude speedup over NeRF.

While NeRF focuses on fitting single scenes,
several works aim to generalize across multiple scenes~\cite{grf2020,yu2020pixelnerf,wang2021ibrnet,chen2021mvsnerf}.
First, a prior is learned over a feature volume by training on a dataset of multiple scenes. Then, presented only with few-shot images of a new scene, the learned prior is utilized to render novel views of the new scene.

\medskip \noindent \textbf{Data synthesizer.}~Given the high demand for large annotated datasets for deep learning, there has been an increase in both synthetic datasets \cite{tremblay2018arx:fat,ros2016cvpr:syn,mayer2016cvpr:flythings,handa2015arx:sn,zhang2016arx:unst,Richter_2016_ECCV,gaidon2016cvpr:vkitti} and in tools for generating such data \cite{kolve2017ai2,to2018ndds,unity2020perception,denninger2019blenderproc,xiang2020sapien}.
The Cycles renderer included in Blender
has been widely used in the research community for generating synthetic data because of its ray tracing ability~\cite{Gu_2017_CVPR,mildenhall2020nerf,iraci2013blender,sajjan2019cleargrasp,wood2015_iccv,ron2020expohd,robotpose_etfa2019_cheind}.
In an attempt to more easily generate synthetic images, Denninger {\em et al.}~\cite{denninger2019blenderproc} introduced an extension to Blender that renders objects falling onto a plane with randomized camera poses.
Other tools, like SAPIEN \cite{xiang2020sapien}, have also used the OptiX backend~\cite{optix} to optimize
ray tracing performance.
We believe that our proposed Python ray tracer adds to this suite by providing scripting capabilities, along with the ease of installation and development.

\medskip \noindent \textbf{Datasets for novel view synthesis.} The original NeRF paper~\cite{mildenhall2020nerf} introduced 8 synthetic scenes with
360-degree views.
Since their release, the accessibility of these scenes has allowed for rapid progress in novel view synthesis, yet, the small number of such scenes makes it difficult to assess performance at scale.
Wang~{\em et al.}~\cite{wang2021ibrnet} introduced a synthetic dataset including Google Scanned Objects.
Other datasets for training multi-view algorithms include
DTU~\cite{jensen2014large},
LLFF~\cite{mildenhall2019llff},
Tanks and Temples~\cite{Knapitsch2017},
Spaces~\cite{flynn2019deepview},
RealEstate10K~\cite{zhou2018stereo},
SRN~\cite{sitzmann2019srns, yu2020pixelnerf},
Transparent Objects~\cite{IchnowskiAvigal2021DexNeRF},
ROBI~\cite{yang2021robi},
CO3D~\cite{reizenstein2021common,henzler2021unsupervised},
SAPIEN~\cite{xiang2020sapien},
and
BlendedMVS~\cite{yao2020blendedmvs}.
See Table~\ref{tab:related_work} for further information.
During the preparation of this manuscript, Reizenstein {\em et al.}~\cite{reizenstein2021common,henzler2021unsupervised} introduced a large real dataset for
multi-view synthesis, using camera
views around the object.
We believe our proposed dataset nevertheless provides unprecedented scale, variety, and quality.

\section{Ray-Traced Multi-View Dataset}

\subsection{NViSII}
Recent advances in ray tracing have allowed researchers
to generate photo-realistic images with unprecedented speed and quality.
Such advancements are accessible via tools integrated within existing rendering frameworks with complex interfaces (\eg, \cite{denninger2019blenderproc}).
In contrast, we propose an accelerated standalone
Python-enabled ray tracing / path tracing renderer built on NVIDIA OptiX~\cite{optix} with a C++/CUDA backend.
The Python API allows for non-graphics experts to easily install the renderer,
quickly create 3D scenes with the full power that scripting provides, and is simply 
installed through {\tt pip} package management.\footnote{{\tt pip install nvisii}, see \url{https://nvisii.com/} for extended documentation.}

NViSII allows for complex visual material definitions using Disney's principled BSDF~\cite{burley2015extending} model.
This model takes sixteen parameters, whose combinations result in physically-plausible, well-behaved results across their entire range.
These parameters, which include color, metallic, transmission, and roughness components, have a perceptually linear effect on the appearance of the BSDF.\@
The linearity property is well-suited to uniform random sampling in order to create a variety of different materials, such as smooth and rough plastics, metals, and dielectrics (\eg, glass and water).
We drive these parameters using either scalar values or textures to create a diverse dataset.

Our solution also supports advanced rendering capabilities, such as multi-GPU ray tracing, physically-based light controls, accurate camera models (including defocus blur and motion blur), native headless rendering, and exporting various metadata (\eg, segmentation, motion vectors, optical flow, depth, surface normals, and albedo).

\subsection{Data generation}

We use NViSII~\cite{morrical2021nvisii} to generate 1927 unique scenes in total from 4 different environments, each environment being a collection of 3D models drawn from a different data source (see Tab.~\ref{table:highlevel}).
For each scene, we normalize it to be contained within a unit cube and generate 150 unique image frames.
Ground truth annotations include a depth/distance map, background alpha mask, segmentation mask, scene point cloud,
and various metadata (\eg, camera poses, entity poses, material properties, and light positions).
Images are exported in the `EXR' format as it allows for storing colors in a high dynamic range without tonemapping and/or gamma correction pre-applied.
Each image is rendered at 4000 samples per pixel. %
We use 2 types of camera placements: hemisphere and free (\eg, Fig.~\ref{fig:cam_single}). The former places the camera on the hemisphere surrounding the
scene, while the latter moves the camera freely within a cube avoiding collisions with the objects
(please refer to the supp.~material for the distribution of camera azimuth and elevation over the entire dataset).
Overall the dataset is orders of magnitude larger than what is typically used for novel
view synthesis.
Since it is time consuming for an algorithm like NeRF~\cite{mildenhall2020nerf} to process the entire dataset,
we also provide a smaller subset composed of 40 scenes, 10 from each environment.

\subsection{Environments}

These 4 different environments are as follows.

\begin{table}
\begin{footnotesize}
\begin{tabular}{rccc}
\toprule
Environment  	  & Camera type & Lighting & \# scenes\\
\midrule
Google Scanned 	  & hemisphere  & white dome 		& 300\\
ABC   	 	      & free 		& spotlight 		& 300\\
Bricks  	  	  & hemisphere  & white dome \& sun & 1027\\
Amazon-Berkeley   & free 		& HDRI dome light 	& 300\\
\bottomrule
\end{tabular}
\end{footnotesize}
\caption{High-level parameters for the four environments.}
\label{table:highlevel}
\vspace{-0.4cm}
\end{table}

\begin{figure*}
    \vspace{-0.1cm}
    \centering
    \includegraphics[width=.9\linewidth]{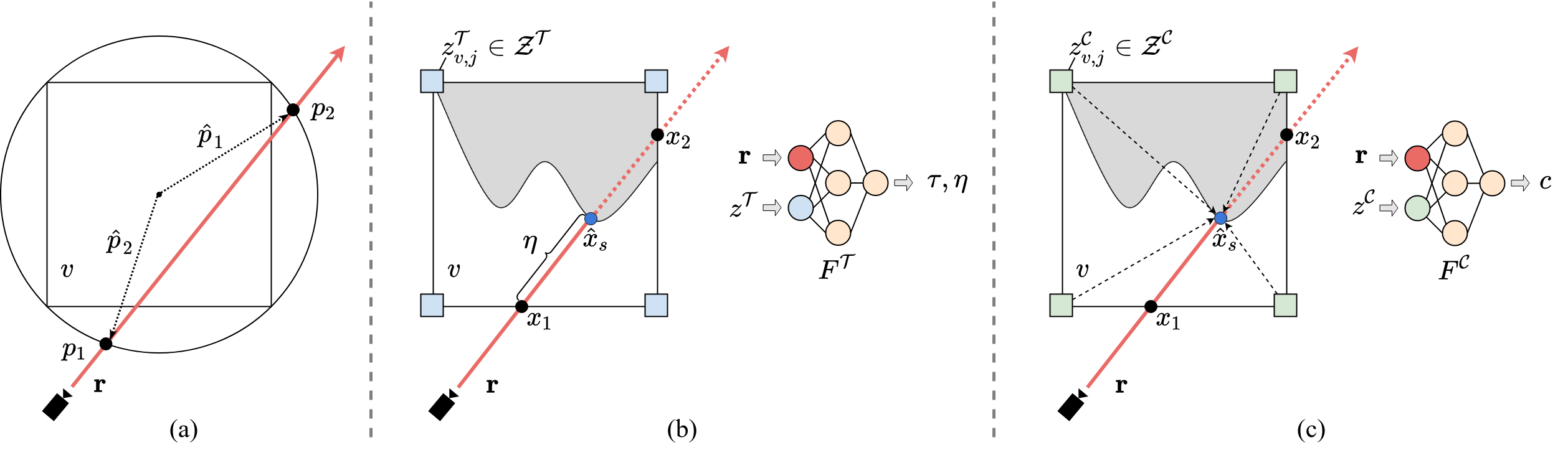}
    \vspace{-0.3cm}
    \caption{SVLF overview for a given voxel $v$. (a) Rays are parameterized by the normalized intersection points with the bounding sphere,
    $\bfr = [\Tilde{p}_1 \,\,|\,\, \Tilde{p}_2]$, where $\tilde{p}_i = \hat{p}_i/\|\hat{p}_i\|$.
        (b) SVLF first estimates the optical thickness $\tau$ and a within-voxel depth $\eta$ to the surface hit, if any.
        (c) Color features are sampled at the estimated surface point $\hat{x}_s$ and a color decoder predicts the final ray color $c$.}
    \label{fig:svlf}
    \vspace{-0.5cm}
\end{figure*}

\medskip \noindent \textbf{Google Scanned Objects.} All the $\sim$1000 Google scanned models~\cite{GoogleScanned} were used to generate 300 falling objects scenes.
In each scene, we randomly selected 20 3D models, added collision shapes, and let the objects fall onto each other and onto a table using the PyBullet physics simulator~\cite{coumans2019:pybullet}.
We applied a random material to the texture since BRDF materials are not provided with the models (they are mostly plastic-like).
Note that the provided textures from these 3D models have light baked into the textures, which can cause the object to look unrealistic, {\em e.g.}, having highlights where it should be dark.
The camera remains at a constant distance from the center and always points toward the center of the scene.
The scenes are lit by a single white dome light.
This dataset is the most similar to NeRF's synthetic data~\cite{mildenhall2020nerf}.
This is also the easiest of our 4 environments: it has a small number of objects and
views distributed on the hemisphere.

\medskip \noindent \textbf{ABC.} We used all the 3D models ($\sim$1.5M) from the ABC dataset~\cite{Koch_2019_CVPR}.
In each scene, we randomly selected 50 3D models and scaled them so that they are all of a similar size.
Similar to the previous environment, we constructed 300 scenes of falling objects using physical simulation.
Each object was also given a random BRDF material (plastic, metallic, rough metallic, {\em etc.}), and a random bright color, so that this environment has multiple view-dependent materials.
The scenes were illuminated by a uniform dome light, and by a bright point light simulating a directional light to produce hard shadows.
The camera was allowed to freely move within the unit cube and look at any of the objects, thus producing both close and far images.
As a result, this environment has the most challenging viewpoints.

\medskip \noindent \textbf{Bricks.} We downloaded 1027 unique models from \href{https://mecabricks.com}{Mecabricks}.
For each model, we generated a single scene where the model was placed in the middle of the scene.
The camera was randomly placed on a hemisphere at a fixed distance from the center.
The camera was aimed at random locations within 1/10 of the unit volume used to scale the object,
thus producing images that are not centered on the model.
Each scene was illuminated by a white dome light and a warm sun placed randomly on the horizon.
These detailed scenes are challenging due to the variety of textures and presence of self-reflections.
This is our largest environment, since it has 1027 scenes.

\medskip \noindent \textbf{Amazon-Berkeley.} Similar to ABC and Google Scanned Objects, we loaded 40 3D models from the \href{https://amazon-berkeley-objects.s3.amazonaws.com/index.html}{Amazon Berkeley Objects (ABO)  dataset} \cite{collins2021abo}, scaled them, and let them fall onto a plane.
Since the 3D models are accompanied with high-quality textures and material definitions, we light the scene with a full HDRI map and apply a random texture on the floor, both from \href{https://polyhaven.com}{Poly Haven}.
Similar to ABC scenes, the camera was allowed to move freely within the unit cube and to look at any object.
Also similar to ABC, the materials are quite challenging, such as metallic (mirror-like) objects.
We believe this is the most challenging environment of the dataset offering interesting views, materials, textures, lighting, and photo-realistic backgrounds.

\section{Sparse Voxel Light Field (SVLF)} \label{sec:method}

Given the size and complexity of our dataset,
in this section we present an efficient and lightweight multi-view voxel-based algorithm for novel view synthesis.
At a high level,
our method learns a voxel-based function that maps a ray $\bfr$ to the optical thickness $\tau_{v_i}$\footnote{Optical thickness is defined, as in~\cite{novak2018monte}, as the integral of the density along a segment of a ray, so that $\alpha=(1 - e^{-\tau_{v_i}})$ is the opacity.} and color $c_{v_i}$ associated with a particular voxel $v_i$.
This mapping is realized by evaluating two small decoder networks.
Rendering a single pixel consists of a volumetric integration that amounts to only a handful of network queries, determined by the number of voxels intersected along the ray.
This is in contrast to evaluating hundreds of queries along the ray as in NeRF~\cite{mildenhall2020nerf}.

\begin{figure*}
\centering
\resizebox{0.9\linewidth}{!}{
\begin{tabular}{ccc|cc|cc|cc}
&
\multicolumn{2}{c}{Google Scanned Objects} &
\multicolumn{2}{c}{ABC} &
\multicolumn{2}{c}{Bricks} &
\multicolumn{2}{c}{Amazon Berkeley} \\
\rotatebox{90}{\small Ground truth} &
\includegraphics[trim=0 0 0 -5, width=0.2\columnwidth]{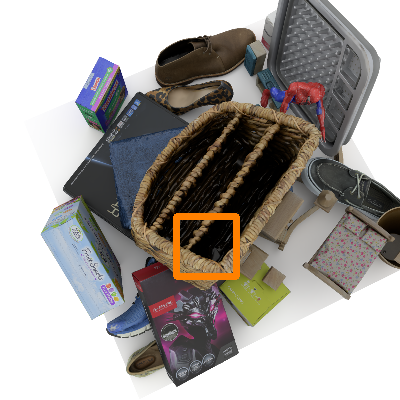} &
\includegraphics[trim=0 0 0 -5, width=0.2\columnwidth]{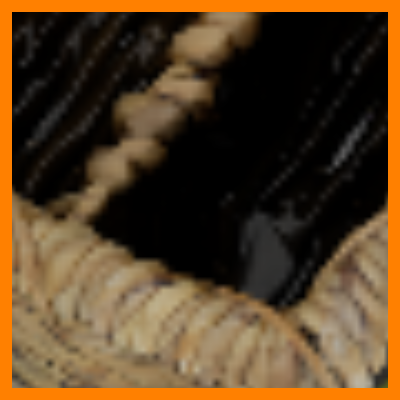} &

\includegraphics[trim=0 0 0 -5, width=0.2\columnwidth]{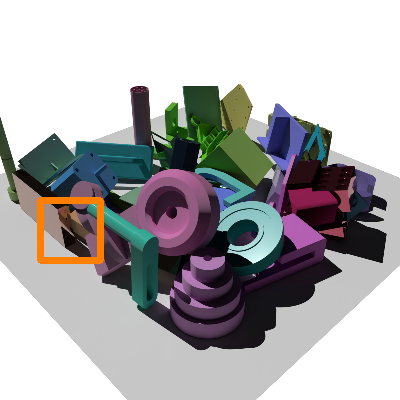} &
\includegraphics[trim=0 0 0 -5, width=0.2\columnwidth]{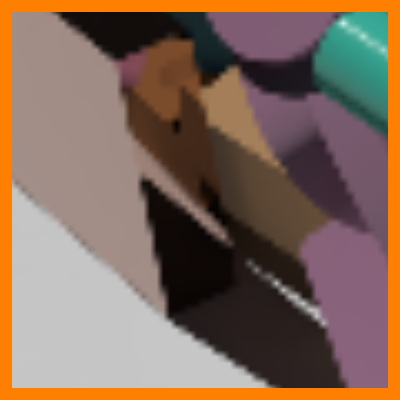} &

\includegraphics[trim=0 0 0 -5, width=0.2\columnwidth]{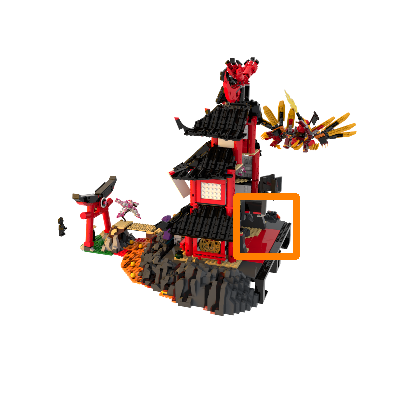} &
\includegraphics[trim=0 0 0 -5, width=0.2\columnwidth]{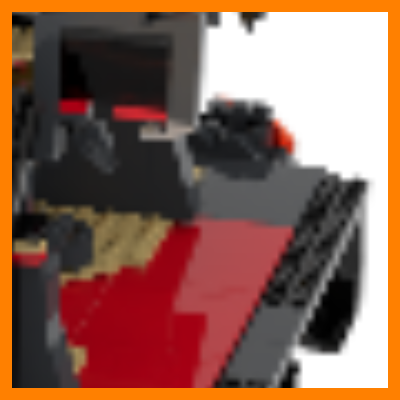} &

\includegraphics[trim=0 0 0 -5, width=0.2\columnwidth]{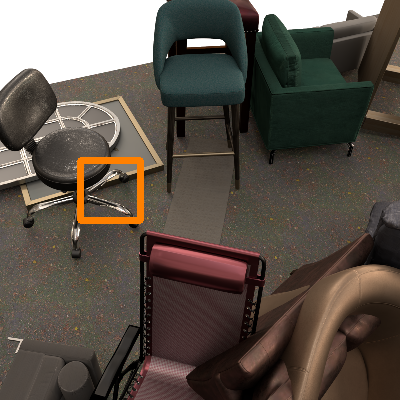} &
\includegraphics[trim=0 0 0 -5, width=0.2\columnwidth]{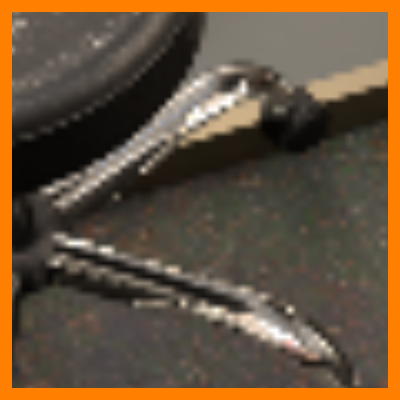}

\\
\rotatebox{90}{\small \ \ \ \ \ \ SVLF} &
\includegraphics[trim=0 0 0 -5, width=0.2\columnwidth]{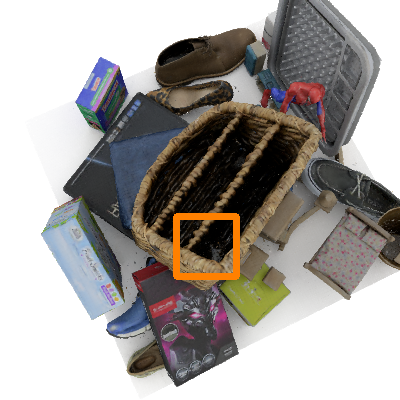} &
\includegraphics[trim=0 0 0 -5, width=0.2\columnwidth]{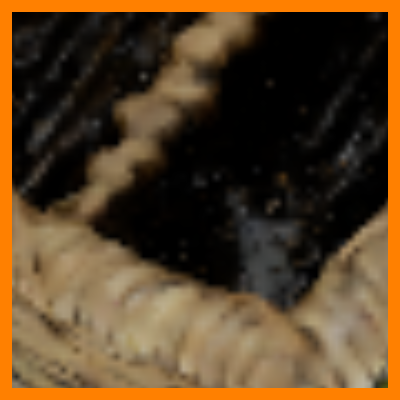} &

\includegraphics[trim=0 0 0 -5, width=0.2\columnwidth]{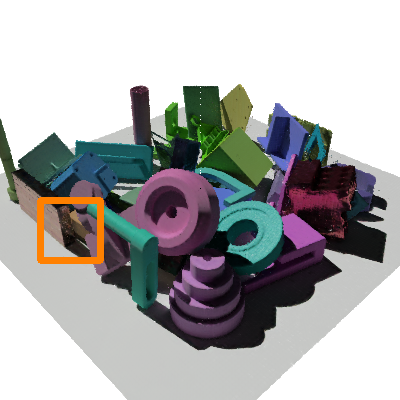} &
\includegraphics[trim=0 0 0 -5, width=0.2\columnwidth]{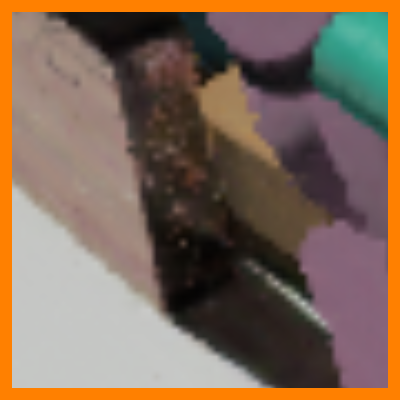} &

\includegraphics[trim=0 0 0 -5, width=0.2\columnwidth]{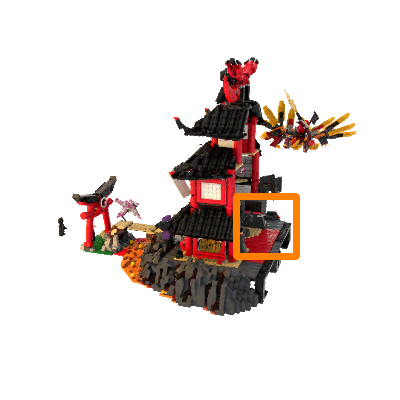} &
\includegraphics[trim=0 0 0 -5, width=0.2\columnwidth]{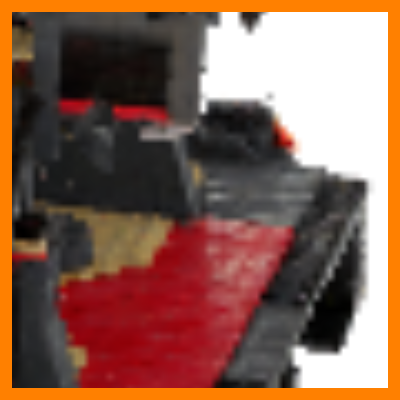} &

\includegraphics[trim=0 0 0 -5, width=0.2\columnwidth]{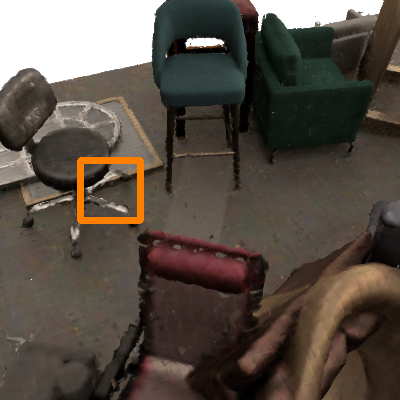} &
\includegraphics[trim=0 0 0 -5, width=0.2\columnwidth]{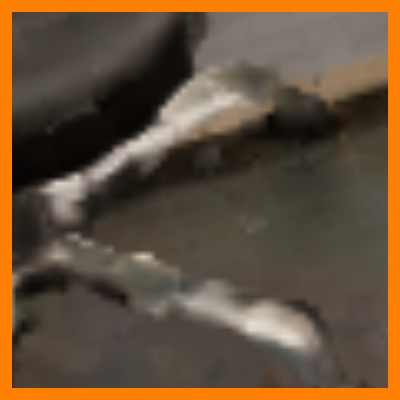}

\\
\rotatebox{90}{\small Ins.-NGP} &
\includegraphics[trim=0 0 0 -5, width=0.2\columnwidth]{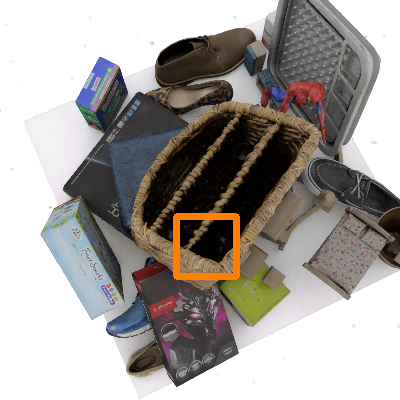} &
\includegraphics[trim=0 0 0 -5, width=0.2\columnwidth]{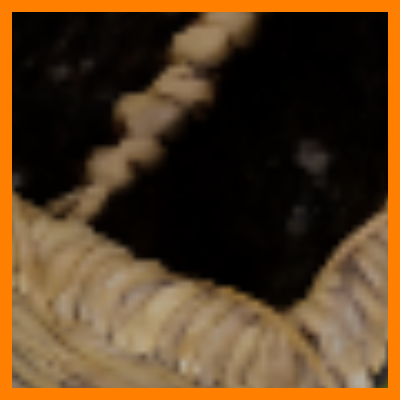} &

\includegraphics[trim=0 0 0 -5, width=0.2\columnwidth]{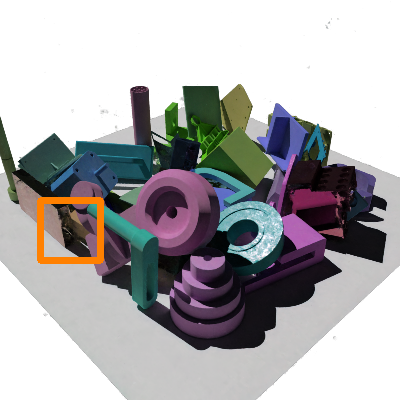} &
\includegraphics[trim=0 0 0 -5, width=0.2\columnwidth]{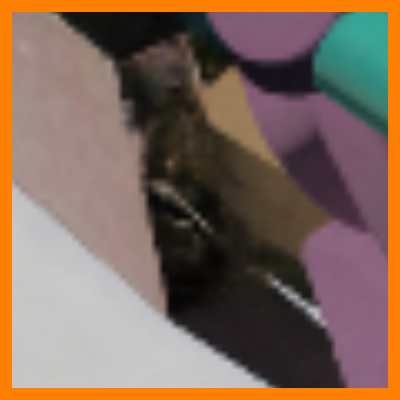} &

\includegraphics[trim=0 0 0 -5, width=0.2\columnwidth]{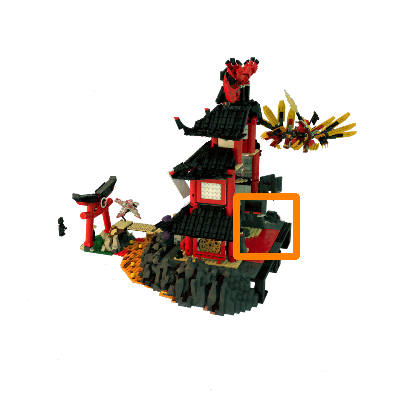} &
\includegraphics[trim=0 0 0 -5, width=0.2\columnwidth]{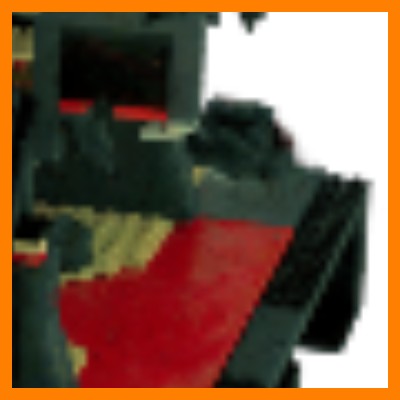} &

\includegraphics[trim=0 0 0 -5, width=0.2\columnwidth]{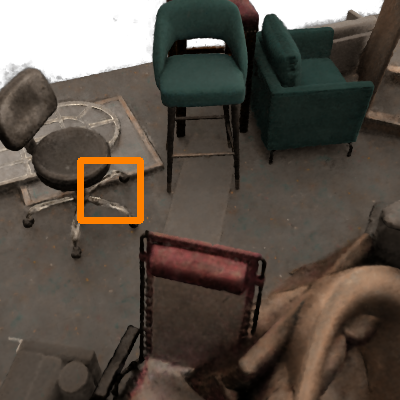} &
\includegraphics[trim=0 0 0 -5, width=0.2\columnwidth]{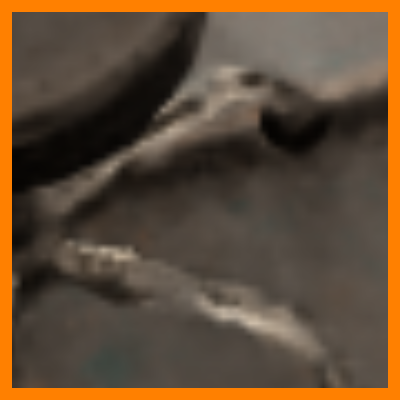}

\\
\rotatebox{90}{\small mip-NeRF} &
\includegraphics[trim=0 0 0 -5, width=0.2\columnwidth]{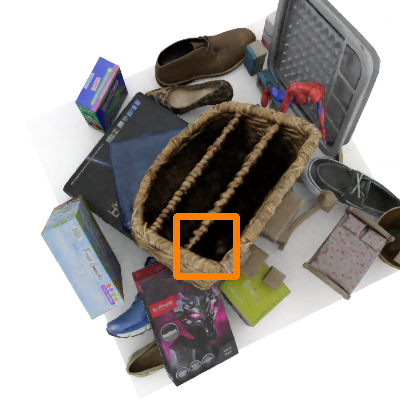} &
\includegraphics[trim=0 0 0 -5, width=0.2\columnwidth]{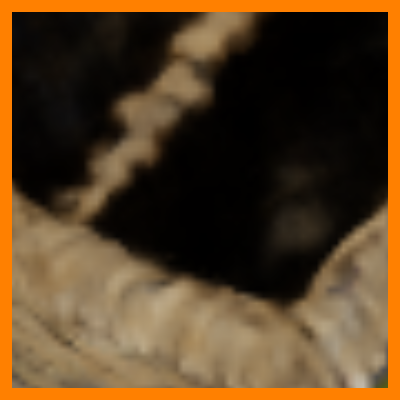} &

\includegraphics[trim=0 0 0 -5, width=0.2\columnwidth]{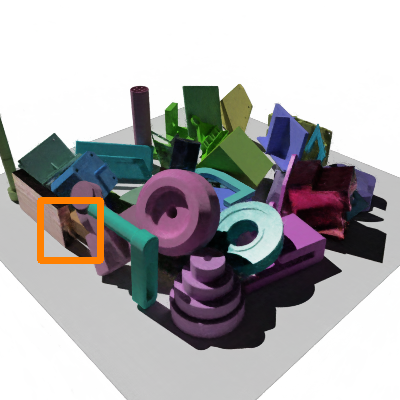} &
\includegraphics[trim=0 0 0 -5, width=0.2\columnwidth]{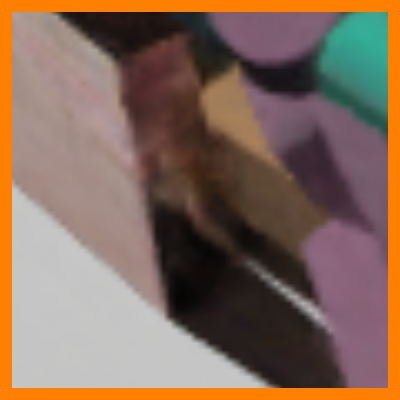} &

\includegraphics[trim=0 0 0 -5, width=0.2\columnwidth]{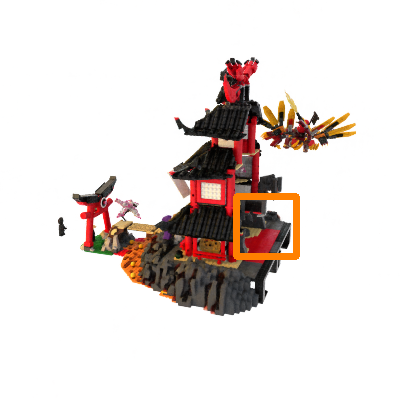} &
\includegraphics[trim=0 0 0 -5, width=0.2\columnwidth]{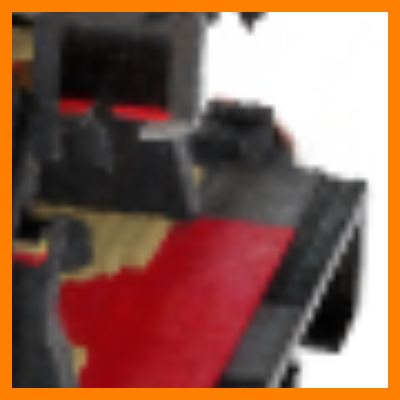} &

\includegraphics[trim=0 0 0 -5, width=0.2\columnwidth]{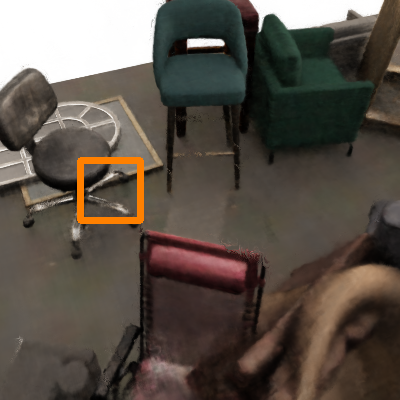} &
\includegraphics[trim=0 0 0 -5, width=0.2\columnwidth]{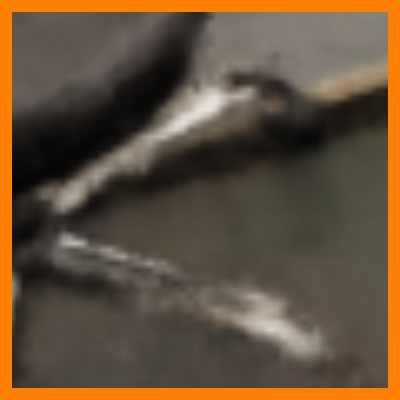}

\\
\rotatebox{90}{\small \ \ \ \ \ \ NeRF} &
\includegraphics[trim=0 0 0 -5, width=0.2\columnwidth]{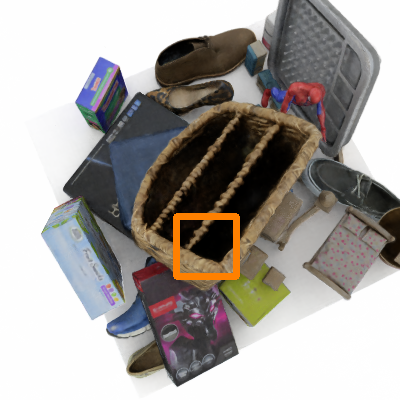} &
\includegraphics[trim=0 0 0 -5, width=0.2\columnwidth]{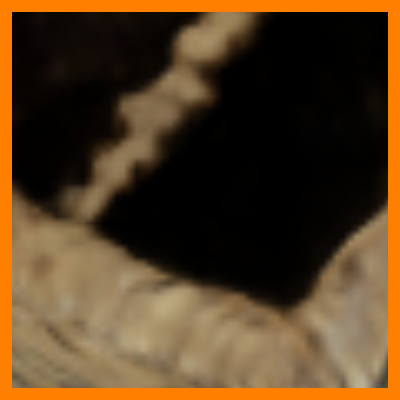} &

\includegraphics[trim=0 0 0 -5, width=0.2\columnwidth]{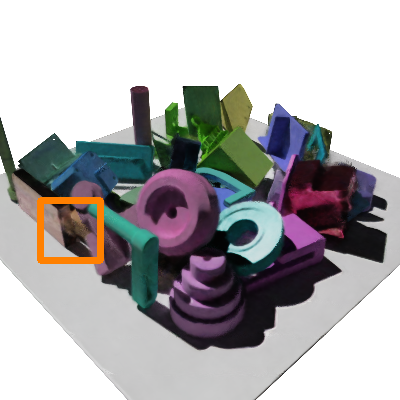} &
\includegraphics[trim=0 0 0 -5, width=0.2\columnwidth]{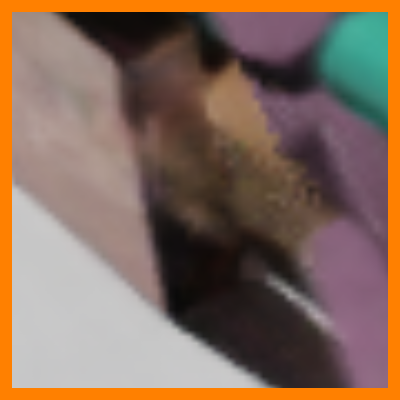} &

\includegraphics[trim=0 0 0 -5, width=0.2\columnwidth]{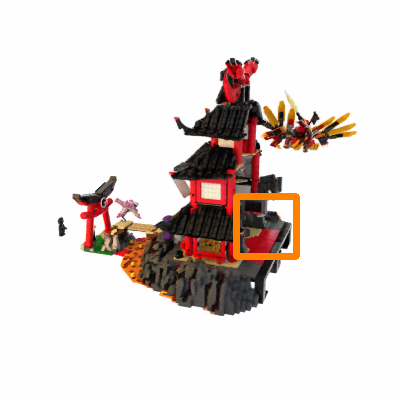} &
\includegraphics[trim=0 0 0 -5, width=0.2\columnwidth]{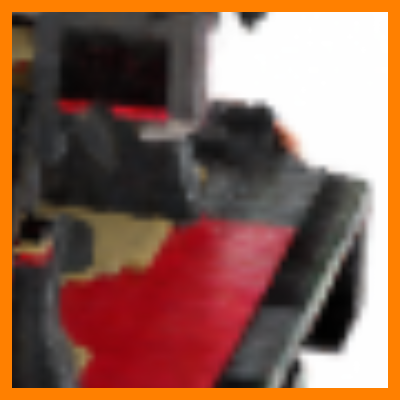} &

\includegraphics[trim=0 0 0 -5, width=0.2\columnwidth]{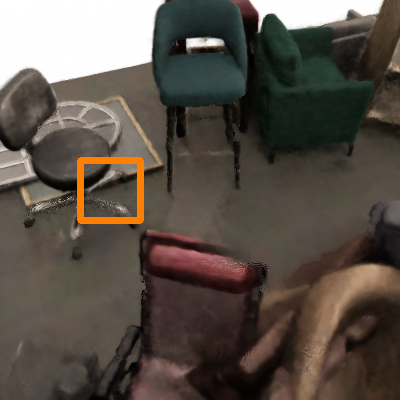} &
\includegraphics[trim=0 0 0 -5, width=0.2\columnwidth]{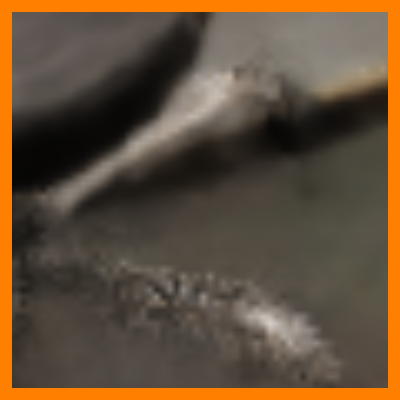}

\end{tabular}
}
\vspace{-0.2cm}
\caption{
Single-scene novel view synthesis results of the baselines on a sample scene from each environment.
}
\label{fig:qual_comp}
\vspace{-0.4cm}
\end{figure*}

\subsection{Representation}
\label{sec:svlf_representation}

\paragraph{Scene volume.}
Given depth maps provided with our dataset,
we partition the scene into a voxel grid $\mathcal{V}$ and represent the space using an octree~\cite{takikawa2021nglod}.
This allows for efficient memory utilization as the octree only represents voxels that contain a surface.
We use $\mathcal{Z}$ to represent a collection of feature vectors over the feature volume.
Each voxel $v$ in the octree stores learnable feature embeddings $z_{v,j} \in \mathcal{Z}$ at each of its eight corners ($j \in \{1, 2, \dots 8\}$), where feature embeddings are shared between spatially adjacent voxels.
In other words, each vertex in the octree is associated with one feature embedding.
The feature embedding $z_p$ is defined for any point $p \in \R^3$ in the space as $z_p = \psi(p; \mathcal{Z})$, where $\psi(.)$ is the tri-linear interpolation between corner features of the voxel in which $p$ is contained.
The collection of feature embeddings $\mathcal{Z}$ is learned to encode local properties of each voxel, while the tri-linear interpolation and shared features between adjacent voxels ensure a degree of continuity in the feature space. %

\medskip \noindent \textbf{Ray parameterization.}
Since we aim at learning voxel-based functions, we parameterize a given ray $\bfr$ in the local coordinates of each voxel.
Specifically, as shown in Fig.~\ref{fig:svlf}~(a), we compute the intersection points $p_1, p_2 \in \R^3$ of $\bfr$ with the minimum bounding sphere around the voxel, define vectors $\hat{p}_i$ from the voxel origin to these intersections, then scale them to unit norm:  $\Tilde{p}_i = \hat{p}_i / \| \hat{p}_i \|$, $i=1,2$.
This gives us the overparameterized representation $\bfr = [\begin{matrix}\Tilde{p}_1 \,\,|\,\, \Tilde{p}_2\end{matrix}] \in \mathbb{R}^6$.

\subsection{Voxel-based light field function}
\label{sec:svlf_fn}

The collection $\mathcal{Z}$ is split into feature embeddings $\mathcal{Z}^\mathcal{T}$ for optical thickness, and feature embeddings $\mathcal{Z}^\mathcal{C}$ for color.
Each of these sub-collections has an associated small decoder network:  $\fdensity$ for optical thickness, and $\fcolor$ for color.

Given a voxel $v$ and a ray $\bfr$, we first compute the ray-voxel intersection points $x_1, x_2$ (see Fig.~\ref{fig:svlf}~(b)) using an efficient ray-AABB intersection algorithm~\cite{majercik2018ray}.
We then estimate the optical thickness and color of the ray in two steps.
The first step estimates whether the ray hits a surface within $v$, and produces an estimate for the location of the hit surface point $\hat{x}_s$ as a convex combination of $x_1$ and $x_2$.
Specifically, we have:
\begin{equation}
    \tau, \eta = \fdensity(\bfr, z^\mathcal{T}),
\end{equation}
where $\tau$ is the estimated optical thickness,
$\eta \in [0, 1]$ is the mixing factor interpolating between $x_1, x_2$
, and $z^\mathcal{T} = [\begin{matrix}\psi(x_1; \zdensity ) \,\,|\,\, \psi(x_2; \zdensity )\end{matrix}]$ is the combined interpolated feature embedding at the two intersection points.
If $\tau\neq 0$, the surface point $\hat{x}_s$ is estimated as $\hat{x}_s = \eta x_1 + (1 - \eta) x_2$, where $\eta$ can be seen as a within-voxel depth.
Otherwise, if $\tau=0$, the ray does not hit any surface.
We estimate surface points to encourage continuous sampling of color features between adjacent voxels.

The second step samples color features $\zcolor$ at the estimated surface point $\hat{x}_s$ and computes the color value $c$ as
\begin{equation}
    c = \fcolor(\bfr, z^\mathcal{C}),
    \label{eq:ceqfc}
\end{equation}
where $z^\mathcal{C} = \psi(\hat{x}_s; \zcolor)$ is the interpolated feature embedding at $\hat{x}_s$ (see Fig.~\ref{fig:svlf}~(c)).

\subsection{Rendering}
\label{sec:svlf_rendering}

To render a pixel from an SVLF, we perform volume rendering.
We first evaluate the color $c$ and optical thickness $\tau$ for all voxels intersected by the ray,
then we aggregate the color values using alpha compositing.

\medskip \noindent \textbf{Evaluating intersected voxels.}
We first compute a list of all voxels intersected by the ray.
We use the sparse ray-octree intersection algorithm of~\cite{takikawa2021nglod} that leverages the hierarchical octree structure and parallel scan kernels~\cite{merrill2017cub} for accelerated computation.
This step yields $x_1$ and $x_2$ for each intersected voxel.
We then evaluate the optical thickness $\tau$ and color $c$ at each intersected voxel as described in Sec.~\ref{sec:svlf_fn}.
Each voxel is evaluated only once by querying two lightweight decoder networks~$\fdensity$ and ~$\fcolor$ to compute the optical thickness and color, respectively.

\medskip \noindent \textbf{Aggregating color values along the ray.}
We use alpha compositing to compute the ray color $c(\bfr)$ as:
\begin{equation}
    c(\bfr) = \sum_{i=1}^N T_{v_i} (1 - e^{-\tau_{v_i}}) c_{v_i}, \quad T_{v_i} = \prod_{j < i} e^{-\tau_{v_j}}
    \label{eqn:rendering}
\end{equation}
where $N$ is the number of intersected voxels, $T_{v_i}$ is the transmittance up until voxel $v_i$, and voxels are indexed by the order of their intersection along the ray.

\begin{table*}
\centering
\begin{footnotesize}
\resizebox{0.82\linewidth}{!}{
\begin{tabular}{ccrclrclrclrcrrcr}
\toprule
\multirow{2}{*}{Method} & \multirow{2}{*}{Env.} & \multicolumn{9}{c}{Image-based metrics} & \multicolumn{6}{c}{Depth-based metrics} \\
\cmidrule(lr){3-11}\cmidrule(lr){12-17}%
& & \multicolumn{3}{c}{PSNR $\uparrow$} & \multicolumn{3}{c}{SSIM $\uparrow$} & \multicolumn{3}{c}{LPIPS $\downarrow$} & \multicolumn{3}{c}{RMSE ($10^3$) $\downarrow$} & \multicolumn{3}{c}{MAE ($10^3$) $\downarrow$} \\\midrule
NeRF &Goog. Scan. &29.627 &\hspace{-2.7ex}$\pm$ &\hspace{-2.6ex}1.863 &0.937 &\hspace{-2.7ex}$\pm$ &\hspace{-2.6ex}0.026 &0.046 &\hspace{-2.7ex}$\pm$ &\hspace{-2.6ex}0.015 &24.101 &\hspace{-2.7ex}$\pm$ &\hspace{-2.6ex}14.234 &43.854 &\hspace{-2.7ex}$\pm$ &\hspace{-2.6ex}7.523 \\
mip-NeRF &Goog. Scan. &31.727 &\hspace{-2.7ex}$\pm$ &\hspace{-2.6ex}1.681 &0.952 &\hspace{-2.7ex}$\pm$ &\hspace{-2.6ex}0.019 &0.032 &\hspace{-2.7ex}$\pm$ &\hspace{-2.6ex}0.010 &18.726 &\hspace{-2.7ex}$\pm$ &\hspace{-2.6ex}9.791 &36.932 &\hspace{-2.7ex}$\pm$ &\hspace{-2.6ex}7.464 \\
Ins.-NGP &Goog. Scan. &\textbf{32.772} &\hspace{-2.7ex}$\pm$ &\hspace{-2.6ex}1.985 &\textbf{0.963} &\hspace{-2.7ex}$\pm$ &\hspace{-2.6ex}0.012 &\textbf{0.017} &\hspace{-2.7ex}$\pm$ &\hspace{-2.6ex}0.007 &\textbf{1.920} &\hspace{-2.7ex}$\pm$ &\hspace{-2.6ex}1.542 &\textbf{7.515} &\hspace{-2.7ex}$\pm$ &\hspace{-2.6ex}2.784 \\
SVLF &Goog. Scan. &30.868 &\hspace{-2.7ex}$\pm$ &\hspace{-2.6ex}1.941 &0.954 &\hspace{-2.7ex}$\pm$ &\hspace{-2.6ex}0.020 &0.019 &\hspace{-2.7ex}$\pm$ &\hspace{-2.6ex}0.009 &9.200 &\hspace{-2.7ex}$\pm$ &\hspace{-2.6ex}4.341 &16.844 &\hspace{-2.7ex}$\pm$ &\hspace{-2.6ex}12.567 \\
\hline
NeRF &ABC &29.956 &\hspace{-2.7ex}$\pm$ &\hspace{-2.6ex}4.267 &0.947 &\hspace{-2.7ex}$\pm$ &\hspace{-2.6ex}0.042 &0.041 &\hspace{-2.7ex}$\pm$ &\hspace{-2.6ex}0.043 &66.312 &\hspace{-2.7ex}$\pm$ &\hspace{-2.6ex}146.052 &167.650 &\hspace{-2.7ex}$\pm$ &\hspace{-2.6ex}330.674 \\
mip-NeRF &ABC &32.191 &\hspace{-2.7ex}$\pm$ &\hspace{-2.6ex}3.328 &0.956 &\hspace{-2.7ex}$\pm$ &\hspace{-2.6ex}0.037 &\textbf{0.024} &\hspace{-2.7ex}$\pm$ &\hspace{-2.6ex}0.027 &22.497 &\hspace{-2.7ex}$\pm$ &\hspace{-2.6ex}11.307 &41.285 &\hspace{-2.7ex}$\pm$ &\hspace{-2.6ex}10.008 \\
Ins.-NGP &ABC &\textbf{34.250} &\hspace{-2.7ex}$\pm$ &\hspace{-2.6ex}4.564 &\textbf{0.971} &\hspace{-2.7ex}$\pm$ &\hspace{-2.6ex}0.027 &0.025 &\hspace{-2.7ex}$\pm$ &\hspace{-2.6ex}0.024 &\textbf{2.121} &\hspace{-2.7ex}$\pm$ &\hspace{-2.6ex}2.798 &\textbf{14.230} &\hspace{-2.7ex}$\pm$ &\hspace{-2.6ex}5.016 \\
SVLF &ABC &29.998 &\hspace{-2.7ex}$\pm$ &\hspace{-2.6ex}3.035 &0.949 &\hspace{-2.7ex}$\pm$ &\hspace{-2.6ex}0.044 &0.029 &\hspace{-2.7ex}$\pm$ &\hspace{-2.6ex}0.032 &10.238 &\hspace{-2.7ex}$\pm$ &\hspace{-2.6ex}7.008 &18.816 &\hspace{-2.7ex}$\pm$ &\hspace{-2.6ex}11.844 \\
\hline
NeRF &Bricks &28.279 &\hspace{-2.7ex}$\pm$ &\hspace{-2.6ex}3.363 &0.940 &\hspace{-2.7ex}$\pm$ &\hspace{-2.6ex}0.036 &0.041 &\hspace{-2.7ex}$\pm$ &\hspace{-2.6ex}0.023 &60.762 &\hspace{-2.7ex}$\pm$ &\hspace{-2.6ex}43.947 &34.122 &\hspace{-2.7ex}$\pm$ &\hspace{-2.6ex}30.115 \\
mip-NeRF &Bricks &31.329 &\hspace{-2.7ex}$\pm$ &\hspace{-2.6ex}3.361 &0.956 &\hspace{-2.7ex}$\pm$ &\hspace{-2.6ex}0.030 &0.023 &\hspace{-2.7ex}$\pm$ &\hspace{-2.6ex}0.014 &40.941 &\hspace{-2.7ex}$\pm$ &\hspace{-2.6ex}29.233 &23.717 &\hspace{-2.7ex}$\pm$ &\hspace{-2.6ex}6.869 \\
Ins.-NGP &Bricks &\textbf{32.255} &\hspace{-2.7ex}$\pm$ &\hspace{-2.6ex}3.277 &\textbf{0.970} &\hspace{-2.7ex}$\pm$ &\hspace{-2.6ex}0.019 &0.023 &\hspace{-2.7ex}$\pm$ &\hspace{-2.6ex}0.013 &\textbf{4.491} &\hspace{-2.7ex}$\pm$ &\hspace{-2.6ex}3.299 &\textbf{7.977} &\hspace{-2.7ex}$\pm$ &\hspace{-2.6ex}3.023 \\
SVLF &Bricks &29.244 &\hspace{-2.7ex}$\pm$ &\hspace{-2.6ex}3.265 &0.949 &\hspace{-2.7ex}$\pm$ &\hspace{-2.6ex}0.032 &\textbf{0.022} &\hspace{-2.7ex}$\pm$ &\hspace{-2.6ex}0.014 &20.300 &\hspace{-2.7ex}$\pm$ &\hspace{-2.6ex}13.938 &13.869 &\hspace{-2.7ex}$\pm$ &\hspace{-2.6ex}9.300 \\
\hline
NeRF &Amz. Ber. &25.778 &\hspace{-2.7ex}$\pm$ &\hspace{-2.6ex}3.704 &0.796 &\hspace{-2.7ex}$\pm$ &\hspace{-2.6ex}0.092 &0.184 &\hspace{-2.7ex}$\pm$ &\hspace{-2.6ex}0.108 &28.490 &\hspace{-2.7ex}$\pm$ &\hspace{-2.6ex}192.051 &74.041 &\hspace{-2.7ex}$\pm$ &\hspace{-2.6ex}21.488 \\
mip-NeRF &Amz. Ber. &\textbf{26.859} &\hspace{-2.7ex}$\pm$ &\hspace{-2.6ex}3.231 &0.784 &\hspace{-2.7ex}$\pm$ &\hspace{-2.6ex}0.088 &\textbf{0.141} &\hspace{-2.7ex}$\pm$ &\hspace{-2.6ex}0.077 &14.919 &\hspace{-2.7ex}$\pm$ &\hspace{-2.6ex}67.881 &66.027 &\hspace{-2.7ex}$\pm$ &\hspace{-2.6ex}17.342 \\
Ins.-NGP &Amz. Ber. &25.405 &\hspace{-2.7ex}$\pm$ &\hspace{-2.6ex}5.657 &\textbf{0.836} &\hspace{-2.7ex}$\pm$ &\hspace{-2.6ex}0.089 &0.161 &\hspace{-2.7ex}$\pm$ &\hspace{-2.6ex}0.103 &\textbf{12.693} &\hspace{-2.7ex}$\pm$ &\hspace{-2.6ex}112.870 &\textbf{36.202} &\hspace{-2.7ex}$\pm$ &\hspace{-2.6ex}33.896 \\
SVLF &Amz. Ber. &25.197 &\hspace{-2.7ex}$\pm$ &\hspace{-2.6ex}3.936 &0.795 &\hspace{-2.7ex}$\pm$ &\hspace{-2.6ex}0.103 &0.161 &\hspace{-2.7ex}$\pm$ &\hspace{-2.6ex}0.106 &15.305 &\hspace{-2.7ex}$\pm$ &\hspace{-2.6ex}49.771 &38.119 &\hspace{-2.7ex}$\pm$ &\hspace{-2.6ex}48.792 \\
\hline
NeRF \hspace{4ex}{\tiny (0.15 fps)} &all &28.363 &\hspace{-2.7ex}$\pm$ &\hspace{-2.6ex}3.819 &0.905 &\hspace{-2.7ex}$\pm$ &\hspace{-2.6ex}0.049 &0.078 &\hspace{-2.7ex}$\pm$ &\hspace{-2.6ex}0.046 &44.916 &\hspace{-2.7ex}$\pm$ &\hspace{-2.6ex}99.071 &79.917 &\hspace{-2.7ex}$\pm$ &\hspace{-2.6ex}97.450 \\
mip-NeRF \hspace{1ex}{\tiny (0.3 fps)} &all &30.526 &\hspace{-2.7ex}$\pm$ &\hspace{-2.6ex}2.900 &0.912 &\hspace{-2.7ex}$\pm$ &\hspace{-2.6ex}0.043 &\textbf{0.055} &\hspace{-2.7ex}$\pm$ &\hspace{-2.6ex}0.032 &24.271 &\hspace{-2.7ex}$\pm$ &\hspace{-2.6ex}29.090 &41.990 &\hspace{-2.7ex}$\pm$ &\hspace{-2.6ex}10.421 \\
Ins.-NGP \hspace{2ex}{\tiny (60 fps)}&all &\textbf{31.170} &\hspace{-2.7ex}$\pm$ &\hspace{-2.6ex}3.871 &\textbf{0.935} &\hspace{-2.7ex}$\pm$ &\hspace{-2.6ex}0.037 &0.056 &\hspace{-2.7ex}$\pm$ &\hspace{-2.6ex}0.037 &\textbf{5.306} &\hspace{-2.7ex}$\pm$ &\hspace{-2.6ex}30.127 &\textbf{16.481} &\hspace{-2.7ex}$\pm$ &\hspace{-2.6ex}11.180 \\
SVLF \hspace{3ex}{\tiny (12.10 fps)} &all &28.827 &\hspace{-2.7ex}$\pm$ &\hspace{-2.6ex}3.044 &0.912 &\hspace{-2.7ex}$\pm$ &\hspace{-2.6ex}0.050 &0.058 &\hspace{-2.7ex}$\pm$ &\hspace{-2.6ex}0.040 &13.761 &\hspace{-2.7ex}$\pm$ &\hspace{-2.6ex}18.764 &21.912 &\hspace{-2.7ex}$\pm$ &\hspace{-2.6ex}20.626 \\
\bottomrule
\end{tabular}
}
\end{footnotesize}
\vspace{-0.2cm}
\caption{Comparison of baseline methods on the 40-scene subset of our dataset at $400 \times 400$ resolution.
}
\label{table:results}
\vspace{-0.2cm}
\end{table*}

\begin{table*}
\centering
\begin{minipage}{.7\linewidth}
\begin{footnotesize}
\resizebox{0.98\linewidth}{!}{
\begin{tabular}{crclrclrclrcrrcr}
\toprule
\multirow{1}{*}{Env.} & \multicolumn{9}{c}{Image-based metrics} & \multicolumn{6}{c}{Depth-based metrics} \\
\cmidrule(lr){2-10}\cmidrule(lr){11-16}
& \multicolumn{3}{c}{PSNR $\uparrow$} & \multicolumn{3}{c}{SSIM $\uparrow$} & \multicolumn{3}{c}{LPIPS $\downarrow$} & \multicolumn{3}{c}{RMSE ($10^3$) $\downarrow$} & \multicolumn{3}{c}{MAE ($10^3$) $\downarrow$}\\
\midrule
Goog. Scan. &31.269 &\hspace{-2.7ex}$\pm$ &\hspace{-2.6ex}1.461 &1.486 &\hspace{-2.7ex}$\pm$ &\hspace{-2.6ex}0.014 &0.054 &\hspace{-2.7ex}$\pm$ &\hspace{-2.6ex}0.013 &0.634 &\hspace{-2.7ex}$\pm$ &\hspace{-2.6ex}0.591 &6.790 &\hspace{-2.7ex}$\pm$ &\hspace{-2.6ex}2.829 \\
ABC &33.888 &\hspace{-2.7ex}$\pm$ &\hspace{-2.6ex}3.842 &0.975 &\hspace{-2.7ex}$\pm$ &\hspace{-2.6ex}0.017 &0.034 &\hspace{-2.7ex}$\pm$ &\hspace{-2.6ex}0.024 &1.311 &\hspace{-2.7ex}$\pm$ &\hspace{-2.6ex}2.075 &13.913 &\hspace{-2.7ex}$\pm$ &\hspace{-2.6ex}5.104 \\
Bricks &31.170 &\hspace{-2.7ex}$\pm$ &\hspace{-2.6ex}3.281 &0.966 &\hspace{-2.7ex}$\pm$ &\hspace{-2.6ex}0.020 &0.045 &\hspace{-2.7ex}$\pm$ &\hspace{-2.6ex}0.029 &2.526 &\hspace{-2.7ex}$\pm$ &\hspace{-2.6ex}8.298 &7.863 &\hspace{-2.7ex}$\pm$ &\hspace{-2.6ex}8.112 \\
Amz. Ber. &23.129 &\hspace{-2.7ex}$\pm$ &\hspace{-2.6ex}4.882 &0.851 &\hspace{-2.7ex}$\pm$ &\hspace{-2.6ex}0.078 &0.254 &\hspace{-2.7ex}$\pm$ &\hspace{-2.6ex}0.097 &17.766 &\hspace{-2.7ex}$\pm$ &\hspace{-2.6ex}164.130 &47.861 &\hspace{-2.7ex}$\pm$ &\hspace{-2.6ex}53.498 \\
\hline
all &30.358 &\hspace{-2.7ex}$\pm$ &\hspace{-2.6ex}3.334 &1.030 &\hspace{-2.7ex}$\pm$ &\hspace{-2.6ex}0.027 &0.077 &\hspace{-2.7ex}$\pm$ &\hspace{-2.6ex}0.036 &4.412 &\hspace{-2.7ex}$\pm$ &\hspace{-2.6ex}30.355 &14.854 &\hspace{-2.7ex}$\pm$ &\hspace{-2.6ex}13.878 \\
\bottomrule
\end{tabular}
}
\end{footnotesize}
\vspace{-0.2cm}
\caption{Results for the full dataset at $1600 \times 1600$ res. with Instant-NGP \cite{mueller2022instant}.}
\label{table:results_full}
\end{minipage}
~\quad
\begin{minipage}{.25\linewidth}
\begin{footnotesize}
    \centering
    \vspace{1ex}  %
    \begin{tabular}{ccc}
         \toprule
         env. & PSNR $\uparrow$ & SSIM $\uparrow$ \\
         \midrule
         Goog. Scan. & 14.588 & 0.483 \\
         ABC & 12.149 &  0.629\\
         Bricks & 12.149 & 0.523\\
         Amz. Ber. & 12.126 & 0.318\\
         \midrule
         all & 12.753 & 0.488 \\
        \bottomrule
    \end{tabular}
\end{footnotesize}
\vspace{-0.2cm}
\caption{Pixel-NeRF results.}
\label{tab:pixelnerf}
\end{minipage}
\vspace{-0.4cm}
\end{table*}

\subsection{Multi-stage training}
\label{sec:svlf_training}
We train SVLF in three stages:
\begin{enumerate}[leftmargin=*]
    \item Voxel training:
    We begin our training using surface rendering.
    For each ray, we use the corresponding depth to find the voxel containing the surface hit by the ray.
    We learn local features for each voxel to produce the correct color $c$, within-voxel depth $\eta$ and optical thickness $\tau$ using the following loss:
    \begin{equation}
        \mathcal{L} = \sum_v \|c_v - c_{\text{gt}}\|^2 + \lambda_{\eta} \|\eta_v - \eta_{\text{gt}}\|^2 + \lambda_\tau \|e^{-\tau_v}\|^2
        \label{eqn:loss}
    \end{equation}
    where the last term encourages predicting a zero transparency for the ray segment through $v$.

    \item Volumetric training of optical thickness: Next, we freeze the color features $\zcolor$ and decoder $\fcolor$ (Eq.~\ref{eq:ceqfc}), and train the optical thickness network to predict the correct integral weights for volume rendering (Eq.~\ref{eqn:rendering}). We use the same loss function (Eq.~\ref{eqn:loss})
    except that
    we compute the photometric loss with the aggregated ray color, $c(\bfr)$, and
    we set $\lambda_\tau$ to zero to turn off direct supervision on the optical thickness $\tau$. We observe that freezing the color network is important for convergence, otherwise the network gets stuck in poor local minima.
    \item Fine-tuning: Finally, we unfreeze the color features $\zcolor$ and decoder $\fcolor$, and fine-tune the full model using the same loss function, but with a lowered learning rate.
\end{enumerate}

\subsection{Implementation details}
\label{sec:implementation_details}
\noindent \textbf{Architecture.}
We represent a scene using a sparse octree, constructed at $128^3$ voxel resolution.
The octree implementation comes from the Kaolin library~\cite{Kaolin}.
Specifically, we leverage the Structured Point Cloud (SPC) representation which supports efficient CUDA kernels for ray tracing.
We have additionally implemented CUDA kernels for volumetric rendering using efficient scan primitives from CUB~\cite{merrill2017cub},
which we leverage for fast and efficient rendering of our SVLF model.
Both the optical thickness and color decoders $\fdensity$, $\fcolor$ are ReLU MLP networks.
We use a single hidden layer for $\fdensity$ and 3 hidden layers for $\fcolor$, all with size 128.
We apply a ReLU activation to the output optical thickness $\tau$, and a sigmoid activation to both the mixing ratio $\eta$ and color $c$ outputs.
We set the embedding size of the feature collections $\zdensity$ and $\zcolor$ to 64 and 32, respectively.

\medskip \noindent \textbf{Training.}
We run our staged training (Sec.~\ref{sec:svlf_training}) for 300 epochs, which takes $\sim$160 minutes on a single NVIDIA V100 GPU.
We use the Adam optimizer~\cite{adamOptimizer} with a learning rate of 1e-3 and a batch size of a single image down-scaled to a $400 \times 400$ resolution.
We train using surface rendering for 100 epochs.
We then freeze the weights of the color features $\zcolor$ and decoder $\fcolor$ and run the volume\-tric training stage for 150 epochs.
Finally, we unfreeze the color weights ($\zcolor, \fcolor$), and continue volumetric training for another 50 epochs with a learning rate of 2e-4.

\section{Experiments}
\label{sec:experiments}

In this section, we benchmark different baselines against our dataset on two main tasks: single scene view synthesis and few-shot view synthesis.
We evaluate the view reconstruction quality of baselines using PSNR and SSIM (higher is better) and LPIPS~\cite{zhang2018perceptual} (lower is better).
We also evaluate the reconstructed depth maps using RMSE and MAE (lower is better).

\subsection{Single scene view synthesis}

In our dataset, each scene is composed of 150 unique views.
We propose to use 100 views for training,
5 views for validation, and 45 views for testing.
For this task we evaluate four baselines:
NeRF~\cite{mildenhall2020nerf}, mip-NeRF~\cite{barron2021mip}, the concurrent work of Instant-NGP~\cite{mueller2022instant} and our proposed SVLF.
We use the author-provided implementations for NeRF\footnote{\url{https://github.com/bmild/nerf}}, mip-NeRF\footnote{\url{https://github.com/google/mipnerf}}, and Instant-NGP\footnote{\url{https://github.com/NVlabs/instant-ngp}} and train all methods until convergence.
The number of training steps was determined by the saturation of the mean PSNR value over the validation set for a few scenes and then was fixed for all scenes.
As NeRF and mip-NeRF have quite large training time, we propose a smaller subset composed of 10 scenes per environment,
for a total of 40 scenes.
Finally, we mask out the background in the Amazon Berkeley environment as the baselines are not designed to handle complex backgrounds.

Table~\ref{table:results} presents high-level results for image and depth-based metrics, and
Figure~\ref{fig:qual_comp} presents qualitative comparisons.
All methods perform reasonably well on the Google Scanned Objects and the Bricks environments
which have a setup similar to
the NeRF Synthetic dataset~\cite{mildenhall2020nerf}.
NeRF and SVLF performance degrade on the ABC environment, and
none of the baselines performs well on our most challenging environment, Amazon Berkeley,
with Instant-NGP being the most robust of the baselines, followed by mip-NeRF as it is trained on multi-scale images.
SVLF benefits from depth supervision during training and therefore constructs better depth maps than NeRF and mip-NeRF.
In our experiments we have observed that SVLF does not always capture view dependent effects due to a lack of global context in our voxel-based formulation and the absence of an explicit view direction input.

Finally, we show results using Instant-NGP for all 1927 scenes of the dataset at full resolution in Table~\ref{table:results_full}.
Please refer to the supp.~material for more qualitative results and depth map visualizations.

\begin{table}
\centering
\begin{footnotesize}
\resizebox{0.92\linewidth}{!}{
\begin{tabular}{ccccc}
\toprule
\multirow{1}{*}{Method} & \multicolumn{2}{c}{Training} & \multicolumn{2}{c}{Rendering} \\
\cmidrule(lr){2-3}\cmidrule(lr){4-5}
& \# steps & time (hrs) & time/frame (sec) & FPS \\
\midrule
NeRF & 500K  & 26.3 & 6.86 & 0.15  \\
mip-NeRF & 500K & 35.0 & 3.30 & 0.30 \\
SVLF & 30K & 2.7 & 0.08 & 12.10\\
\bottomrule
\end{tabular}
}
\end{footnotesize}
\vspace{-0.2cm}
\caption{Training and inference time comparison on the single-scene novel view synthesis task.}
\label{table:runtime}
\vspace{-0.4cm}
\end{table}

\medskip
\noindent \textbf{Runtime comparison.}
We compare the training and rendering speed of SVLF to the baselines in Table~\ref{table:runtime}.
Training is done on a single V100 GPU, while inference is done on a single RTX 3090 GPU.
SVLF trains an order of magnitude faster than the other baselines as it utilizes depth maps during training.
This observation is also consistent with \cite{neff2021donerf}, a depth-supervised NeRF.

To compare inference time, we compute the average rendering time over the 45 test images of the first scene of the Google Scanned Objects environment.
SVLF is over 80x faster to render than NeRF, and approximately 40x faster than mip-NeRF on our test scene.
There are two main reasons for this speedup.
First, rendering SVLF requires significantly fewer network queries per ray compared to NeRF.
The number of queries is determined by the number of voxels intersected along the ray, which is efficiently computed using the ray-octree intersection algorithm of~\cite{takikawa2021nglod}.
On average, we have less than~$10$ voxel evaluations per ray on our scenes.
Second, the cost per query is much cheaper for SVLF.
This is due to moving much of the network capacity to the learned local features, and thus enabling using a much smaller decoder architecture ($\sim$16x smaller) than the NeRF network.

\subsection{Few-shot view synthesis}

\begin{figure}
     \centering
    \begin{footnotesize}
    \resizebox{0.96\linewidth}{!}{
     \begin{tabular}{cc|c|c}
    \multicolumn{2}{c}{input views} & GT & PixelNeRF~\cite{yu2020pixelnerf} \\
        \rotatebox{90}{\footnotesize Goog. Scan.} &
	     \includegraphics[height=1.2cm]{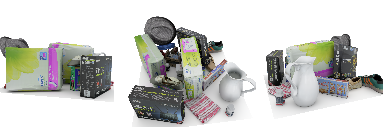}
        &
         \includegraphics[height=1.2cm]{figures/pixelnerf/goog/1134_0000_vis_crop.png}
         &
         \includegraphics[height=1.2cm]{figures/pixelnerf/goog/1134_0000_vis_crop.png}
         \\
          \rotatebox{90}{\footnotesize \ \ \ \ \ \ Bricks} &
	     \includegraphics[height=1.2cm]{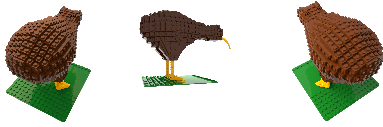}
        &
         \includegraphics[height=1.2cm]{figures/pixelnerf/lego/0746_0200_vis_crop.png}
         &
         \includegraphics[height=1.2cm]{figures/pixelnerf/lego/0746_0200_vis_crop.png}
         \\
                     \rotatebox{90}{\footnotesize \ \ \ \ \ \ ABC} &
         \includegraphics[height=1.2cm]{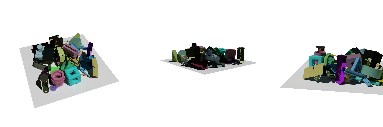}
         &
         \includegraphics[height=1.2cm]{figures/pixelnerf/abc/1181_0000_vis_crop.png}
         &
         \includegraphics[height=1.2cm]{figures/pixelnerf/abc/1181_0000_vis_crop.png}
         \\
         \rotatebox{90}{\footnotesize Amz. Ber.} &
	     \includegraphics[height=1.2cm]{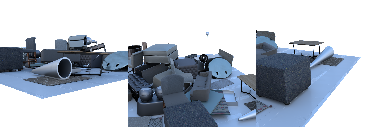}
        &
         \includegraphics[height=1.2cm]{figures/pixelnerf/amz/0144_0000_vis_crop.png}
         &
         \includegraphics[height=1.2cm]{figures/pixelnerf/amz/0144_0000_vis_crop.png}
     \end{tabular}
     }
    \end{footnotesize}
    \vspace{-0.3cm}
    \caption{Sample results of PixelNeRF~\cite{yu2020pixelnerf} on our dataset. }
    \vspace{-0.5cm}
    \label{fig:pixelnerf}
\end{figure}

Given that each environment includes multiple scenes with shared characteristics, camera distributions, object types,
and/or lighting, we can use any environment to train few-shot view synthesis algorithms.
Different methods \cite{sitzmann2019srns,yu2020pixelnerf} have shown that a single or few views ($<5$)
can be used to predict new views.
We propose to partition our dataset as follows: 280 scenes for training, 2 scenes for validation,
and 18 scenes for testing.
Since the Bricks environment is bigger, we propose to use 900 scenes for training, 30 for validation, and 100 for testing.
We use the freely available training code\footnote{\url{https://github.com/sxyu/pixel-nerf}} for PixelNeRF~\cite{yu2020pixelnerf}.
We found the algorithm unstable to train,
especially when having random camera poses, and eventually diverging after a few hundred epochs.
We terminate the training when the model diverges and use the latest proper checkpoint for evaluation.
We show qualitative results for a 3-shot input experiment in Fig.~\ref{fig:pixelnerf} and report quantitative metrics in Table~\ref{tab:pixelnerf}.
Even though the results are encouraging, it is clear that future work is needed
in the area of few-shot view synthesis.

\section{Conclusion}
In this work, we proposed RTMV, a large-scale, high quality and ray-traced synthetic dataset for novel view synthesis.
Our dataset is orders of magnitude larger than currently used datasets and offers a challenging variety in terms of camera poses, lighting conditions, object material and textures.
Thus, RTMV is suitable as a demanding benchmark to evaluate view synthesis algorithms that will help advance novel view synthesis research.
We also proposed SVLF, a voxel-based accelerated algorithm for novel view synthesis.
SVLF is an order of magnitude faster to train on synthetic images, and two orders of magnitude faster to render than NeRF, while retaining output quality.
Finally, we propose a Python-based ray tracing renderer that is easy to use by non-experts and suitable for fast scripting and rendering high-quality physically-based images.

\vspace{-0.2cm}
\begin{small}
  \paragraph{Acknowledgments.}
  We thank Chen-Hsuan Lin for his valuable feedback. MM was partially funded by DARPA SemaFor (HR001119S0085).
\end{small}

\begin{figure*}[hbt!]
    \centering
    \captionsetup{type=figure}
    \includegraphics[width=1\textwidth]{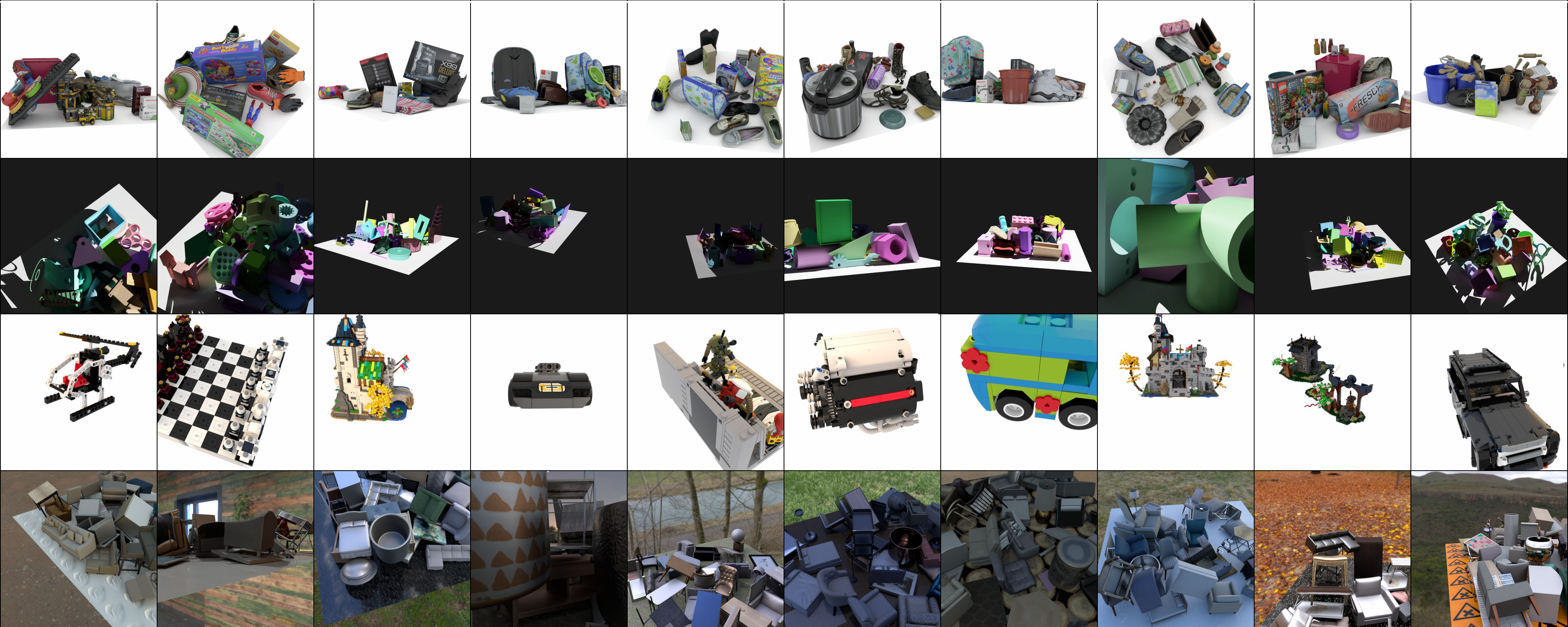} \\
    \label{tab:dataset}
    \caption{Examples of multiple views from different environments:
      Google Scanned ($1^\mathit{st}$ row),
      ABC ($2^\mathit{nd}$ row),
      Bricks ($3^\mathit{rd}$ row), and
      Amazon-Berkeley ($4^\mathit{th}$ row).
      (Best viewed when zoomed.)
    }
    \vspace{-0.1cm}
    \label{fig:dataset}
    \vspace{-0.2cm}
\end{figure*}%

\appendix

\input{appendix.tex}

\clearpage
\clearpage

{\small
\bibliographystyle{ieee_fullname}
\bibliography{main}
}

\end{document}

%% file: appendix.tex
\begin{figure}
     \centering
     \begin{tabular}{ccc}
     \rotatebox[origin=c]{90}{\footnotesize Goog. scan} &
	     \includegraphics[align=c,width=0.4\columnwidth]{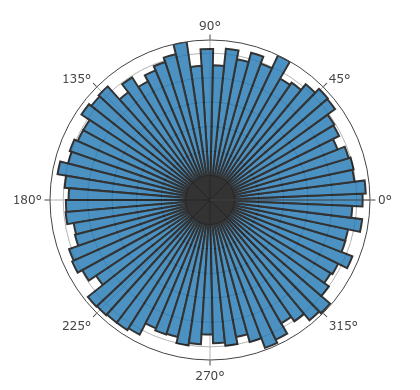} &
	     \includegraphics[align=c,width=0.4\columnwidth]{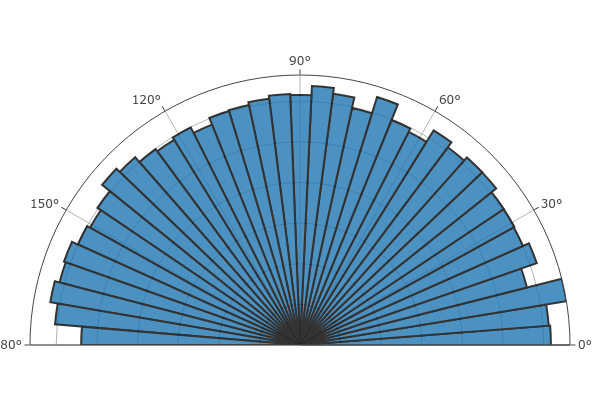}
     \end{tabular}
     \centering
     \begin{tabular}{ccc}
     \rotatebox[origin=c]{90}{\footnotesize ABC} &
	     \includegraphics[align=c,width=0.4\columnwidth]{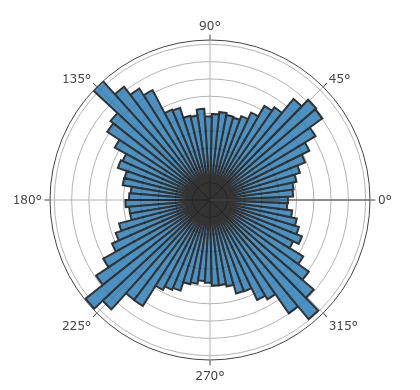} &
	     \includegraphics[align=c,width=0.4\columnwidth]{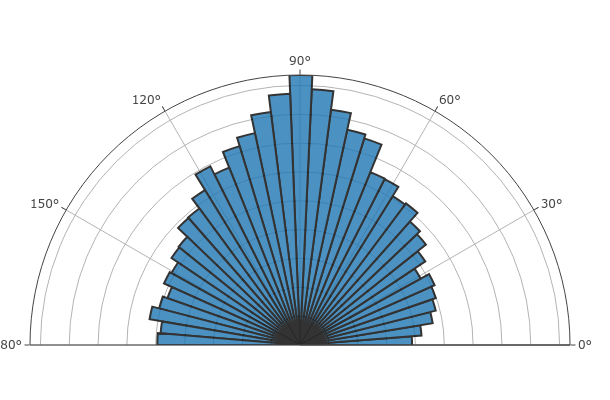}
     \end{tabular}
     \centering
     \begin{tabular}{ccc}
     \rotatebox[origin=c]{90}{\footnotesize Bricks} &
	     \includegraphics[align=c,width=0.4\columnwidth]{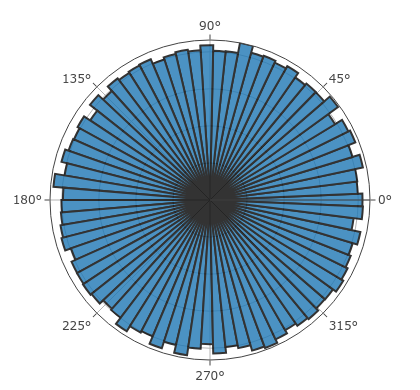} &
	     \includegraphics[align=c,width=0.4\columnwidth]{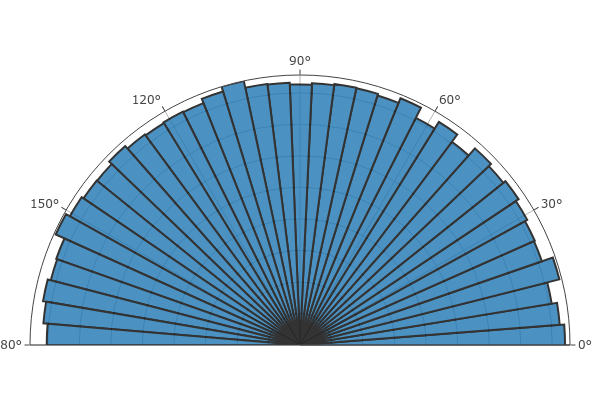}
     \end{tabular}
     \centering
     \begin{tabular}{ccc}
         \rotatebox[origin=c]{90}{\footnotesize Amz. Ber.} &
	     \includegraphics[align=c,clip,width=0.4\columnwidth]{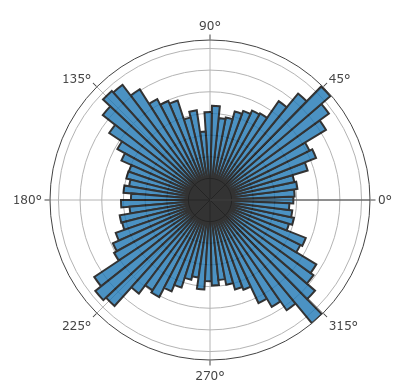} &
	     \includegraphics[align=c,clip,width=0.4\columnwidth]{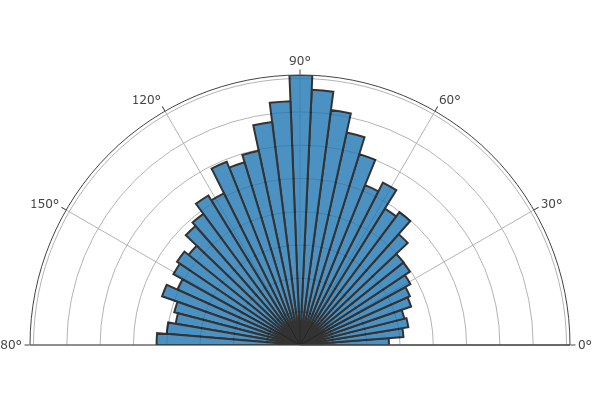}
     \end{tabular}

\vspace{-0.2cm}
\caption{Camera azimuth and elevation distributions for our different environments}
\label{fig:camera_dist}
\vspace{-0.4cm}
\end{figure}
\section{Appendix}
\subsection{Data generation}

\paragraph{Camera pose distribution.}
Our RTMV dataset is designed to be more challenging than existing novel view synthesis datasets.
In addition to its diversity in terms of scene lighting and objects materials and textures (see Figure~\ref{fig:dataset}),
it contains two types of camera placements: hemisphere and free.
The `hemisphere' camera placement is an easy setup where cameras are placed at an equal distance from the scene (\ie, on a hemisphere).
Therefore, trained multi view synthesis models do not need to reason about multi-scale.
In the `free' camera placement, the camera is allowed to roam freely within a cube surrounding the scene,
thus providing a variety of multi-scale images of the scene.
Figure~\ref{fig:camera_dist} shows the distribution of camera azimuth and elevation for each of our four environments.

\begin{figure}
\begin{lstlisting}[language=Python]
  import raytracer 
  raytracer.initialize()
  # Create camera
  my_camera = raytracer.entity.create(
      name      = 'cam',
      transform = raytracer.transform.create('c_tfm'),
      camera    = raytracer.camera.create('c_cam')
  )
  my_camera.get_transform().look_at(
      eye = [3, 3, 3], at = [0, 0, 0], up = [0, 0, 1]
  )
  raytracer.set_camera_entity(my_camera)
  # Create object
  my_object = raytracer.entity.create(
      name      = 'obj', 
      transform = raytracer.transform.create('o_tfm'),
      mesh      = raytracer.mesh.create_sphere('o_mesh'),
      material  = raytracer.material.create('o_mat')
  )
  raytracer.material.get('o_mat').set_base_color([1, 0, 0])
  # Render image
  raytracer.render_to_file(
      width = 512, height = 512, 
      samples_per_pixel = 1024, 
      file_path = 'image.png'
  )
  raytracer.deinitialize()
\end{lstlisting}
\vspace*{-5.5em}
\raggedleft\includegraphics[width=1.5cm]{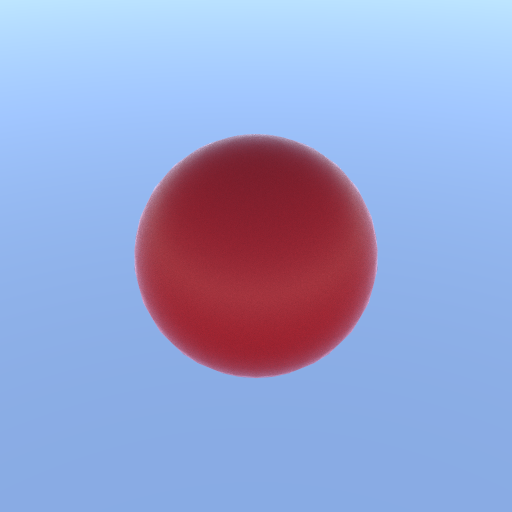}\hspace{.1em}
\caption{A minimal Python script that renders the inset image.}
\label{fig:nvisii}
\end{figure}

\paragraph{Python-based ray tracer.}
Our Python-based ray-tracing renderer is built on NVIDIA OptiX~\cite{optix} with a C++/CUDA backend.
It uses path tracing to render high-quality and physically-based images.
In addition, it appeals to non-graphics experts through its simple Python API.
The scripting capability makes it easy to install, use and share.
Figure~\ref{fig:nvisii} shows a simple code example highlighting the simplicity of our Python API.

\subsection{Additional qualitative results}
\begin{figure*}
\centering
\begin{tabular}{cc|c|c|c}
\rotatebox[origin=lt]{90}{\small \ \ \ \ \ \ \ \ \ Ground truth} &
\includegraphics[trim=0 0 0 -5, width=0.40\columnwidth]{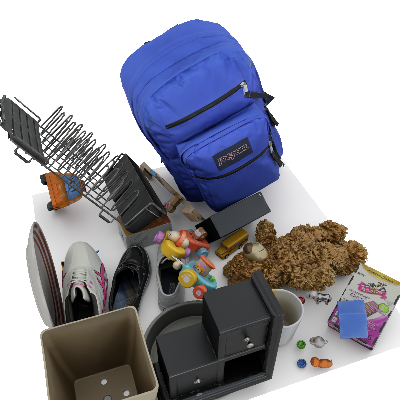}
&

\includegraphics[trim=0 0 0 -5, width=0.40\columnwidth]{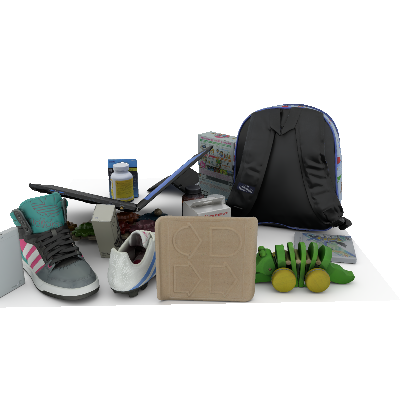}
&

\includegraphics[trim=0 0 0 -5, width=0.40\columnwidth]{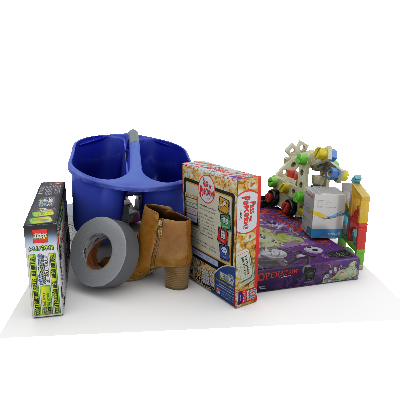}
&

\includegraphics[trim=0 0 0 -5, width=0.40\columnwidth]{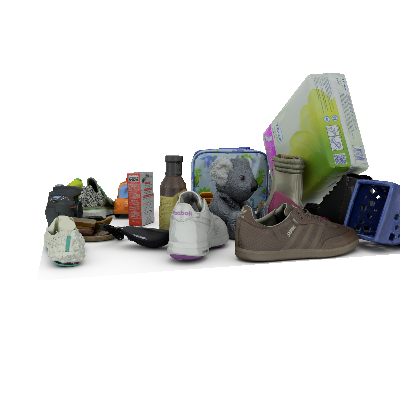}
\\

\rotatebox[origin=lt]{90}{\small \ \ \ \ \ \ \ \ \ \ \ \ \ \ SVLF} &
\includegraphics[trim=0 0 0 -5, width=0.4\columnwidth]{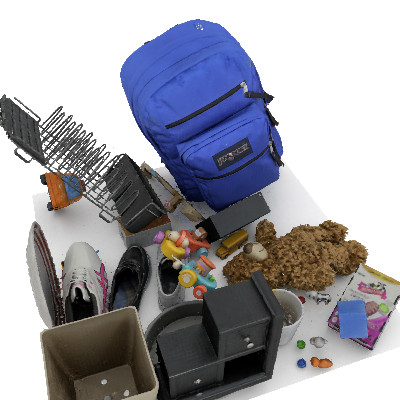}
&

\includegraphics[trim=0 0 0 -5, width=0.4\columnwidth]{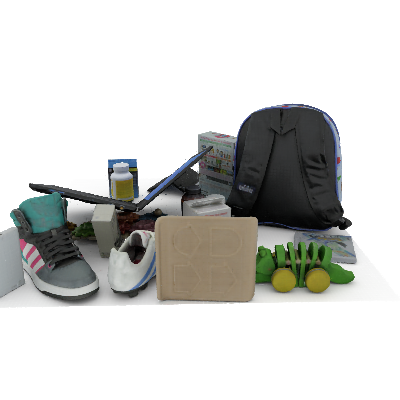}
&

\includegraphics[trim=0 0 0 -5, width=0.4\columnwidth]{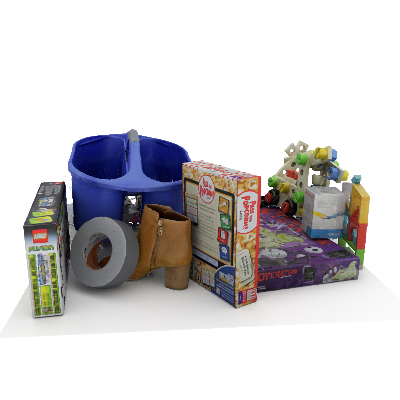}
&

\includegraphics[trim=0 0 0 -5, width=0.4\columnwidth]{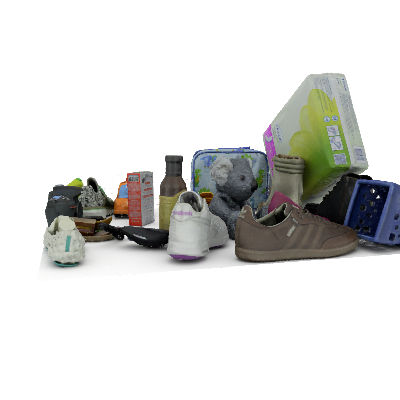}
\\

\rotatebox[origin=lt]{90}{\small \ \ \ \ \ \ \ \ \ \ Ins.-NGP} &
\includegraphics[trim=0 0 0 -5, width=0.4\columnwidth]{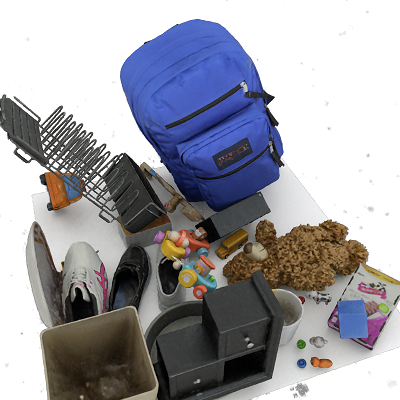}
&

\includegraphics[trim=0 0 0 -5, width=0.4\columnwidth]{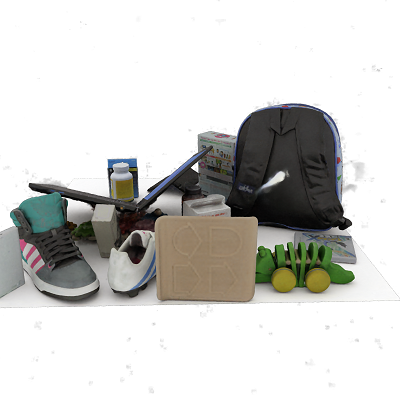}
&

\includegraphics[trim=0 0 0 -5, width=0.4\columnwidth]{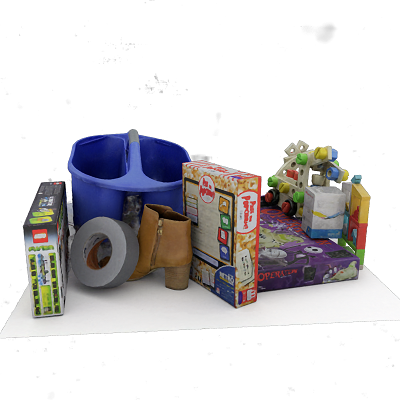}
&

\includegraphics[trim=0 0 0 -5, width=0.4\columnwidth]{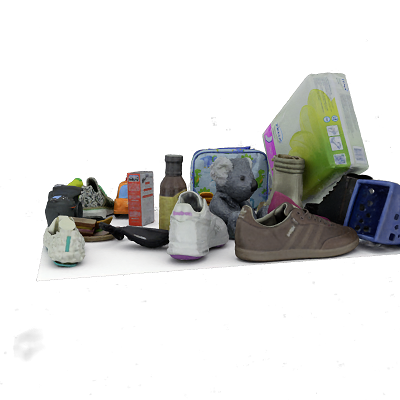}
\\

\rotatebox[origin=lt]{90}{\small \ \ \ \ \ \ \ \ \ mip-NeRF} &
\includegraphics[trim=0 0 0 -5, width=0.4\columnwidth]{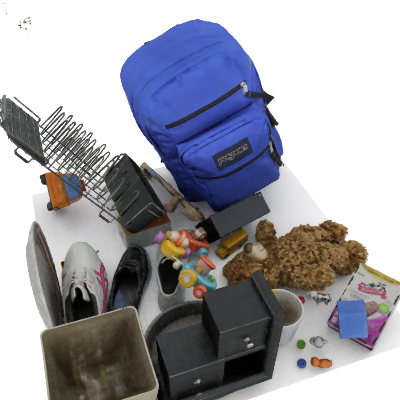}
&

\includegraphics[trim=0 0 0 -5, width=0.4\columnwidth]{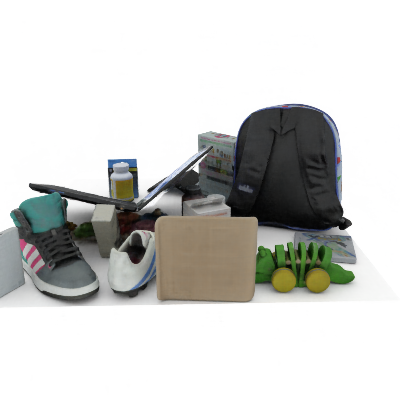}
&

\includegraphics[trim=0 0 0 -5, width=0.4\columnwidth]{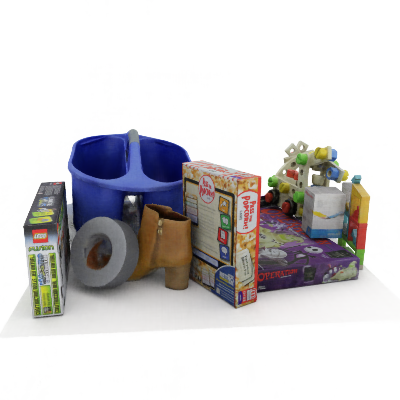}
&

\includegraphics[trim=0 0 0 -5, width=0.4\columnwidth]{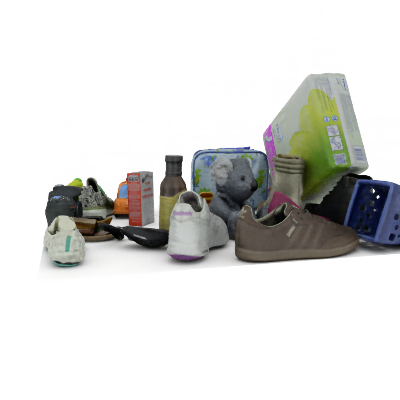}
\\

\rotatebox[origin=lt]{90}{\small \ \ \ \ \ \ \ \ \ \ \ \ \ \ NeRF} &
\includegraphics[trim=0 0 0 -5, width=0.4\columnwidth]{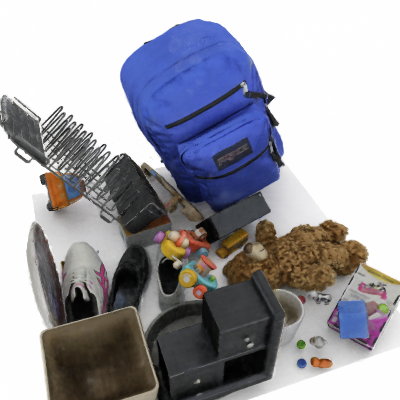}
&

\includegraphics[trim=0 0 0 -5, width=0.4\columnwidth]{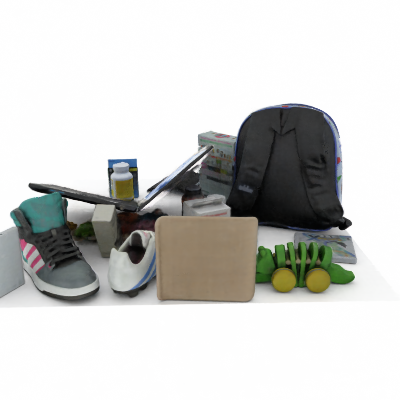}
&

\includegraphics[trim=0 0 0 -5, width=0.4\columnwidth]{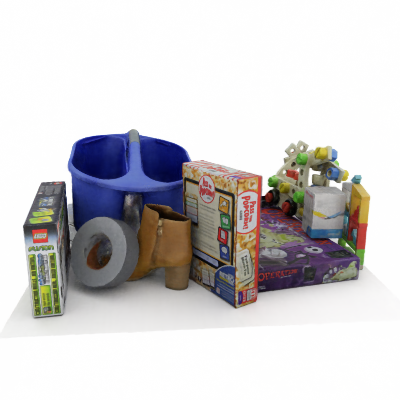}
&

\includegraphics[trim=0 0 0 -5, width=0.4\columnwidth]{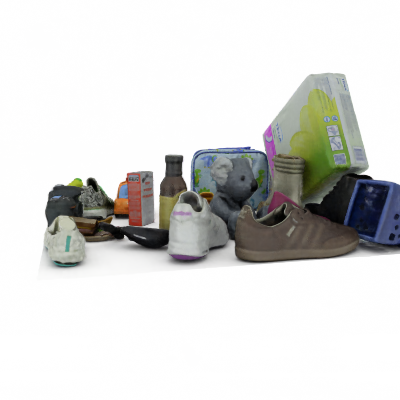}
\\

\end{tabular}
\caption{More qualitative results on the Google Scanned Objects environment (figure best seen in zoom).}
\label{fig:more_qual_results_1}
\end{figure*}

\begin{figure*}
\centering
\begin{tabular}{cc|c|c|c}
\rotatebox[origin=lt]{90}{\small \ \ \ \ \ \ \ \ \ Ground truth} &
\includegraphics[trim=0 0 0 -5, width=0.4\columnwidth]{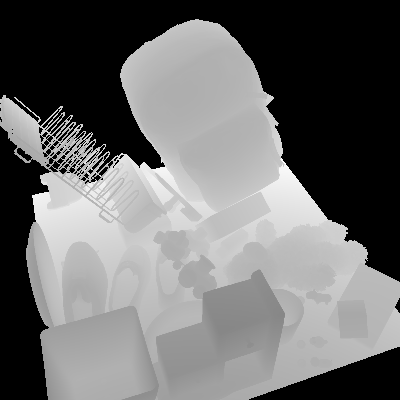}
&

\includegraphics[trim=0 0 0 -5, width=0.4\columnwidth]{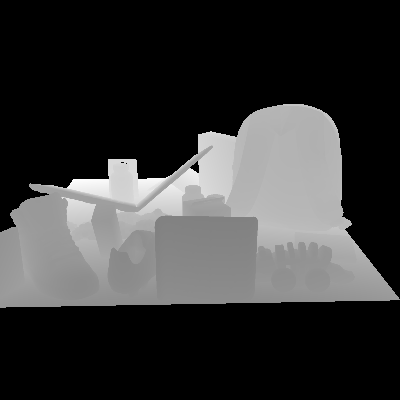}
&

\includegraphics[trim=0 0 0 -5, width=0.4\columnwidth]{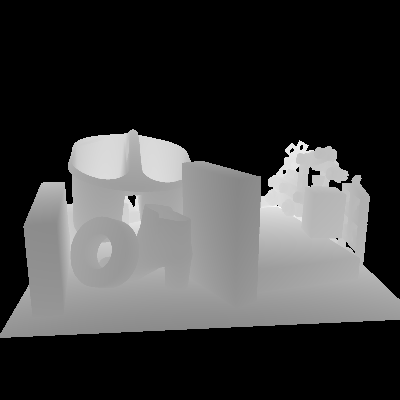}
&

\includegraphics[trim=0 0 0 -5, width=0.4\columnwidth]{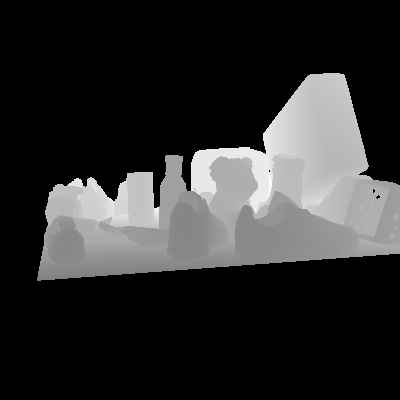}
\\

\rotatebox[origin=lt]{90}{\small \ \ \ \ \ \ \ \ \ \ \ \ \ \ SVLF} &
\includegraphics[trim=0 0 0 -5, width=0.4\columnwidth]{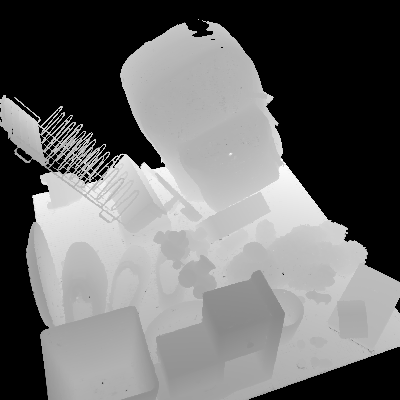}
&

\includegraphics[trim=0 0 0 -5, width=0.4\columnwidth]{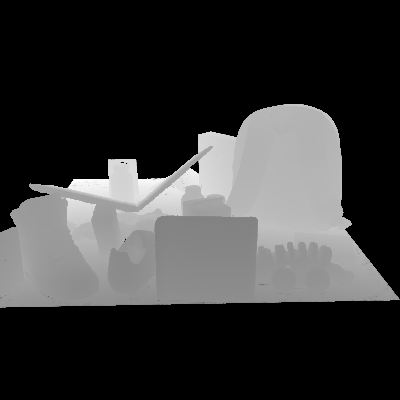}
&

\includegraphics[trim=0 0 0 -5, width=0.4\columnwidth]{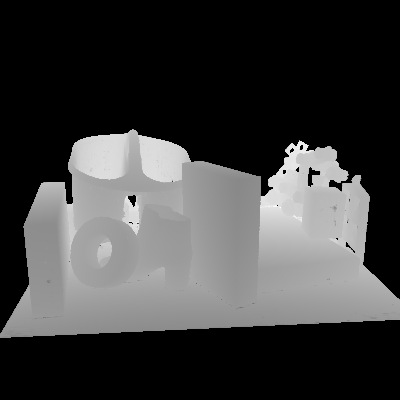}
&

\includegraphics[trim=0 0 0 -5, width=0.4\columnwidth]{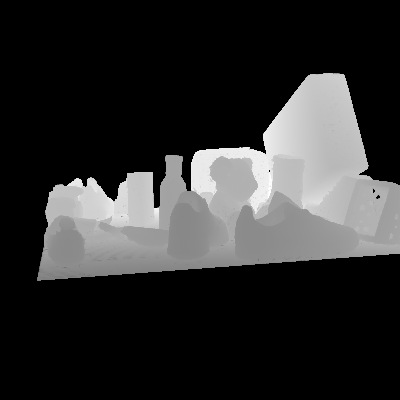}
\\

\rotatebox[origin=lt]{90}{\small \ \ \ \ \ \ \ \ \ \ Ins.-NGP} &
\includegraphics[trim=0 0 0 -5, width=0.4\columnwidth]{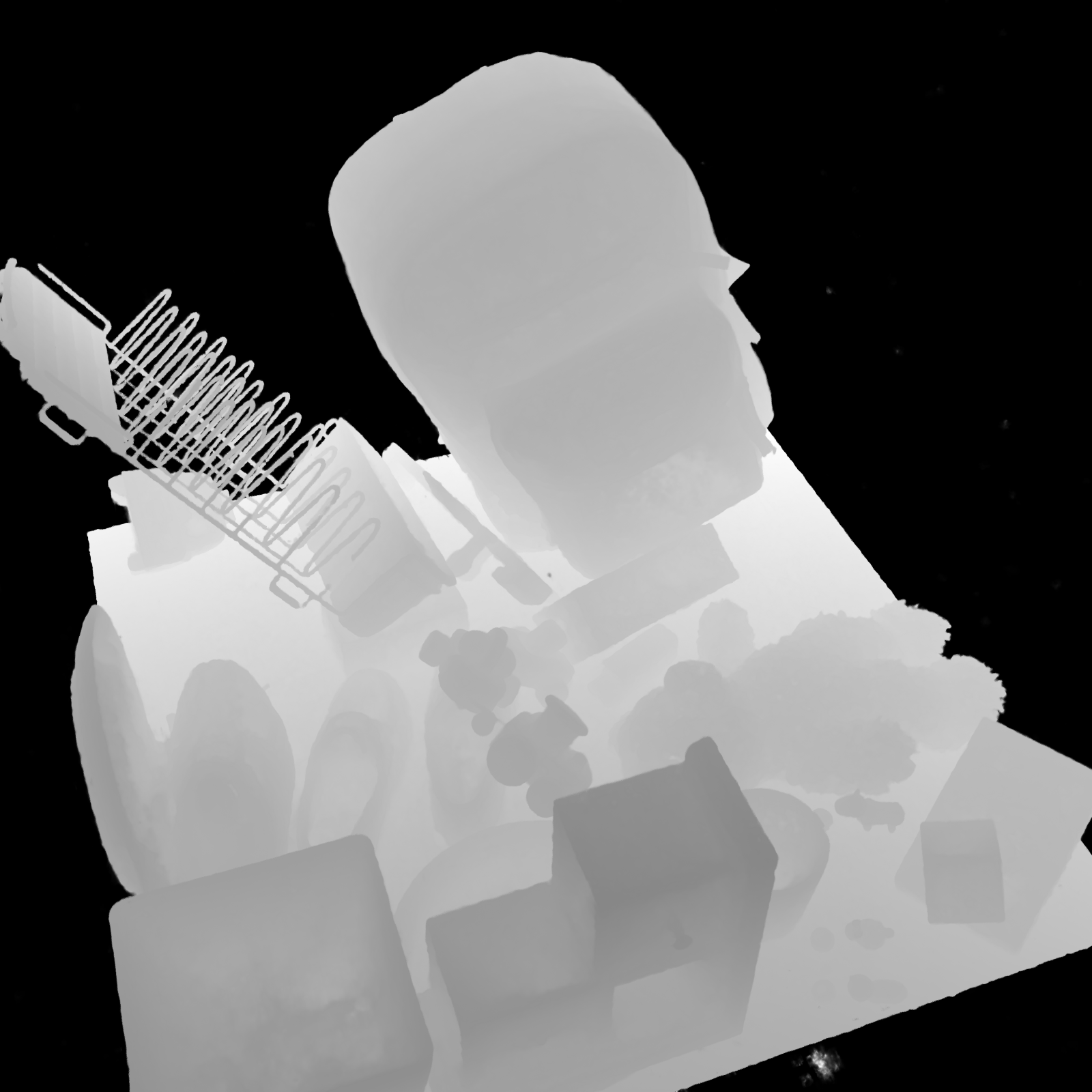}
&

\includegraphics[trim=0 0 0 -5, width=0.4\columnwidth]{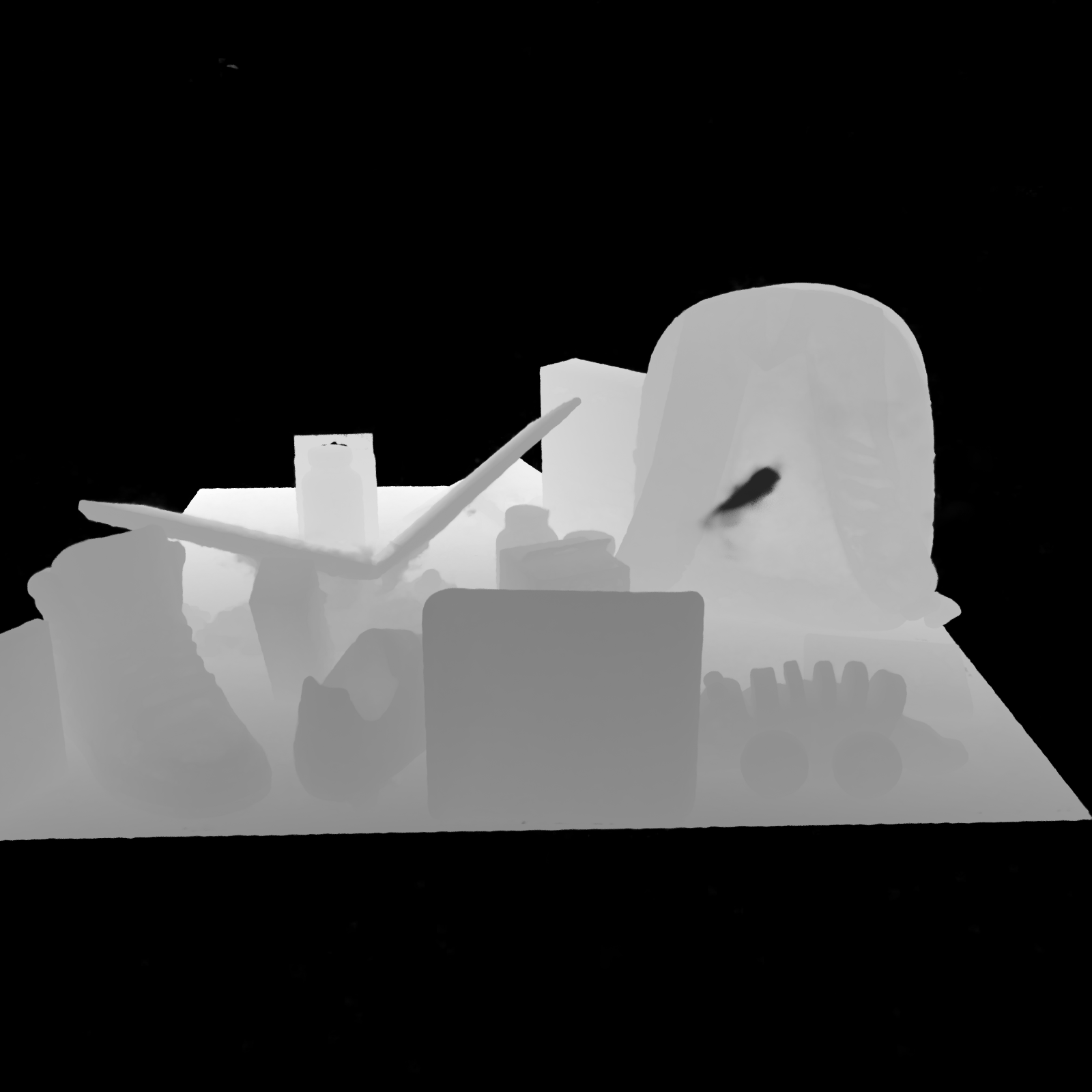}
&

\includegraphics[trim=0 0 0 -5, width=0.4\columnwidth]{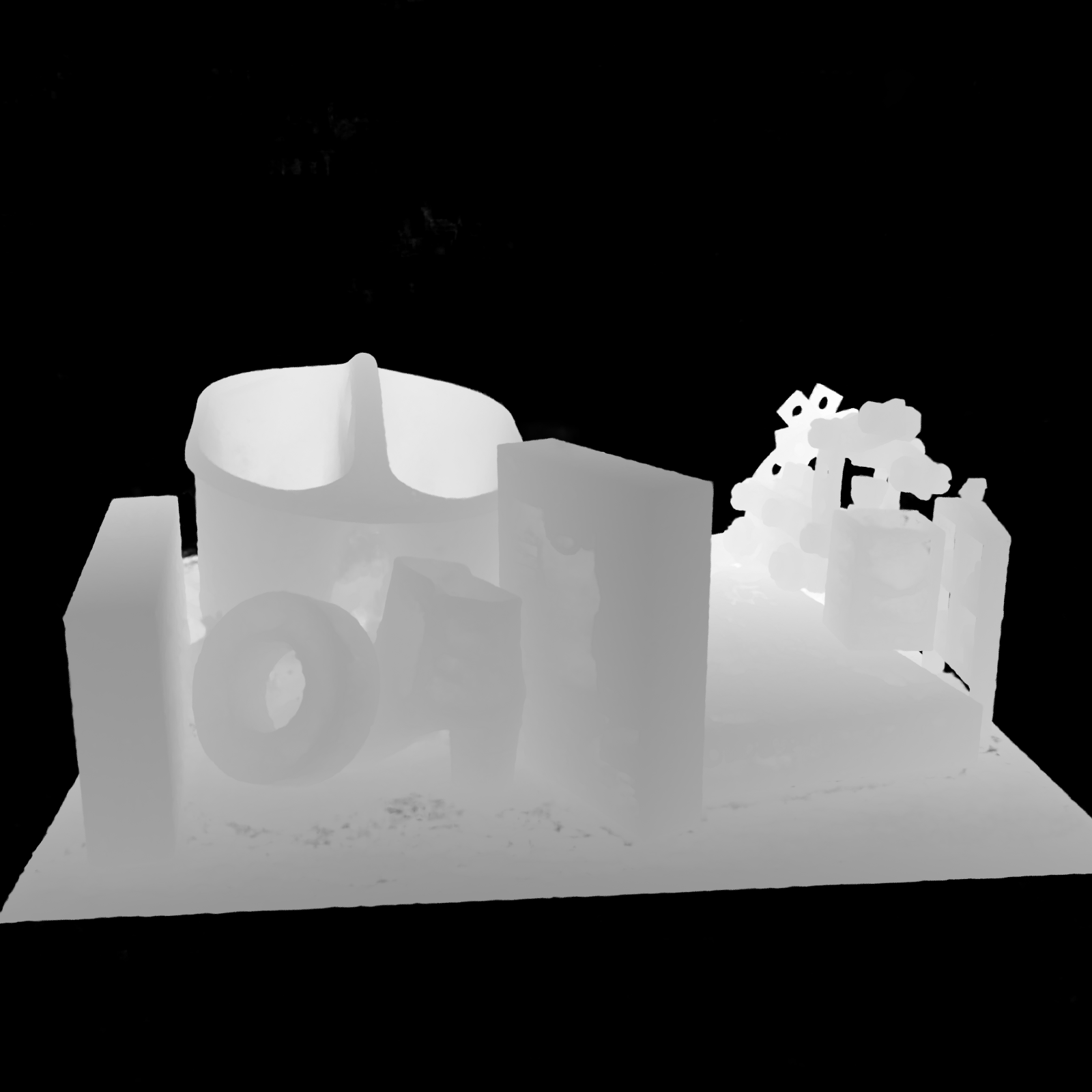}
&

\includegraphics[trim=0 0 0 -5, width=0.4\columnwidth]{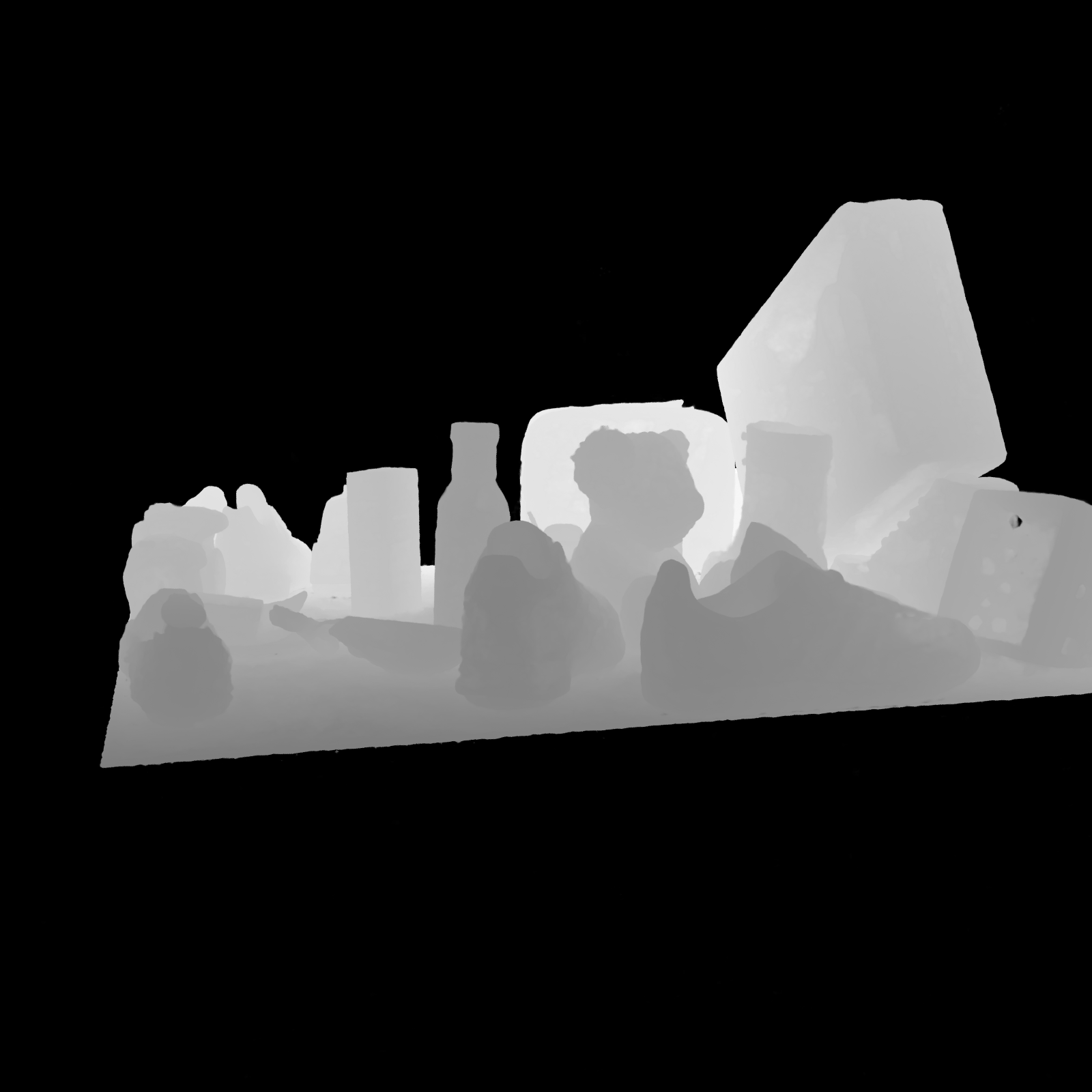}
\\

\rotatebox[origin=lt]{90}{\small \ \ \ \ \ \ \ \ \ mip-NeRF} &
\includegraphics[trim=0 0 0 -5, width=0.4\columnwidth]{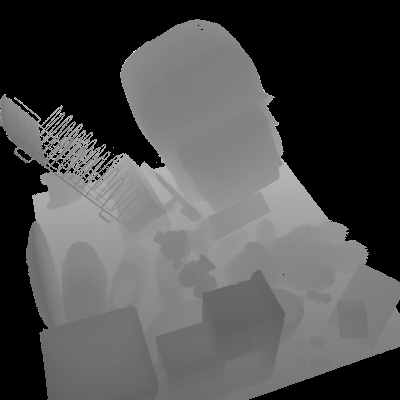}
&

\includegraphics[trim=0 0 0 -5, width=0.4\columnwidth]{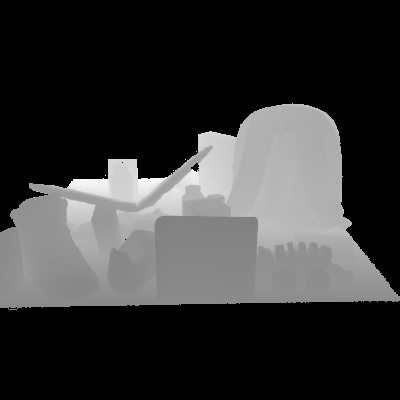}
&

\includegraphics[trim=0 0 0 -5, width=0.4\columnwidth]{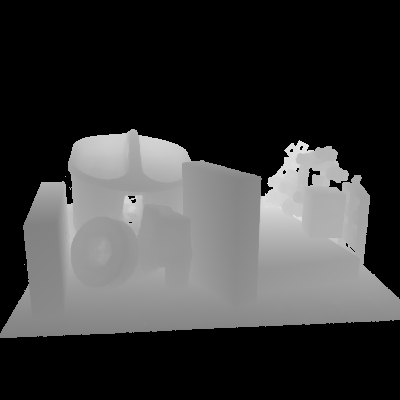}
&

\includegraphics[trim=0 0 0 -5, width=0.4\columnwidth]{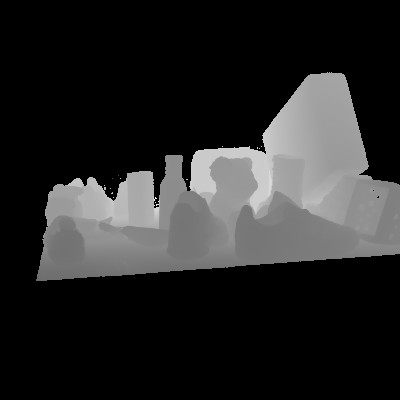}
\\

\rotatebox[origin=lt]{90}{\small \ \ \ \ \ \ \ \ \ \ \ \ \ \ NeRF} &
\includegraphics[trim=0 0 0 -5, width=0.4\columnwidth]{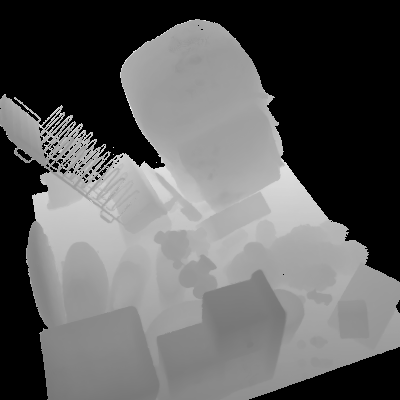}
&

\includegraphics[trim=0 0 0 -5, width=0.4\columnwidth]{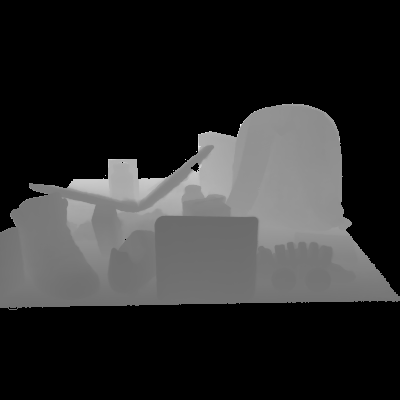}
&

\includegraphics[trim=0 0 0 -5, width=0.4\columnwidth]{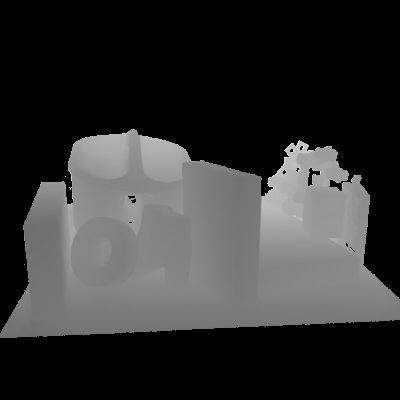}
&

\includegraphics[trim=0 0 0 -5, width=0.4\columnwidth]{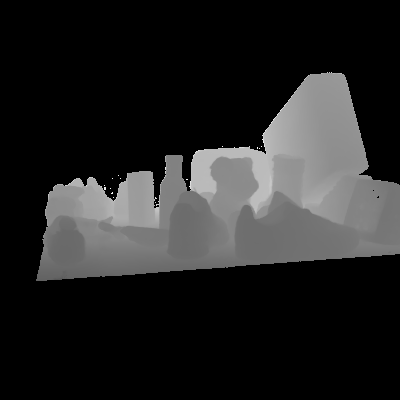}
\\

\end{tabular}
\caption{Depth qualitative results on the Google Scanned Objects environment (figure best seen in zoom).}
\label{fig:more_qual_results_1_depth}
\end{figure*}

\begin{figure*}
\centering
\begin{tabular}{cc|c|c|c}
\rotatebox[origin=lt]{90}{\small \ \ \ \ \ \ \ \ \ Ground truth} &
\includegraphics[trim=0 0 0 -5, width=0.4\columnwidth]{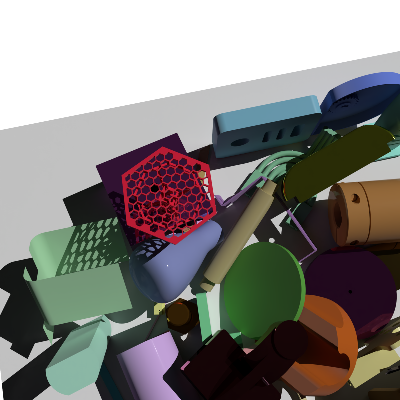}
&

\includegraphics[trim=0 0 0 -5, width=0.4\columnwidth]{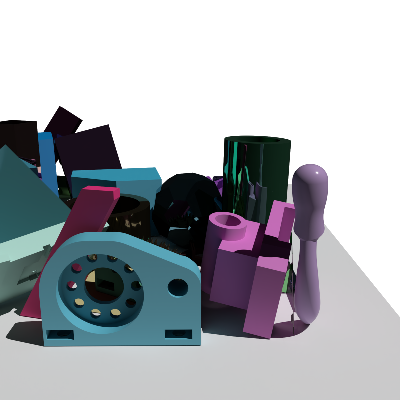}
&

\includegraphics[trim=0 0 0 -5, width=0.4\columnwidth]{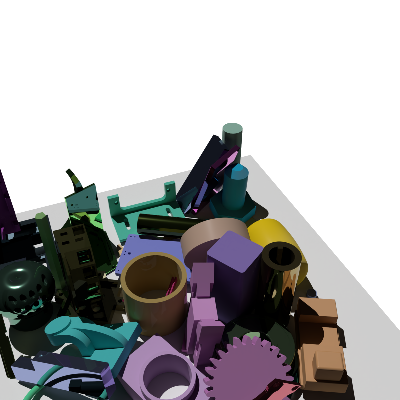}
&

\includegraphics[trim=0 0 0 -5, width=0.4\columnwidth]{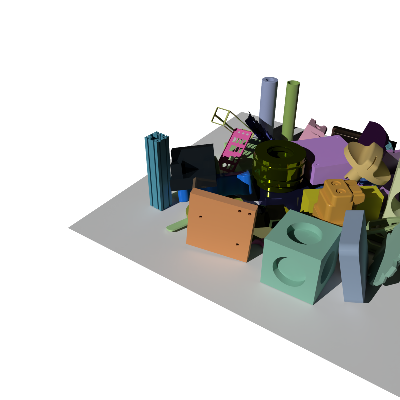}
\\

\rotatebox[origin=lt]{90}{\small \ \ \ \ \ \ \ \ \ \ \ \ \ \ SVLF} &
\includegraphics[trim=0 0 0 -5, width=0.4\columnwidth]{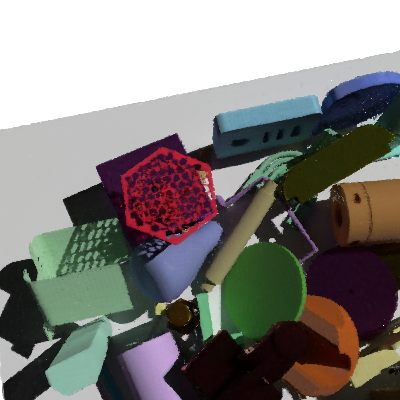}
&

\includegraphics[trim=0 0 0 -5, width=0.4\columnwidth]{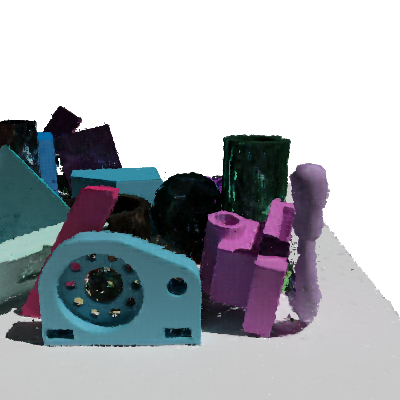}
&

\includegraphics[trim=0 0 0 -5, width=0.4\columnwidth]{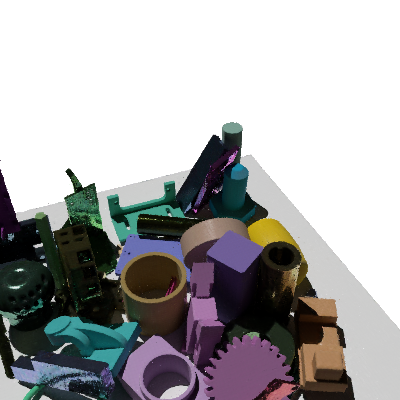}
&

\includegraphics[trim=0 0 0 -5, width=0.4\columnwidth]{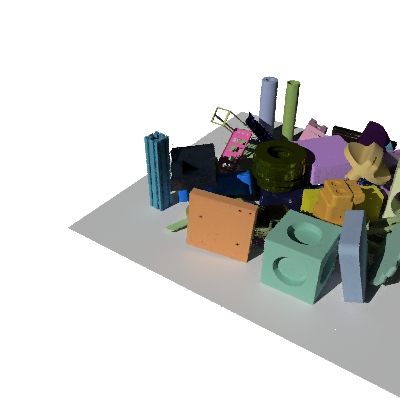}
\\

\rotatebox[origin=lt]{90}{\small \ \ \ \ \ \ \ \ \ \ Ins.-NGP} &
\includegraphics[trim=0 0 0 -5, width=0.4\columnwidth]{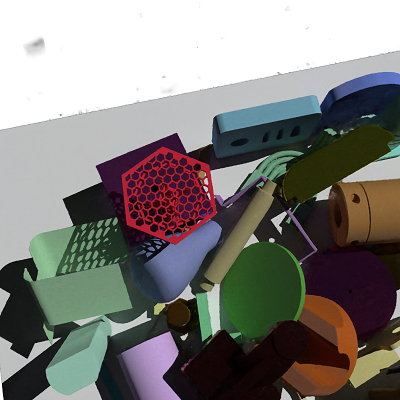}
&

\includegraphics[trim=0 0 0 -5, width=0.4\columnwidth]{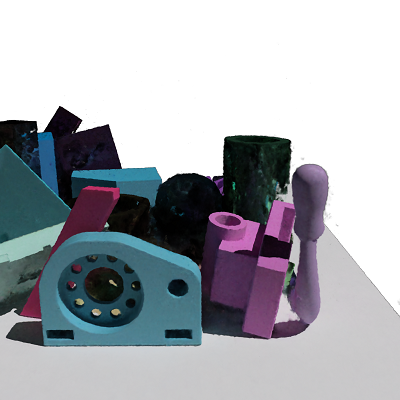}
&

\includegraphics[trim=0 0 0 -5, width=0.4\columnwidth]{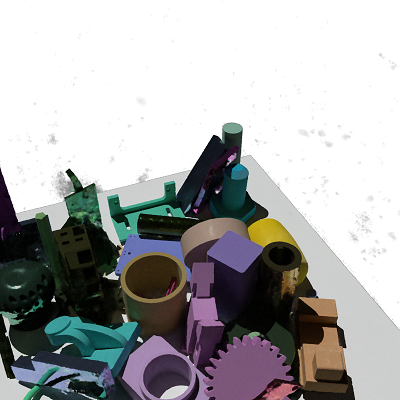}
&

\includegraphics[trim=0 0 0 -5, width=0.4\columnwidth]{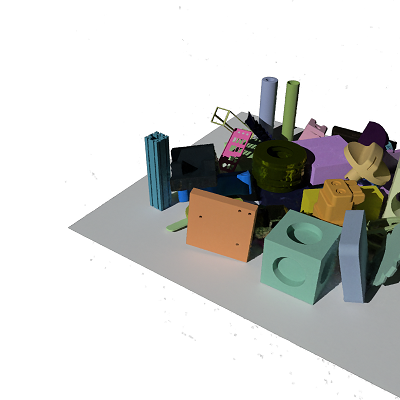}
\\

\rotatebox[origin=lt]{90}{\small \ \ \ \ \ \ \ \ \ mip-NeRF} &
\includegraphics[trim=0 0 0 -5, width=0.4\columnwidth]{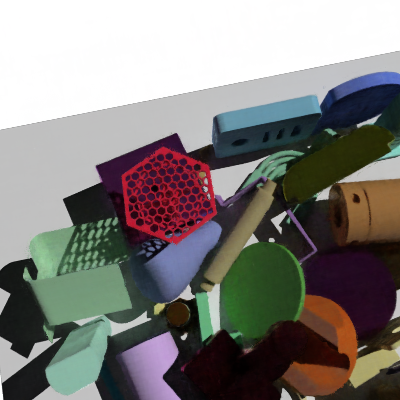}
&

\includegraphics[trim=0 0 0 -5, width=0.4\columnwidth]{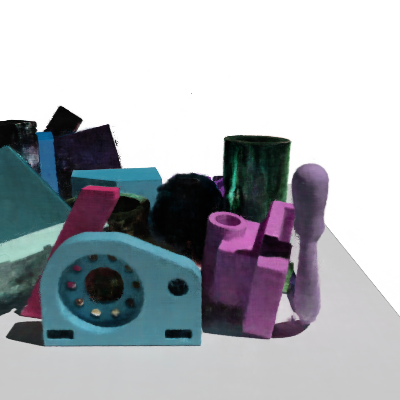}
&

\includegraphics[trim=0 0 0 -5, width=0.4\columnwidth]{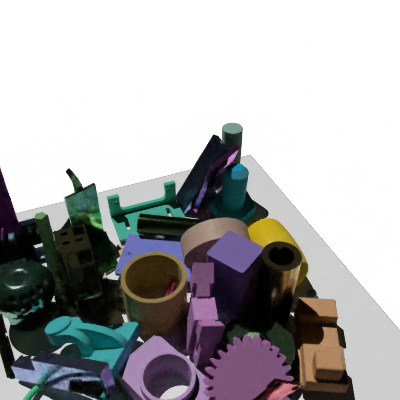}
&

\includegraphics[trim=0 0 0 -5, width=0.4\columnwidth]{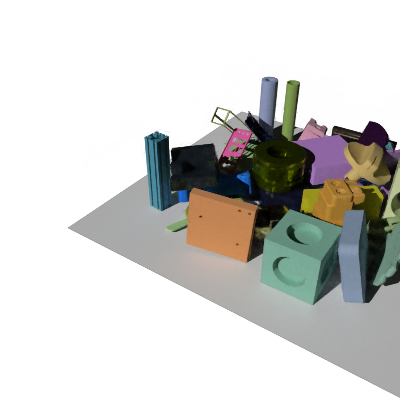}
\\

\rotatebox[origin=lt]{90}{\small \ \ \ \ \ \ \ \ \ \ \ \ \ \ NeRF} &
\includegraphics[trim=0 0 0 -5, width=0.4\columnwidth]{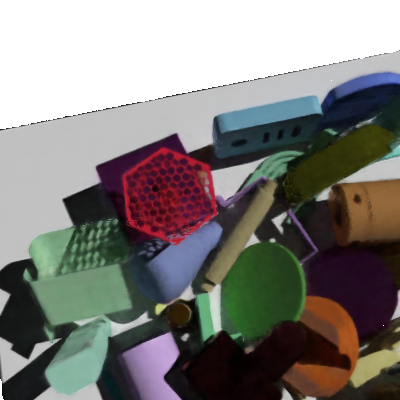}
&

\includegraphics[trim=0 0 0 -5, width=0.4\columnwidth]{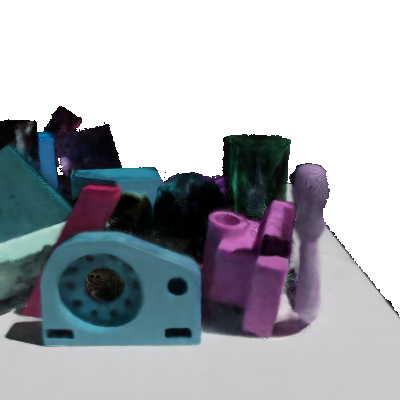}
&

\includegraphics[trim=0 0 0 -5, width=0.4\columnwidth]{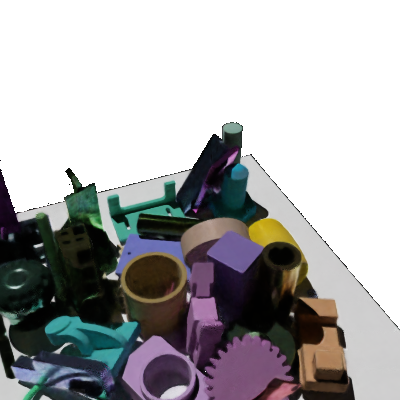}
&

\includegraphics[trim=0 0 0 -5, width=0.4\columnwidth]{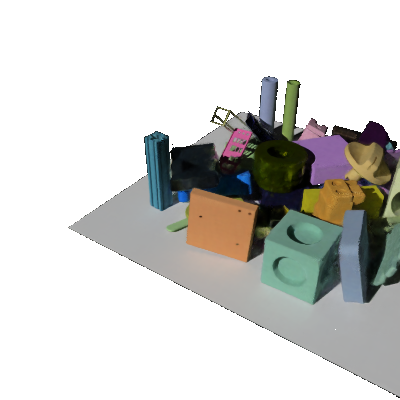}
\\

\end{tabular}
\caption{More qualitative results on the ABC environment (figure best seen in zoom).}
\label{fig:more_qual_results_2}
\end{figure*}

\begin{figure*}
\centering
\begin{tabular}{cc|c|c|c}
\rotatebox[origin=lt]{90}{\small \ \ \ \ \ \ \ \ \ Ground truth} &
\includegraphics[trim=0 0 0 -5, width=0.4\columnwidth]{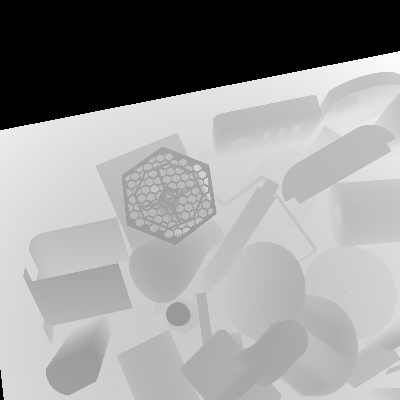}
&

\includegraphics[trim=0 0 0 -5, width=0.4\columnwidth]{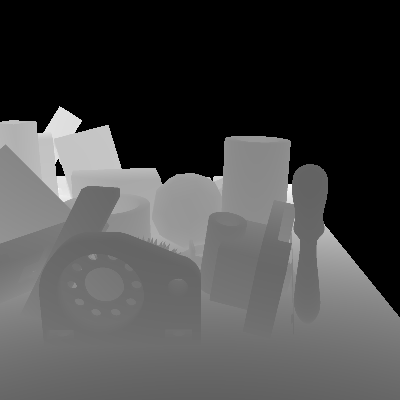}
&

\includegraphics[trim=0 0 0 -5, width=0.4\columnwidth]{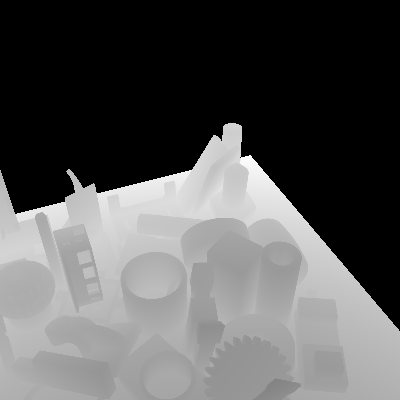}
&

\includegraphics[trim=0 0 0 -5, width=0.4\columnwidth]{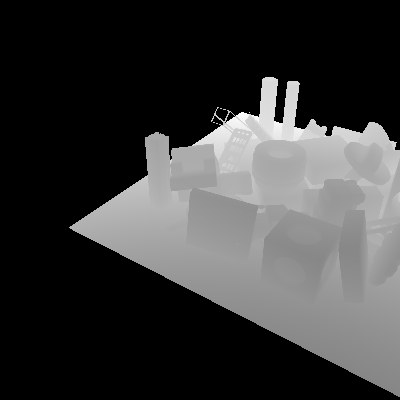}
\\

\rotatebox[origin=lt]{90}{\small \ \ \ \ \ \ \ \ \ \ \ \ \ \ SVLF} &
\includegraphics[trim=0 0 0 -5, width=0.4\columnwidth]{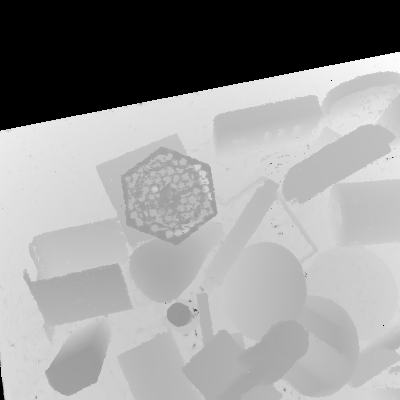}
&

\includegraphics[trim=0 0 0 -5, width=0.4\columnwidth]{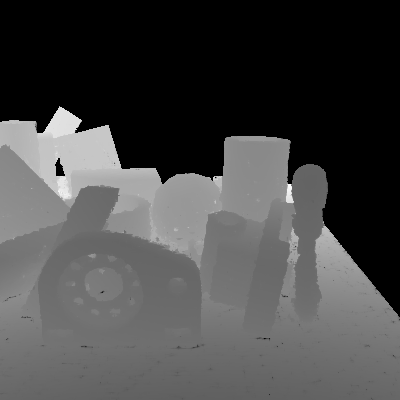}
&

\includegraphics[trim=0 0 0 -5, width=0.4\columnwidth]{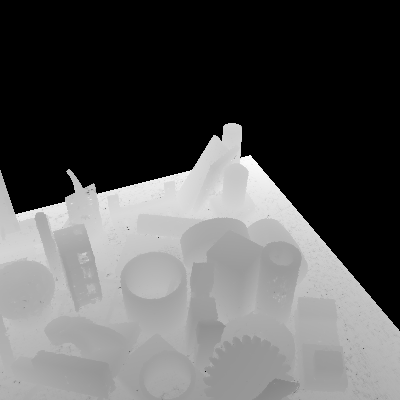}
&

\includegraphics[trim=0 0 0 -5, width=0.4\columnwidth]{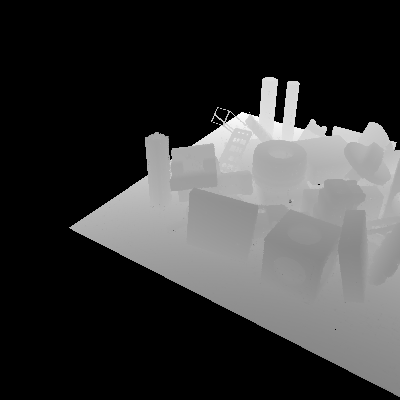}
\\

\rotatebox[origin=lt]{90}{\small \ \ \ \ \ \ \ \ \ \ Ins.-NGP} &
\includegraphics[trim=0 0 0 -5, width=0.4\columnwidth]{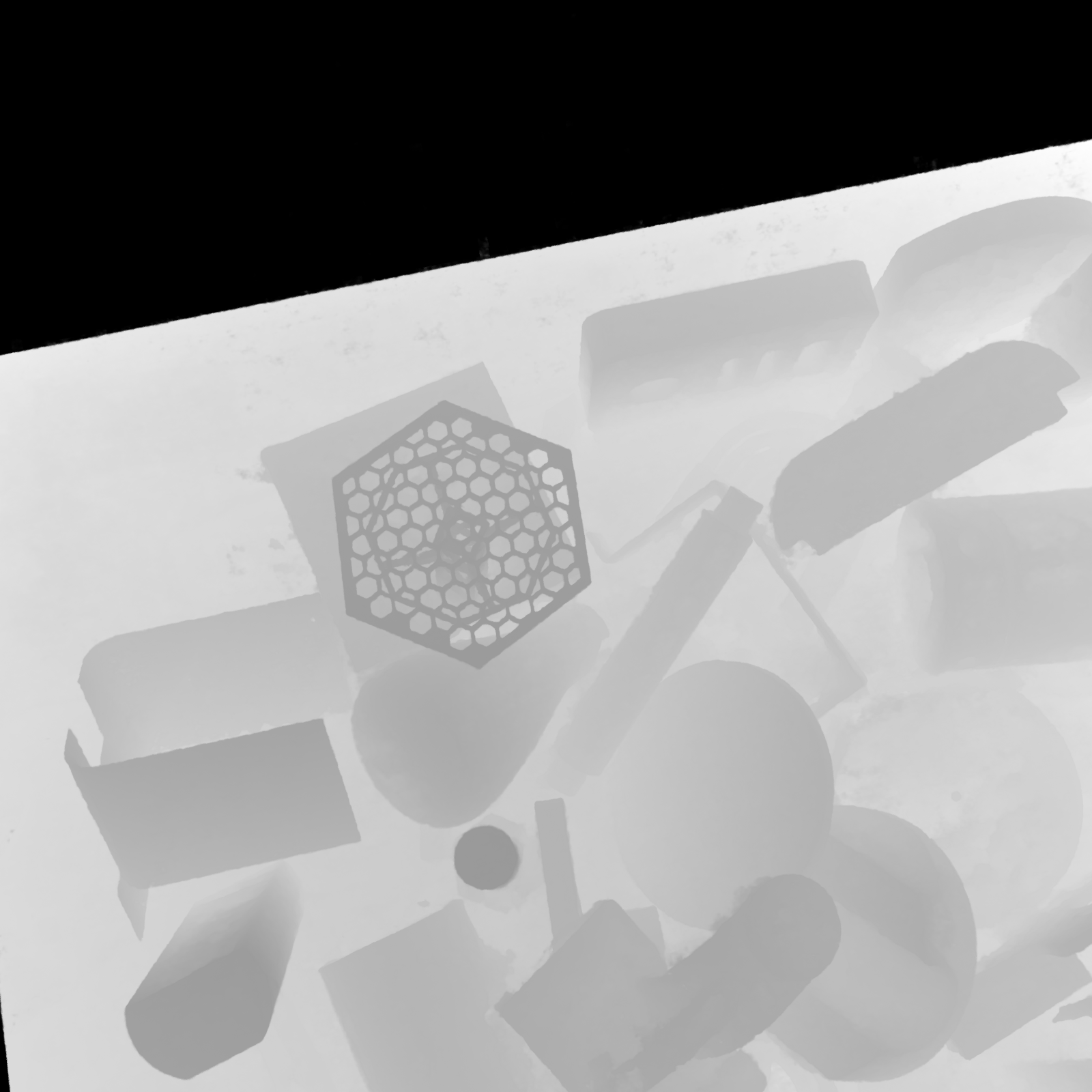}
&

\includegraphics[trim=0 0 0 -5, width=0.4\columnwidth]{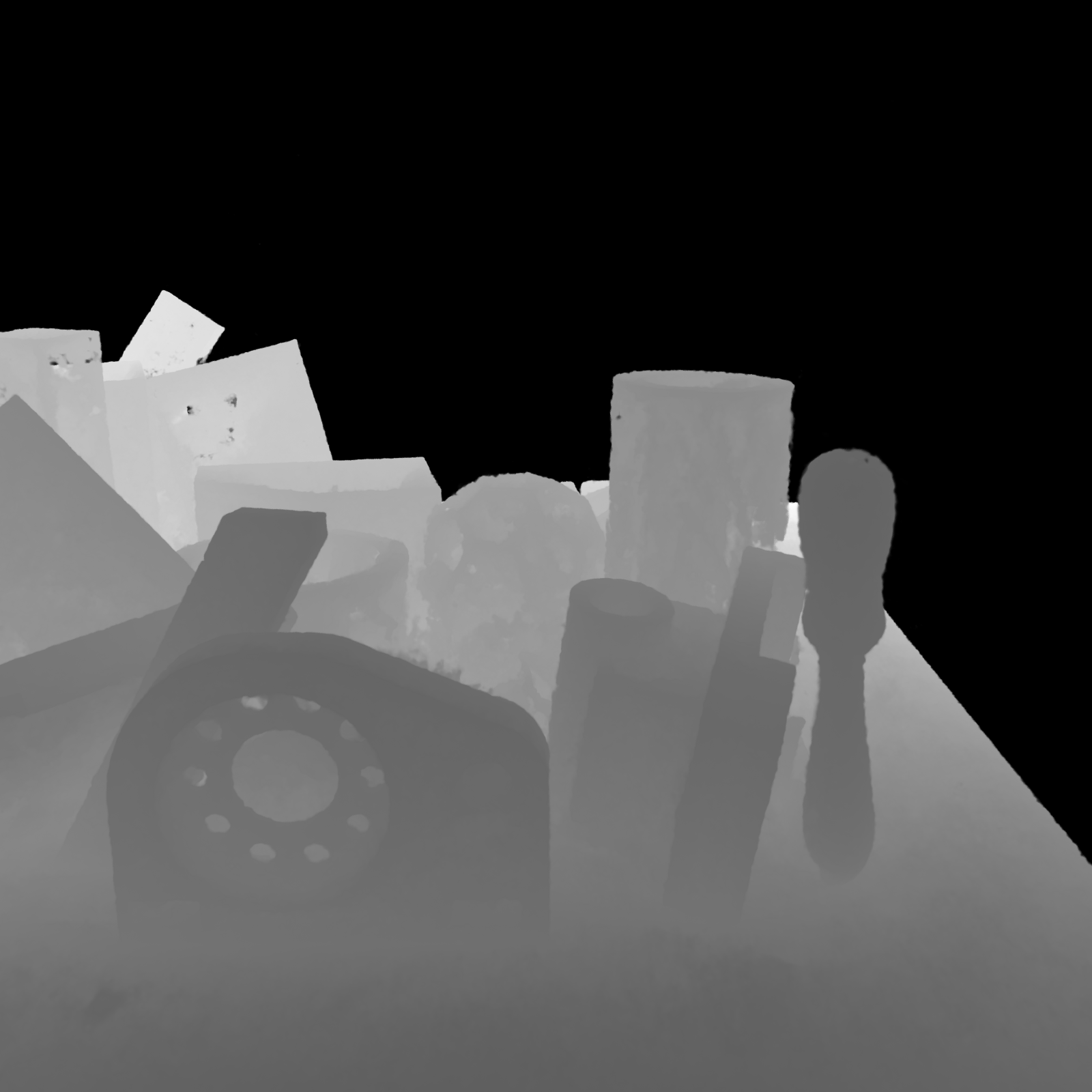}
&

\includegraphics[trim=0 0 0 -5, width=0.4\columnwidth]{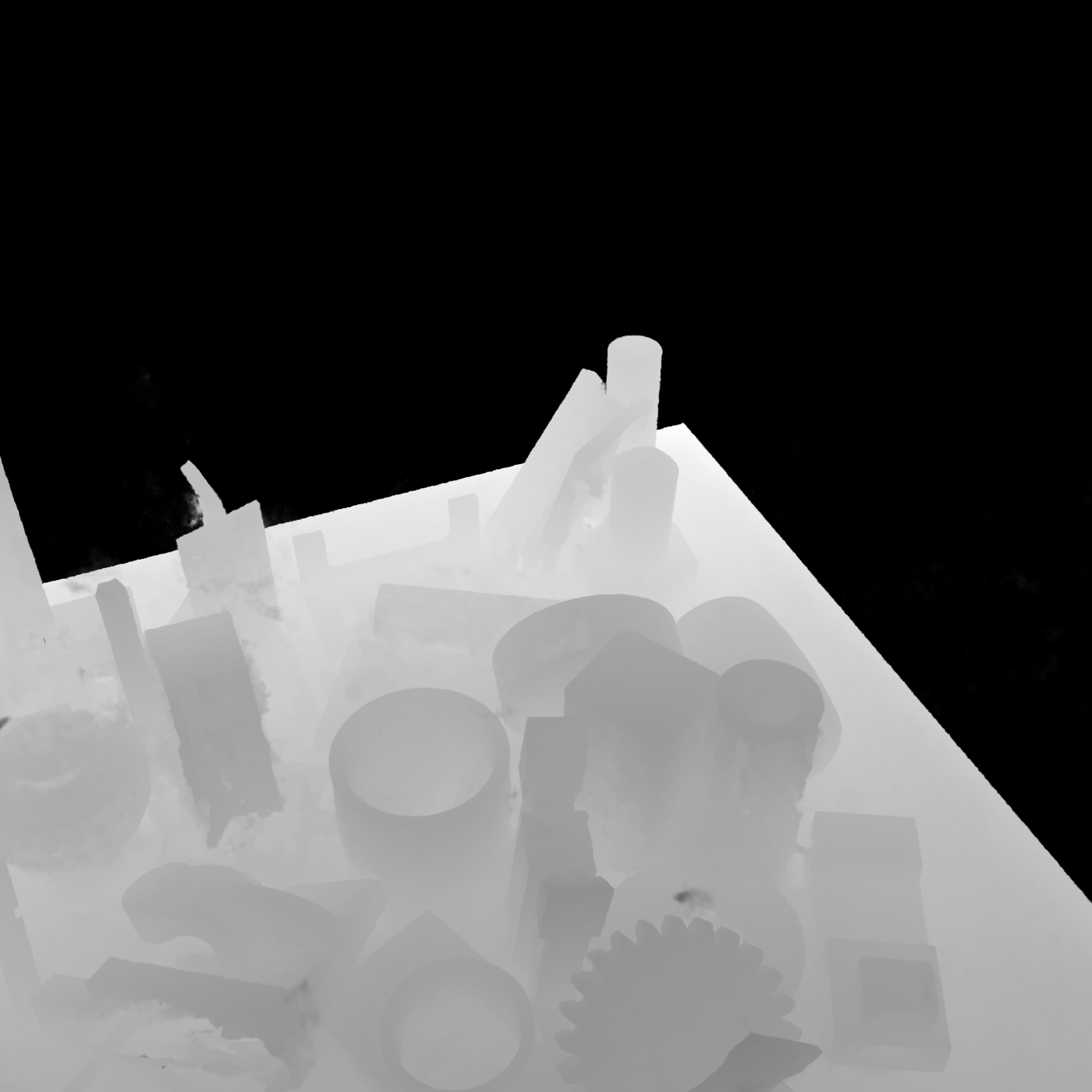}
&

\includegraphics[trim=0 0 0 -5, width=0.4\columnwidth]{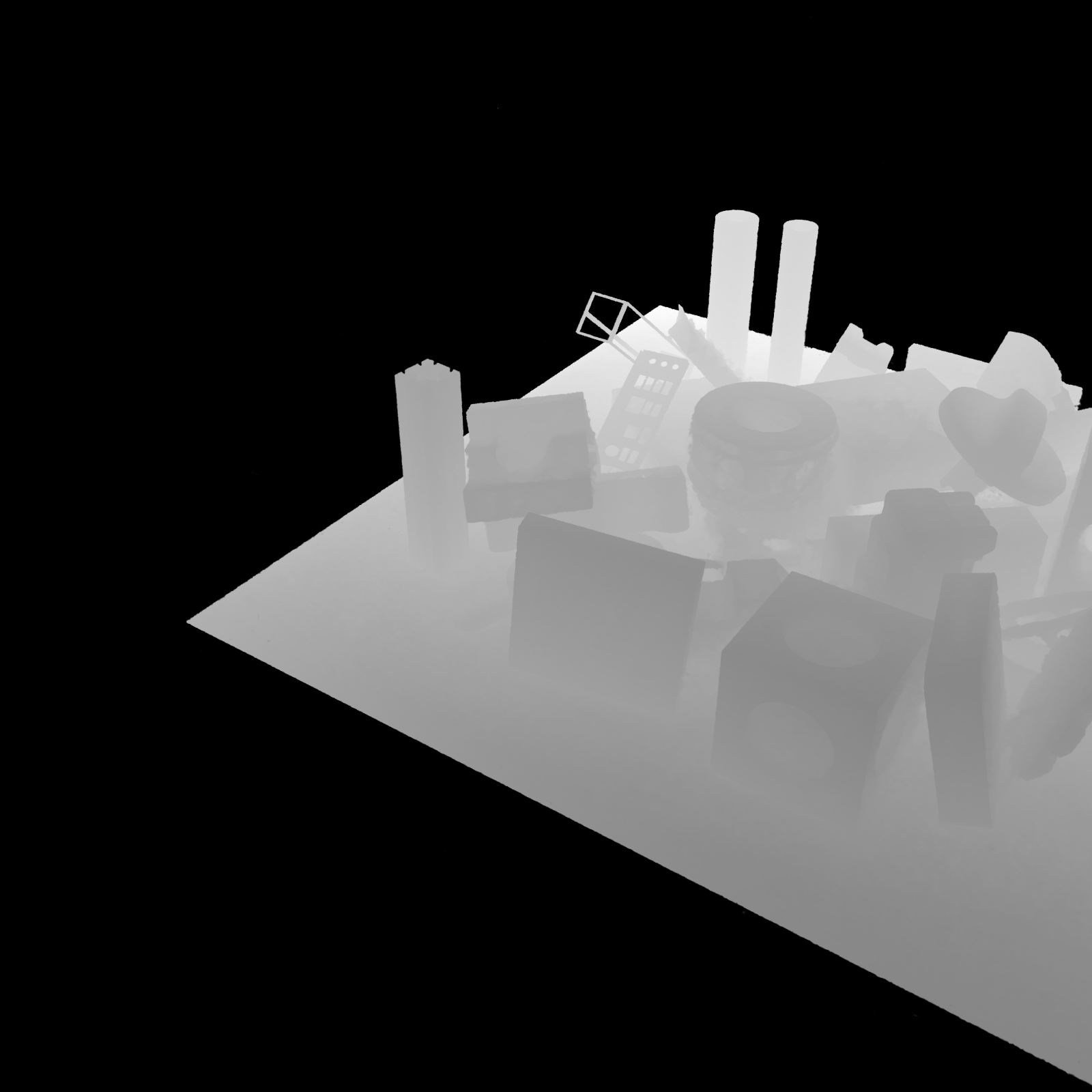}
\\

\rotatebox[origin=lt]{90}{\small \ \ \ \ \ \ \ \ \ mip-NeRF} &
\includegraphics[trim=0 0 0 -5, width=0.4\columnwidth]{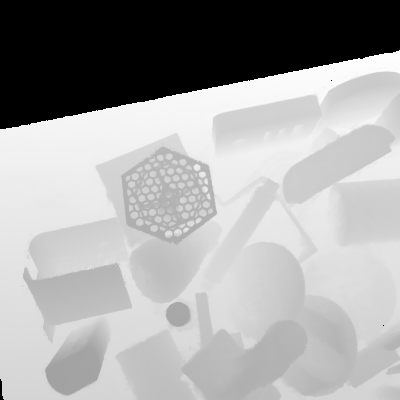}
&

\includegraphics[trim=0 0 0 -5, width=0.4\columnwidth]{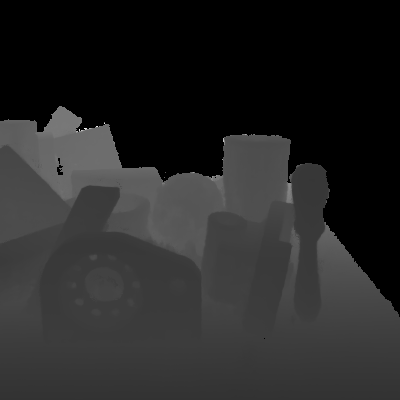}
&

\includegraphics[trim=0 0 0 -5, width=0.4\columnwidth]{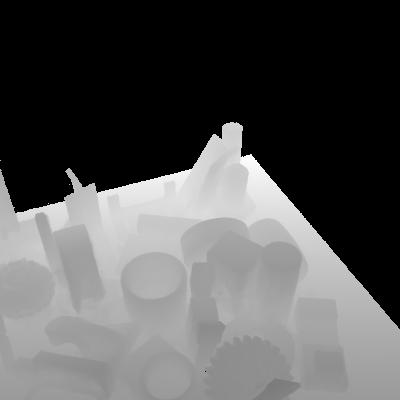}
&

\includegraphics[trim=0 0 0 -5, width=0.4\columnwidth]{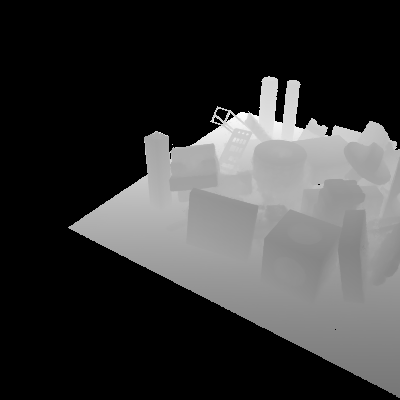}
\\

\rotatebox[origin=lt]{90}{\small \ \ \ \ \ \ \ \ \ \ \ \ \ \ NeRF} &
\includegraphics[trim=0 0 0 -5, width=0.4\columnwidth]{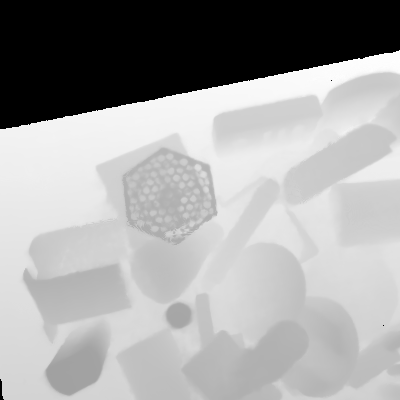}
&

\includegraphics[trim=0 0 0 -5, width=0.4\columnwidth]{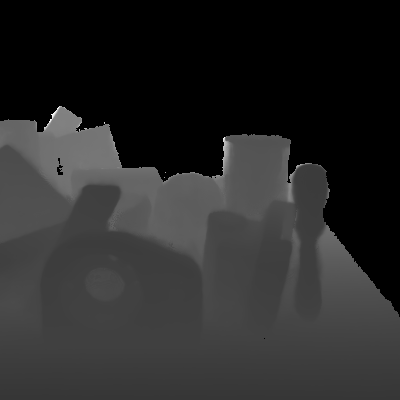}
&

\includegraphics[trim=0 0 0 -5, width=0.4\columnwidth]{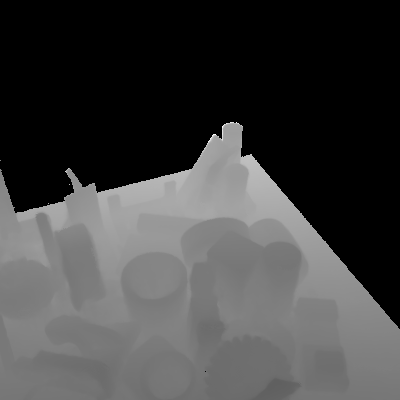}
&

\includegraphics[trim=0 0 0 -5, width=0.4\columnwidth]{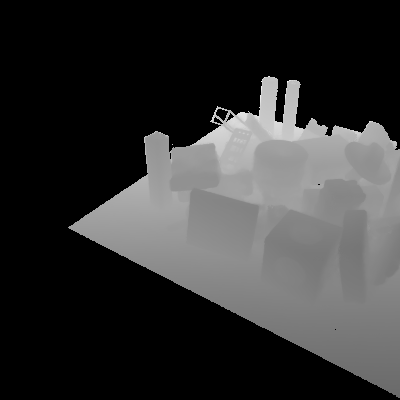}
\\

\end{tabular}
\caption{Depth qualitative results on the ABC environment (figure best seen in zoom).}
\label{fig:more_qual_results_2_depth}
\end{figure*}

\begin{figure*}
\centering
\begin{tabular}{cc|c|c|c}
\rotatebox[origin=lt]{90}{\small \ \ \ \ \ \ \ \ \ Ground truth} &
\includegraphics[trim=0 0 0 -5, width=0.4\columnwidth]{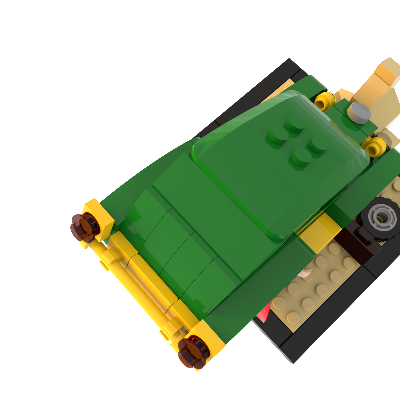}
&

\includegraphics[trim=0 0 0 -5, width=0.4\columnwidth]{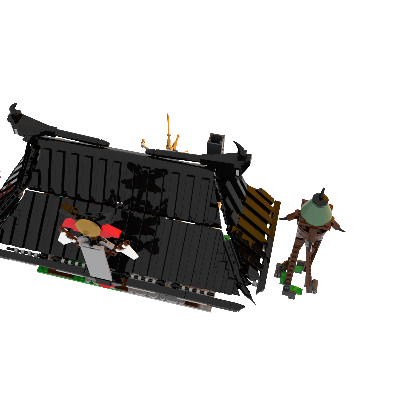}
&

\includegraphics[trim=0 0 0 -5, width=0.4\columnwidth]{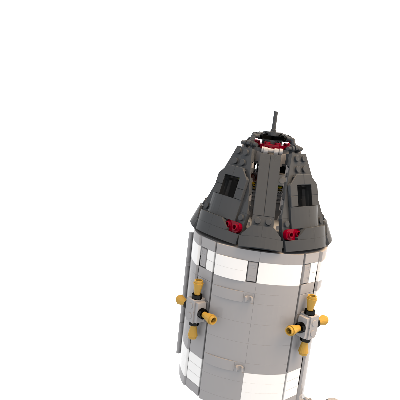}
&

\includegraphics[trim=0 0 0 -5, width=0.4\columnwidth]{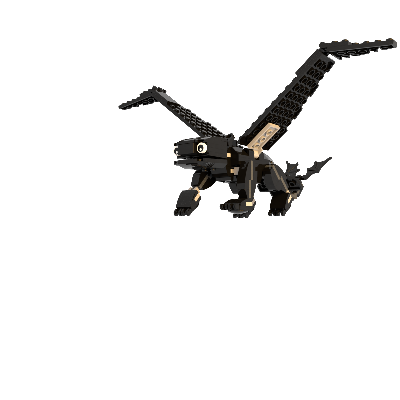}
\\

\rotatebox[origin=lt]{90}{\small \ \ \ \ \ \ \ \ \ \ \ \ \ \ SVLF} &
\includegraphics[trim=0 0 0 -5, width=0.4\columnwidth]{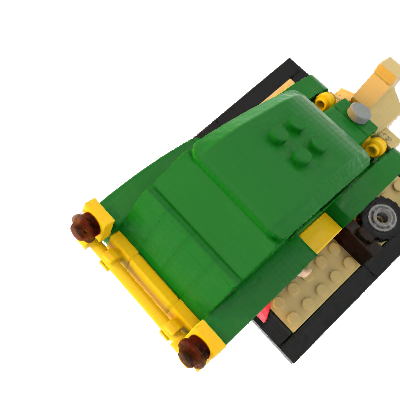}
&

\includegraphics[trim=0 0 0 -5, width=0.4\columnwidth]{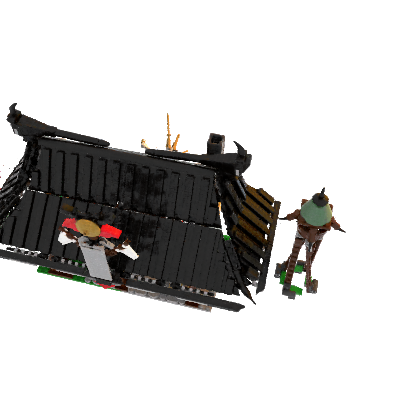}
&

\includegraphics[trim=0 0 0 -5, width=0.4\columnwidth]{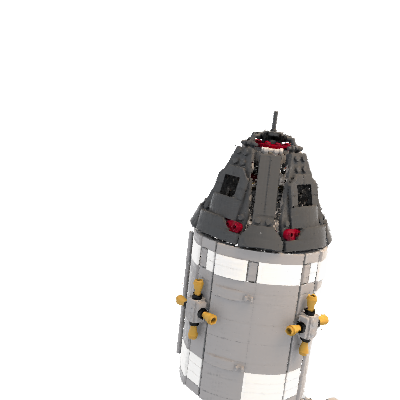}
&

\includegraphics[trim=0 0 0 -5, width=0.4\columnwidth]{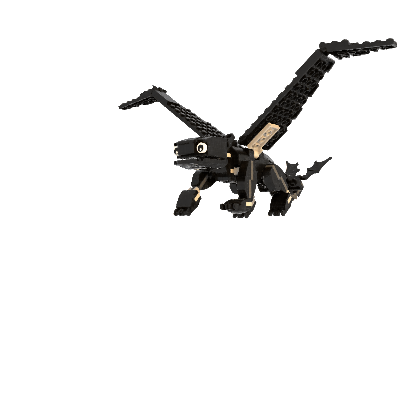}
\\

\rotatebox[origin=lt]{90}{\small \ \ \ \ \ \ \ \ \ \ Ins.-NGP} &
\includegraphics[trim=0 0 0 -5, width=0.4\columnwidth]{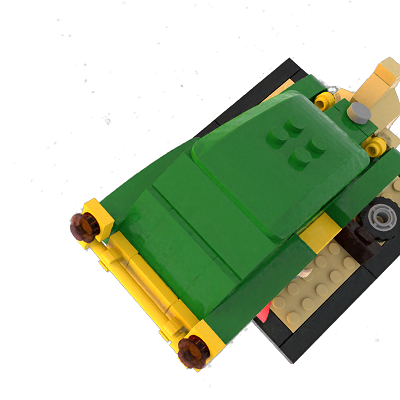}
&

\includegraphics[trim=0 0 0 -5, width=0.4\columnwidth]{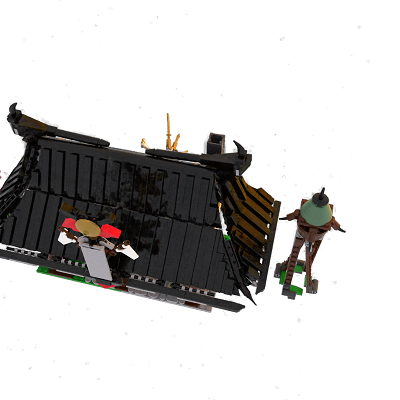}
&

\includegraphics[trim=0 0 0 -5, width=0.4\columnwidth]{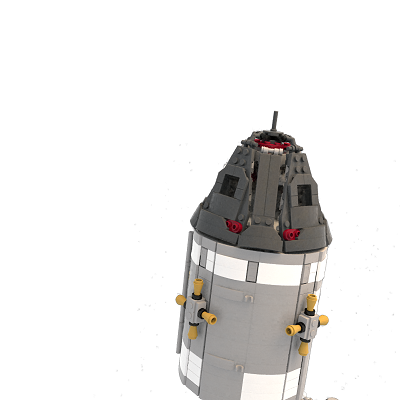}
&

\includegraphics[trim=0 0 0 -5, width=0.4\columnwidth]{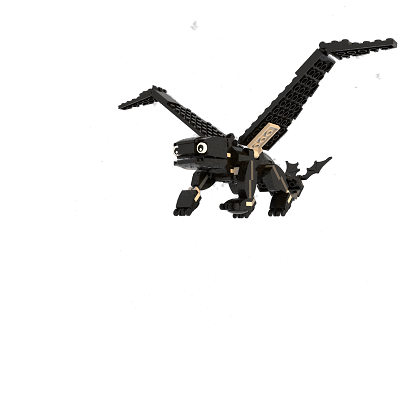}
\\

\rotatebox[origin=lt]{90}{\small \ \ \ \ \ \ \ \ \ mip-NeRF} &
\includegraphics[trim=0 0 0 -5, width=0.4\columnwidth]{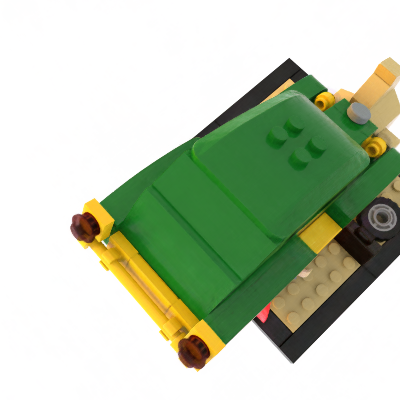}
&

\includegraphics[trim=0 0 0 -5, width=0.4\columnwidth]{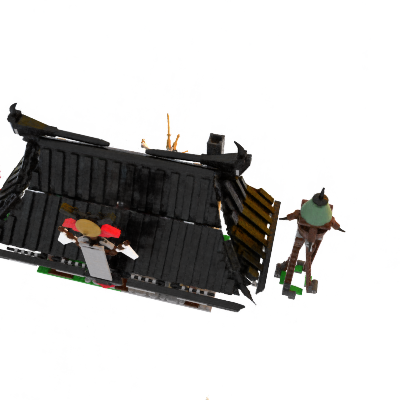}
&

\includegraphics[trim=0 0 0 -5, width=0.4\columnwidth]{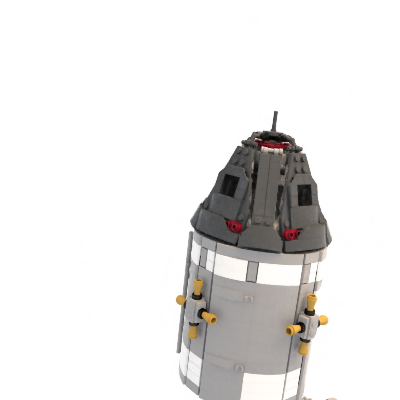}
&

\includegraphics[trim=0 0 0 -5, width=0.4\columnwidth]{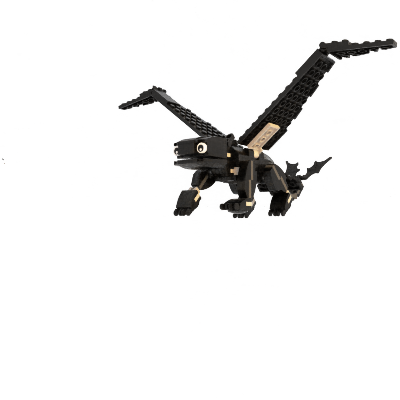}
\\

\rotatebox[origin=lt]{90}{\small \ \ \ \ \ \ \ \ \ \ \ \ \ \ NeRF} &
\includegraphics[trim=0 0 0 -5, width=0.4\columnwidth]{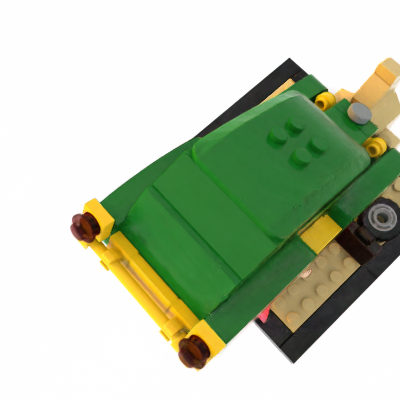}
&

\includegraphics[trim=0 0 0 -5, width=0.4\columnwidth]{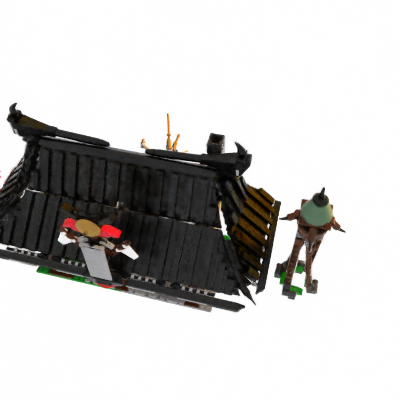}
&

\includegraphics[trim=0 0 0 -5, width=0.4\columnwidth]{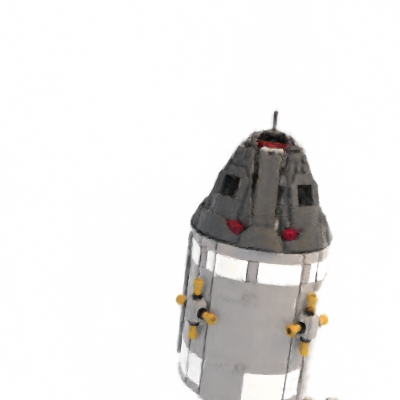}
&

\includegraphics[trim=0 0 0 -5, width=0.4\columnwidth]{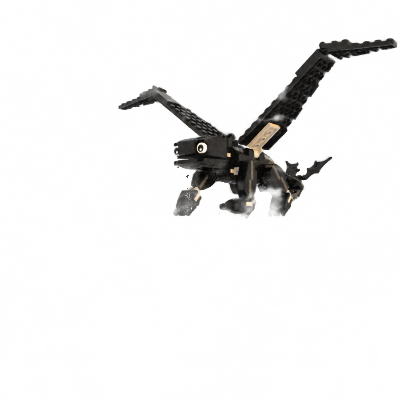}
\\

\end{tabular}
\caption{More qualitative results on the Bricks environment (figure best seen in zoom).}
\label{fig:more_qual_results_3}
\end{figure*}

\begin{figure*}
\centering
\begin{tabular}{cc|c|c|c}
\rotatebox[origin=lt]{90}{\small \ \ \ \ \ \ \ \ \ Ground truth} &
\includegraphics[trim=0 0 0 -5, width=0.4\columnwidth]{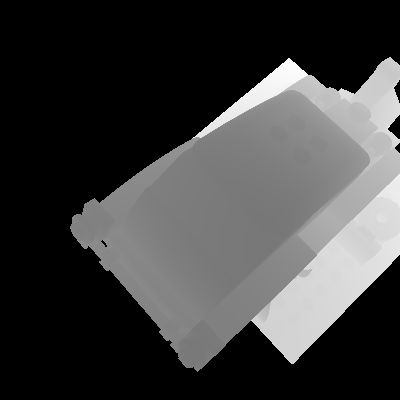}
&

\includegraphics[trim=0 0 0 -5, width=0.4\columnwidth]{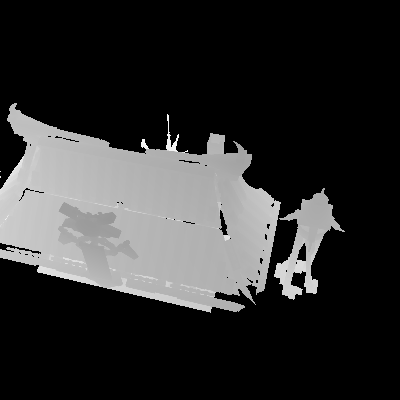}
&

\includegraphics[trim=0 0 0 -5, width=0.4\columnwidth]{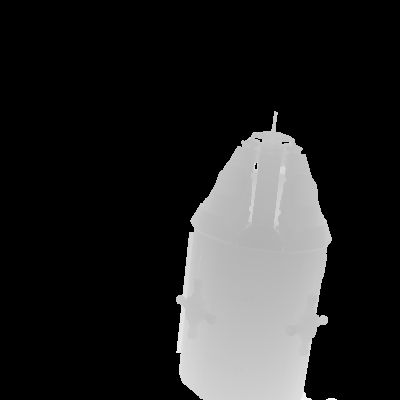}
&

\includegraphics[trim=0 0 0 -5, width=0.4\columnwidth]{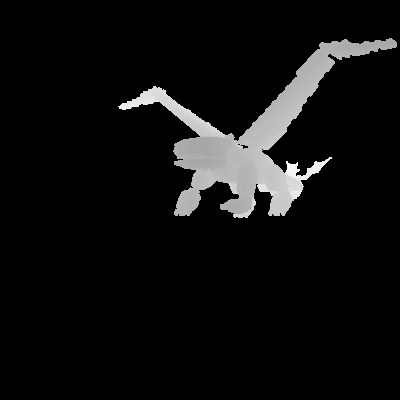}
\\

\rotatebox[origin=lt]{90}{\small \ \ \ \ \ \ \ \ \ \ \ \ \ \ SVLF} &
\includegraphics[trim=0 0 0 -5, width=0.4\columnwidth]{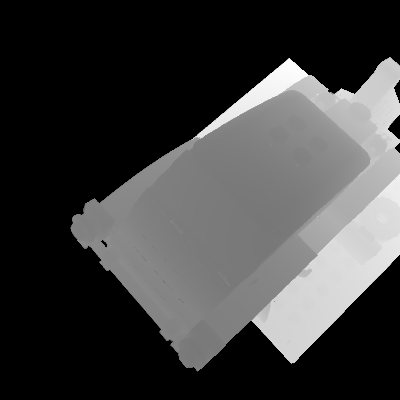}
&

\includegraphics[trim=0 0 0 -5, width=0.4\columnwidth]{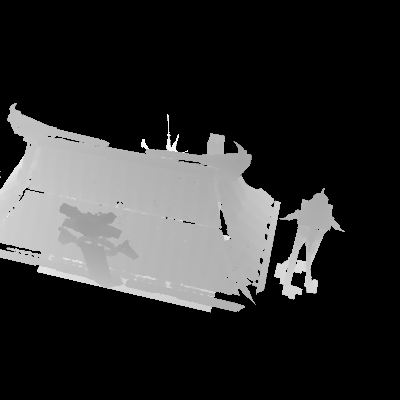}
&

\includegraphics[trim=0 0 0 -5, width=0.4\columnwidth]{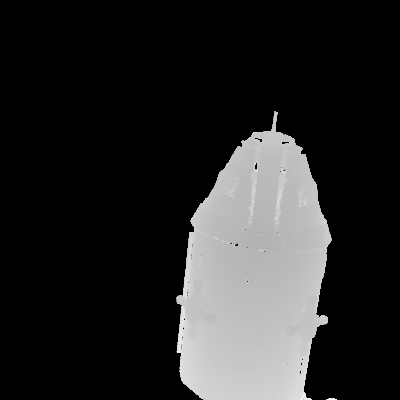}
&

\includegraphics[trim=0 0 0 -5, width=0.4\columnwidth]{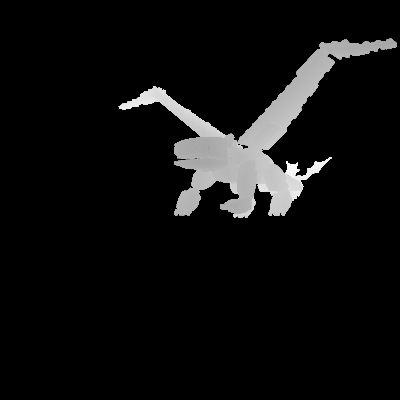}
\\

\rotatebox[origin=lt]{90}{\small \ \ \ \ \ \ \ \ \ \ Ins.-NGP} &
\includegraphics[trim=0 0 0 -5, width=0.4\columnwidth]{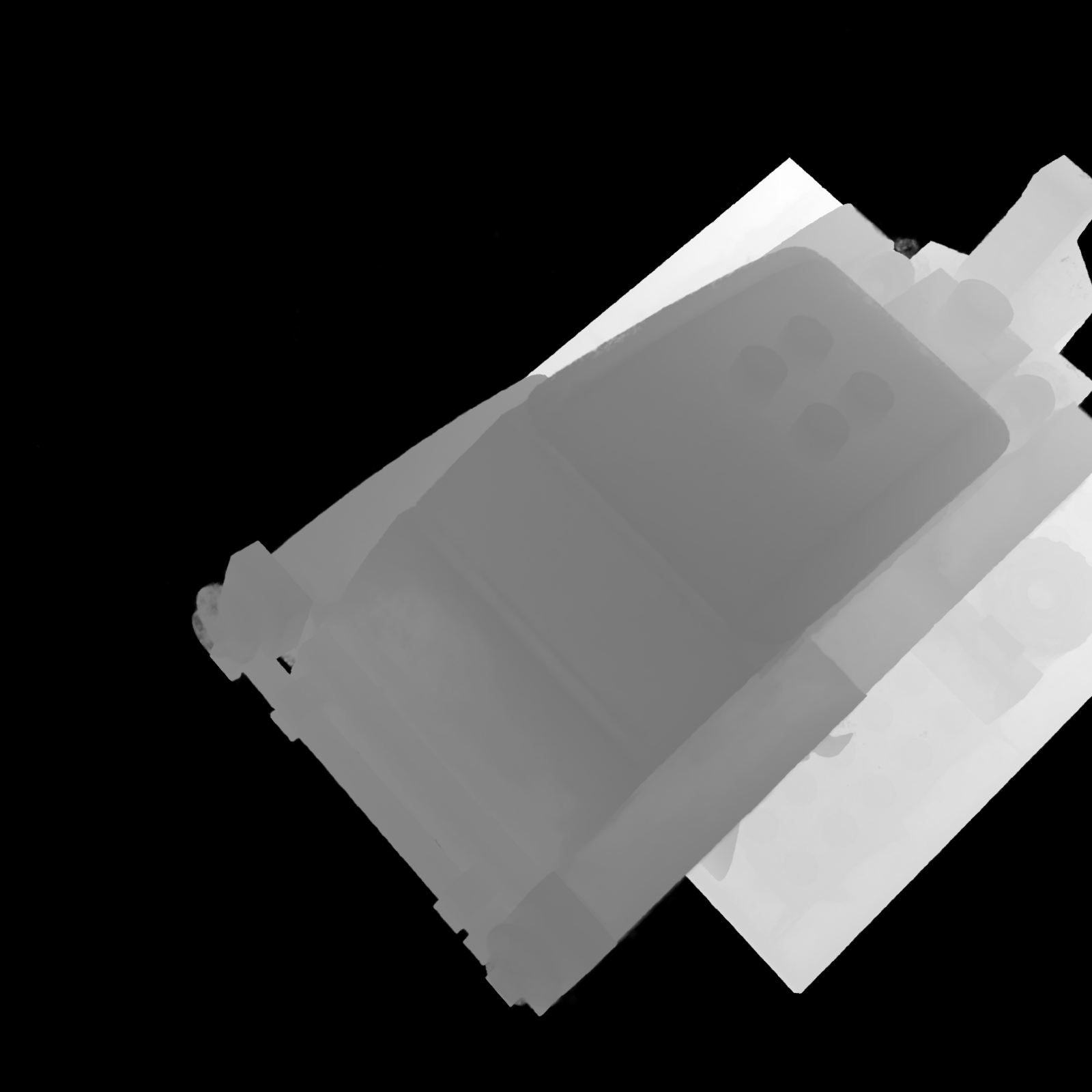}
&

\includegraphics[trim=0 0 0 -5, width=0.4\columnwidth]{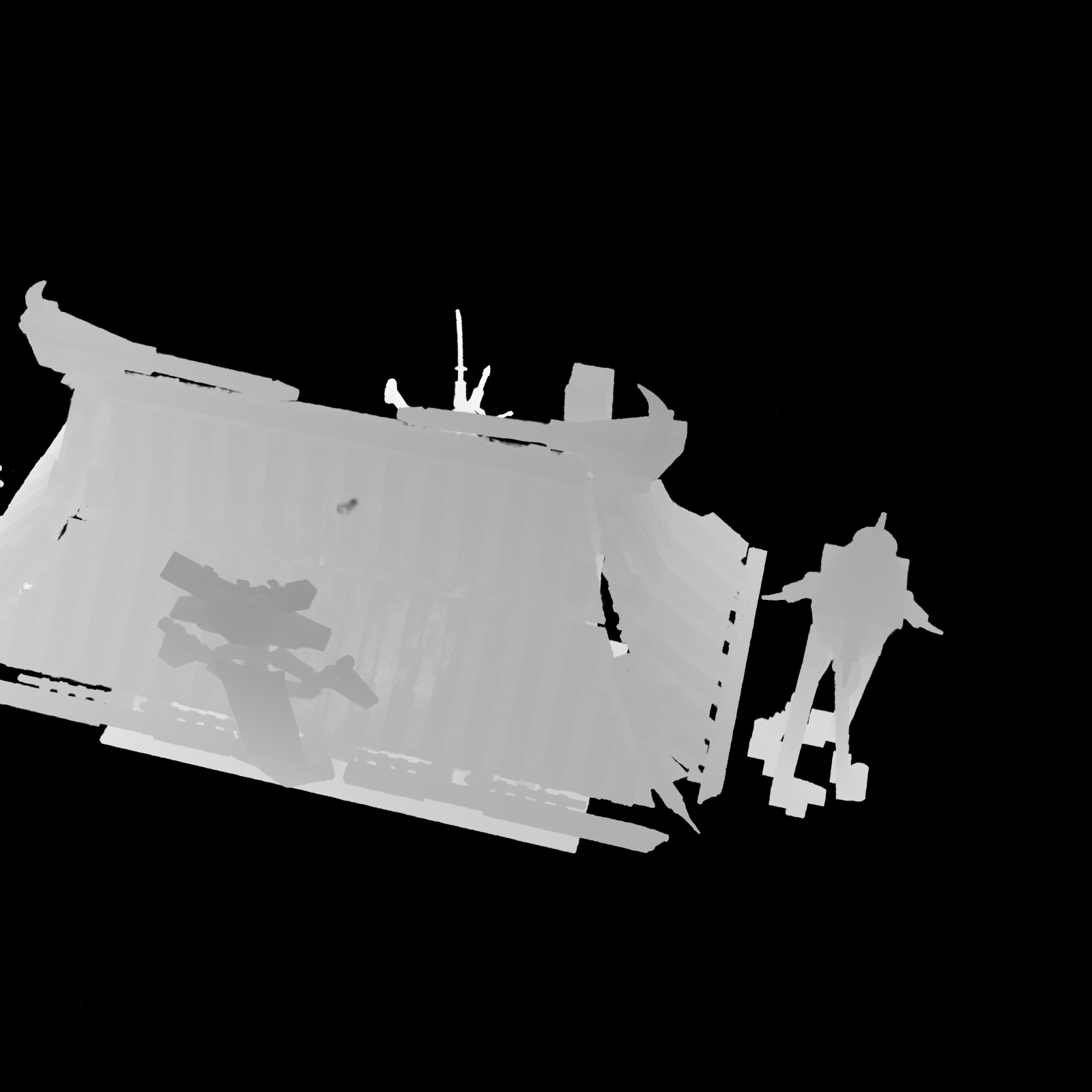}
&

\includegraphics[trim=0 0 0 -5, width=0.4\columnwidth]{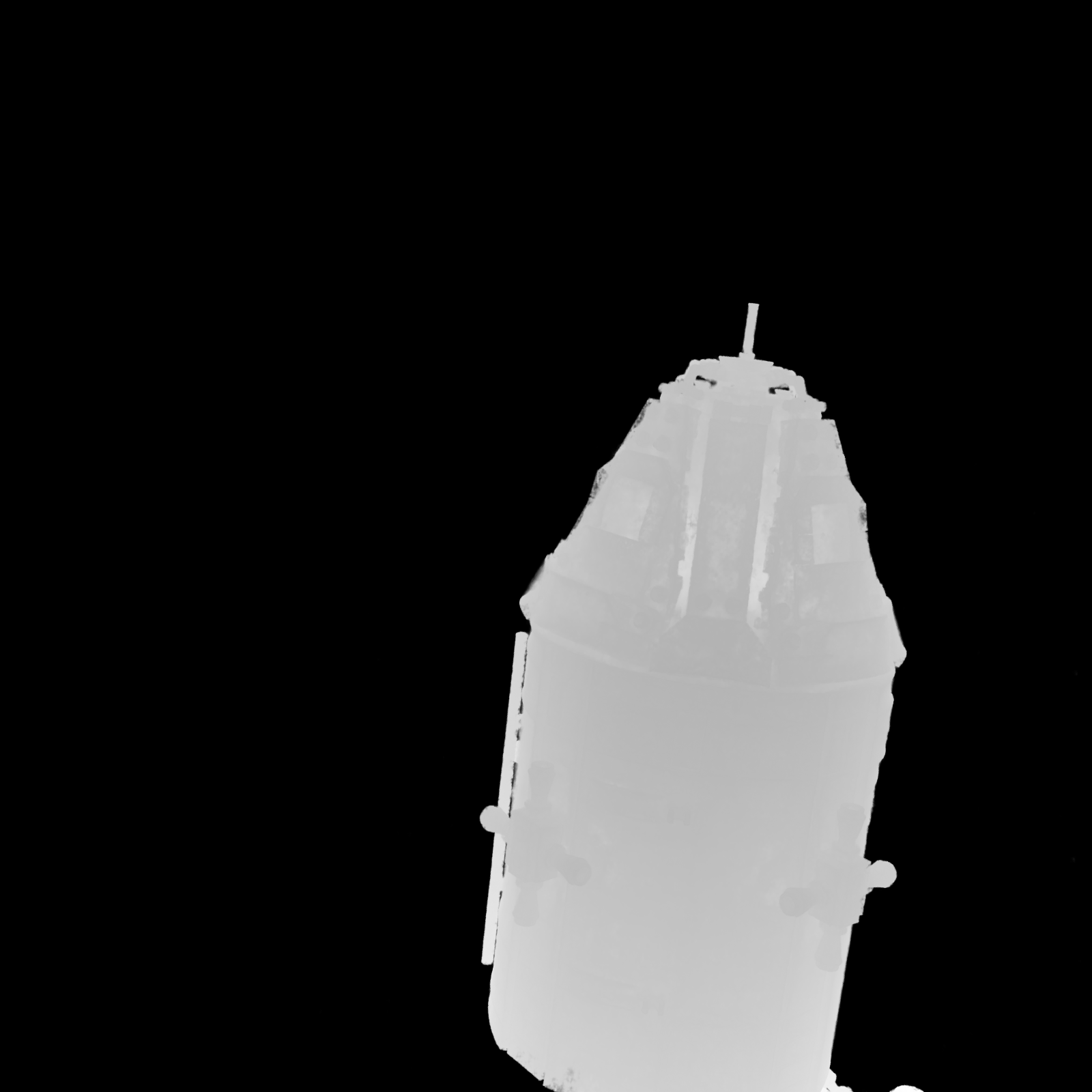}
&

\includegraphics[trim=0 0 0 -5, width=0.4\columnwidth]{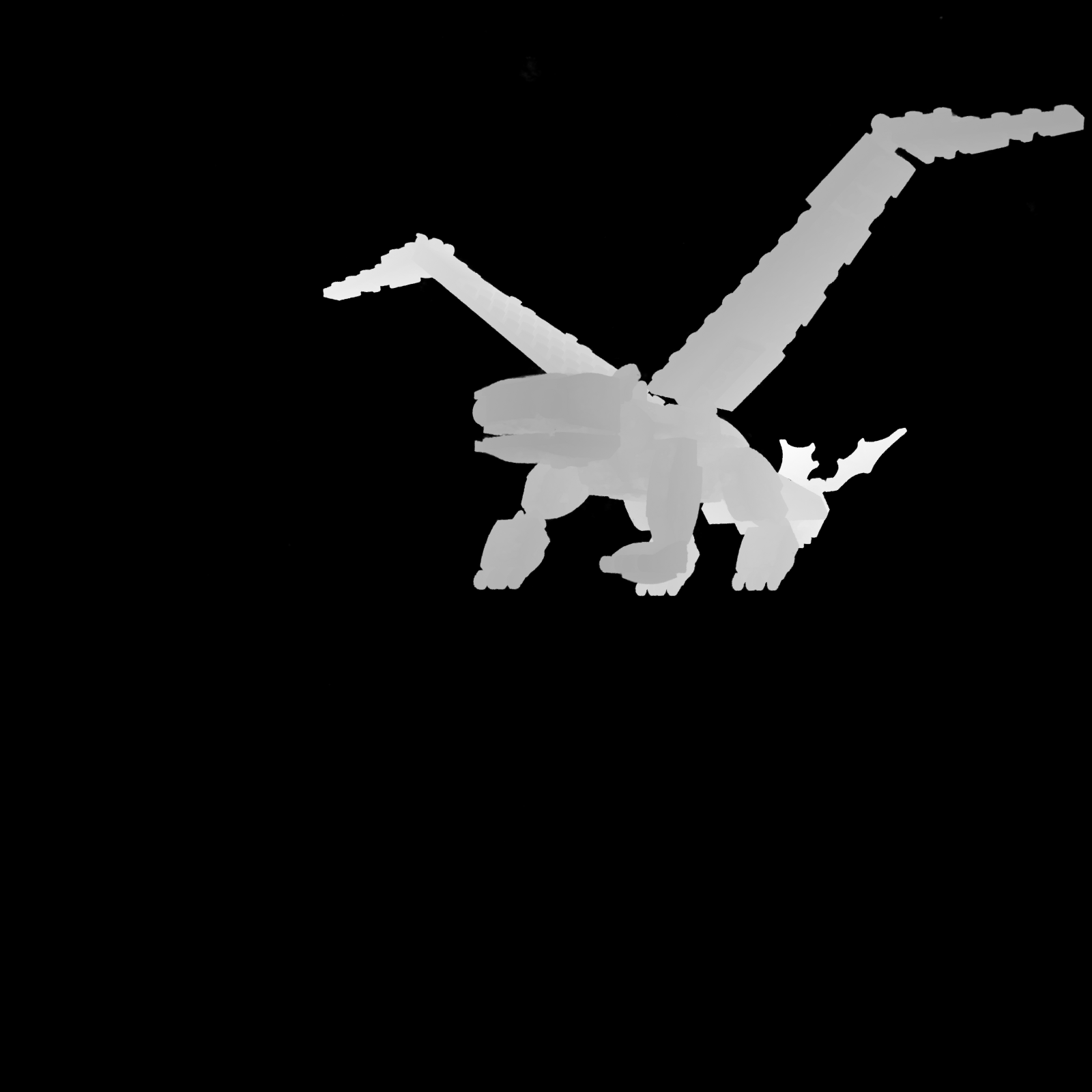}
\\

\rotatebox[origin=lt]{90}{\small \ \ \ \ \ \ \ \ \ mip-NeRF} &
\includegraphics[trim=0 0 0 -5, width=0.4\columnwidth]{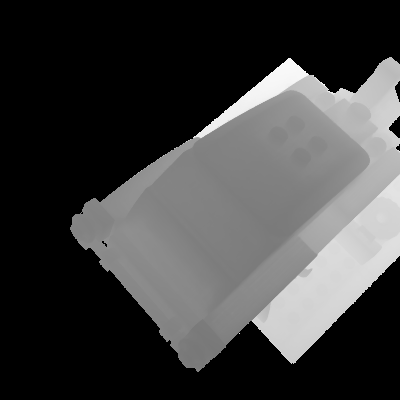}
&

\includegraphics[trim=0 0 0 -5, width=0.4\columnwidth]{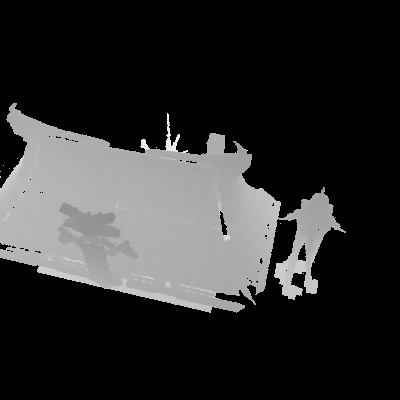}
&

\includegraphics[trim=0 0 0 -5, width=0.4\columnwidth]{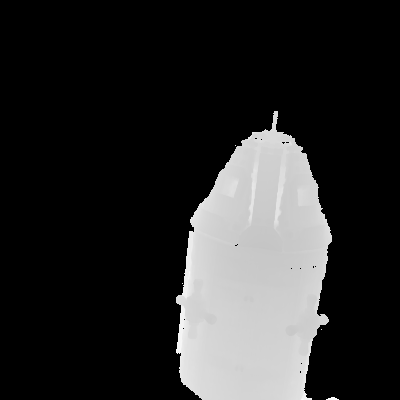}
&

\includegraphics[trim=0 0 0 -5, width=0.4\columnwidth]{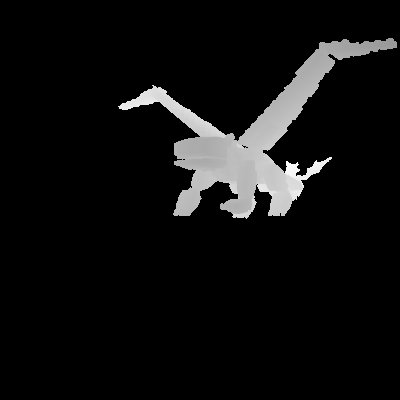}
\\

\rotatebox[origin=lt]{90}{\small \ \ \ \ \ \ \ \ \ \ \ \ \ \ NeRF} &
\includegraphics[trim=0 0 0 -5, width=0.4\columnwidth]{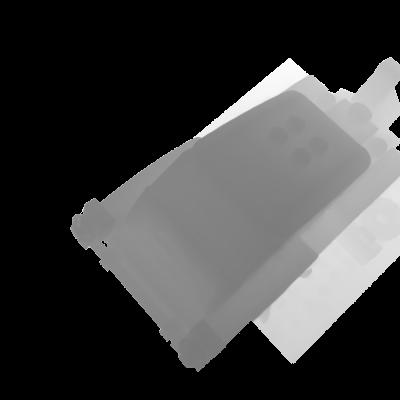}
&

\includegraphics[trim=0 0 0 -5, width=0.4\columnwidth]{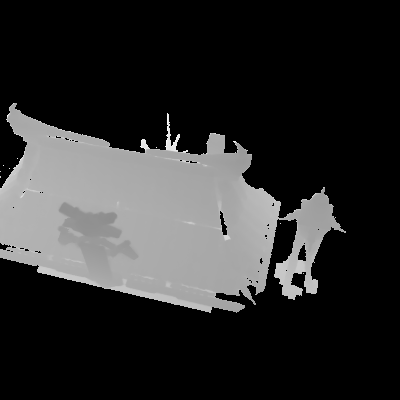}
&

\includegraphics[trim=0 0 0 -5, width=0.4\columnwidth]{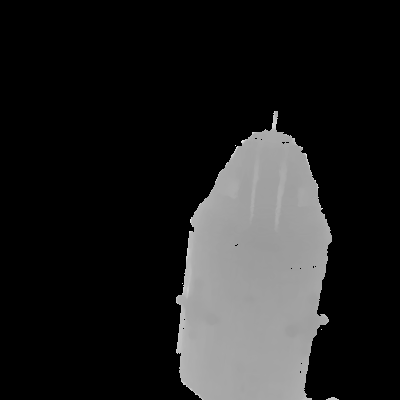}
&

\includegraphics[trim=0 0 0 -5, width=0.4\columnwidth]{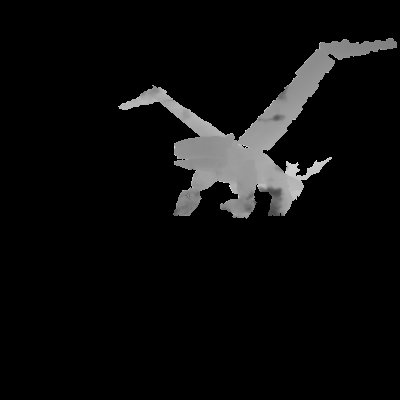}
\\

\end{tabular}
\caption{Depth qualitative results on the Bricks environment (figure best seen in zoom).}
\label{fig:more_qual_results_3_depth}
\end{figure*}

\begin{figure*}
\centering
\begin{tabular}{cc|c|c|c}
\rotatebox[origin=lt]{90}{\small \ \ \ \ \ \ \ \ \ Ground truth} &
\includegraphics[trim=0 0 0 -5, width=0.4\columnwidth]{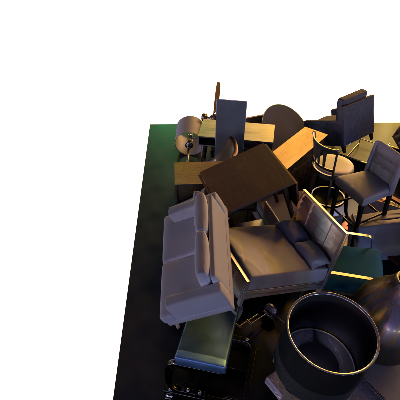}
&

\includegraphics[trim=0 0 0 -5, width=0.4\columnwidth]{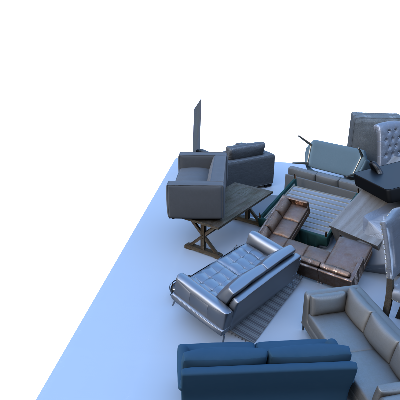}
&

\includegraphics[trim=0 0 0 -5, width=0.4\columnwidth]{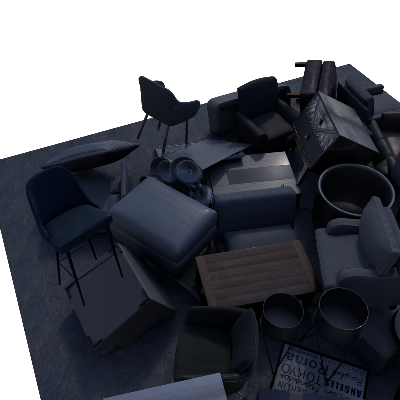}
&

\includegraphics[trim=0 0 0 -5, width=0.4\columnwidth]{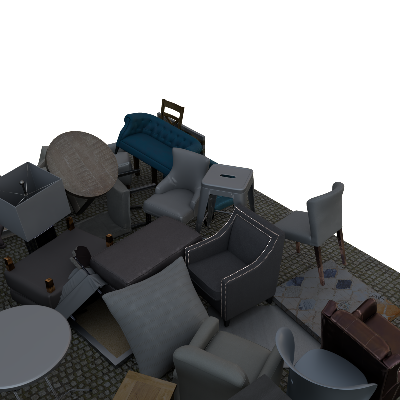}
\\

\rotatebox[origin=lt]{90}{\small \ \ \ \ \ \ \ \ \ \ \ \ \ \ SVLF} &
\includegraphics[trim=0 0 0 -5, width=0.4\columnwidth]{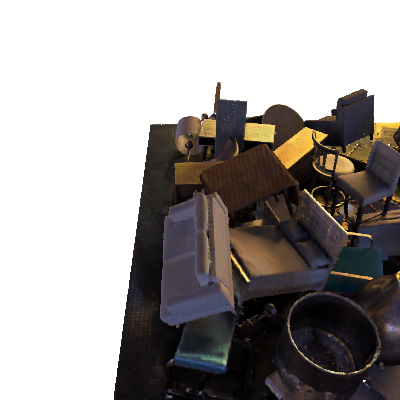}
&

\includegraphics[trim=0 0 0 -5, width=0.4\columnwidth]{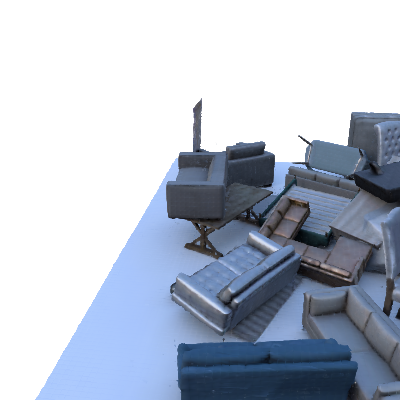}
&

\includegraphics[trim=0 0 0 -5, width=0.4\columnwidth]{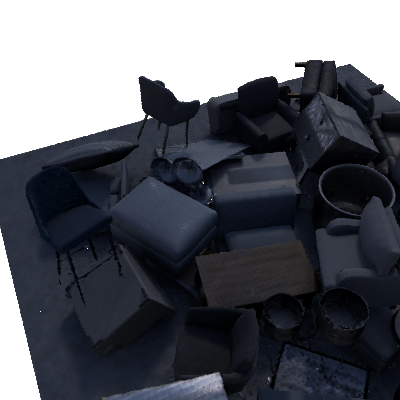}
&

\includegraphics[trim=0 0 0 -5, width=0.4\columnwidth]{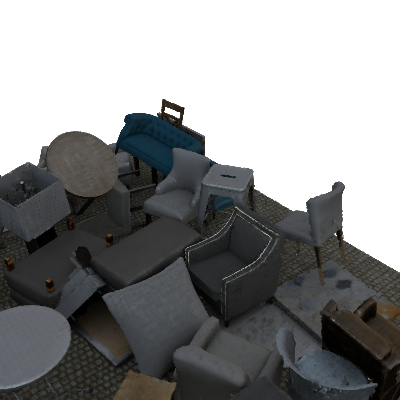}
\\

\rotatebox[origin=lt]{90}{\small \ \ \ \ \ \ \ \ \ \ Ins.-NGP} &
\includegraphics[trim=0 0 0 -5, width=0.4\columnwidth]{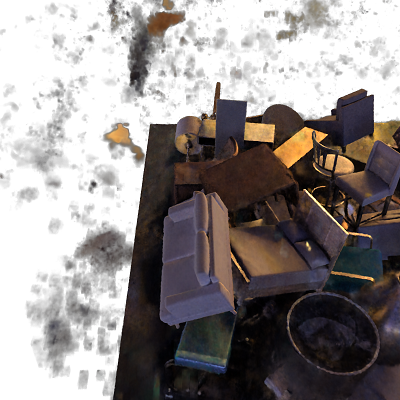}
&

\includegraphics[trim=0 0 0 -5, width=0.4\columnwidth]{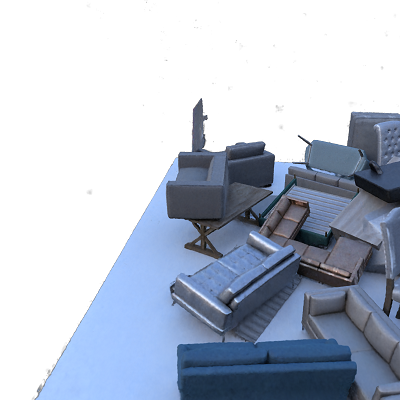}
&

\includegraphics[trim=0 0 0 -5, width=0.4\columnwidth]{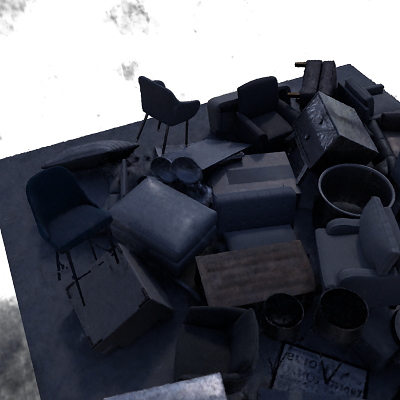}
&

\includegraphics[trim=0 0 0 -5, width=0.4\columnwidth]{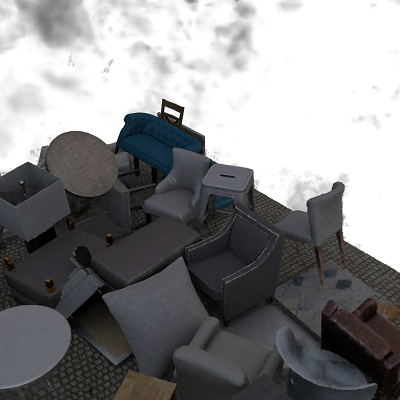}
\\

\rotatebox[origin=lt]{90}{\small \ \ \ \ \ \ \ \ \ mip-NeRF} &
\includegraphics[trim=0 0 0 -5, width=0.4\columnwidth]{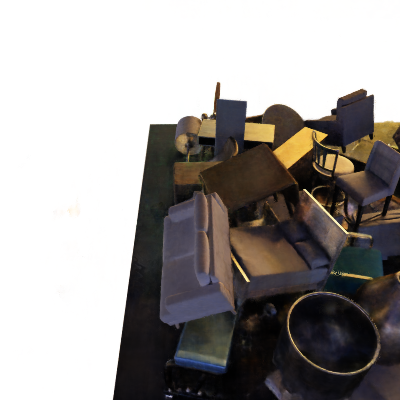}
&

\includegraphics[trim=0 0 0 -5, width=0.4\columnwidth]{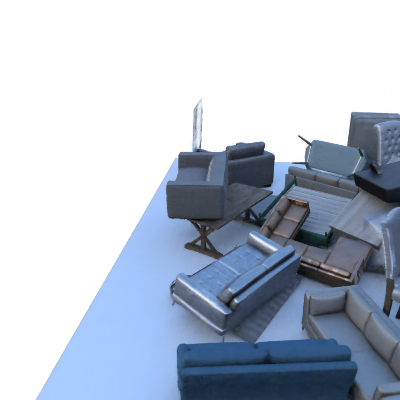}
&

\includegraphics[trim=0 0 0 -5, width=0.4\columnwidth]{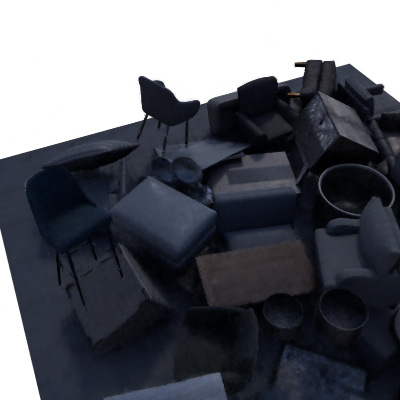}
&

\includegraphics[trim=0 0 0 -5, width=0.4\columnwidth]{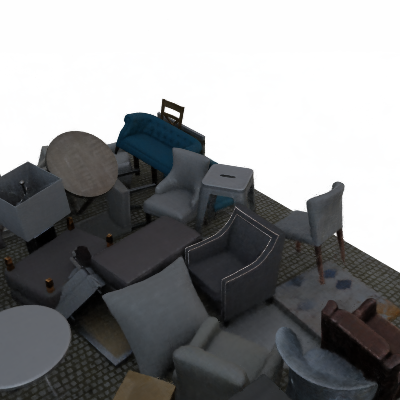}
\\

\rotatebox[origin=lt]{90}{\small \ \ \ \ \ \ \ \ \ \ \ \ \ \ NeRF} &
\includegraphics[trim=0 0 0 -5, width=0.4\columnwidth]{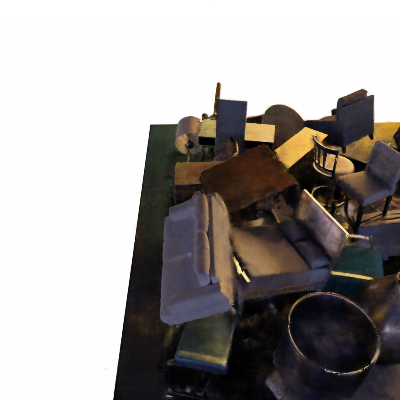}
&

\includegraphics[trim=0 0 0 -5, width=0.4\columnwidth]{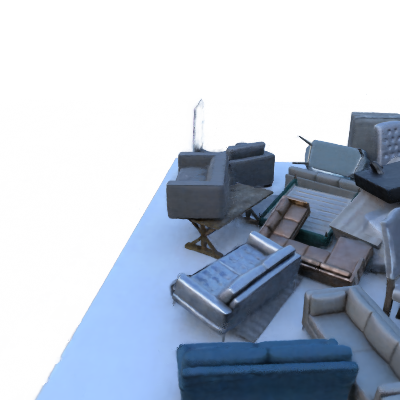}
&

\includegraphics[trim=0 0 0 -5, width=0.4\columnwidth]{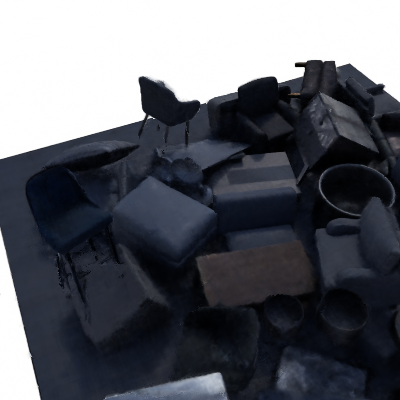}
&

\includegraphics[trim=0 0 0 -5, width=0.4\columnwidth]{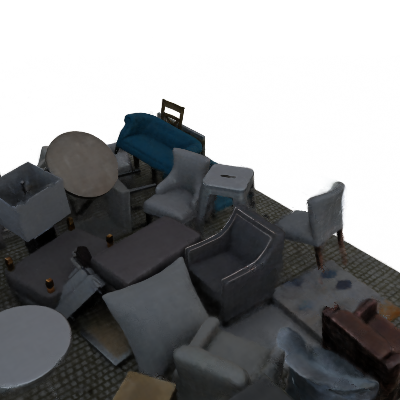}
\\

\end{tabular}
\caption{More qualitative results on the Amazon Berkeley environment (figure best seen in zoom).}
\label{fig:more_qual_results_4}
\end{figure*}

\begin{figure*}
\centering
\begin{tabular}{cc|c|c|c}
\rotatebox[origin=lt]{90}{\small \ \ \ \ \ \ \ \ \ Ground truth} &
\includegraphics[trim=0 0 0 -5, width=0.4\columnwidth]{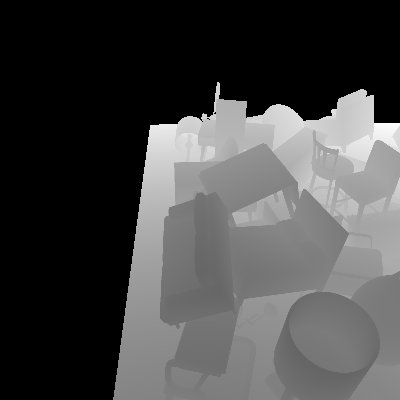}
&

\includegraphics[trim=0 0 0 -5, width=0.4\columnwidth]{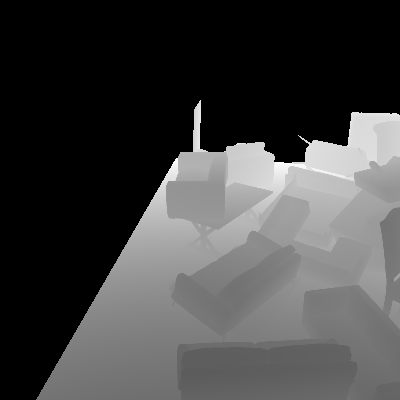}
&

\includegraphics[trim=0 0 0 -5, width=0.4\columnwidth]{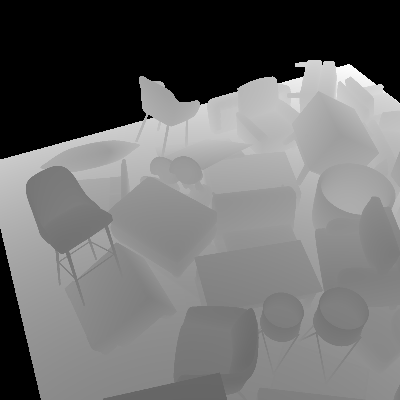}
&

\includegraphics[trim=0 0 0 -5, width=0.4\columnwidth]{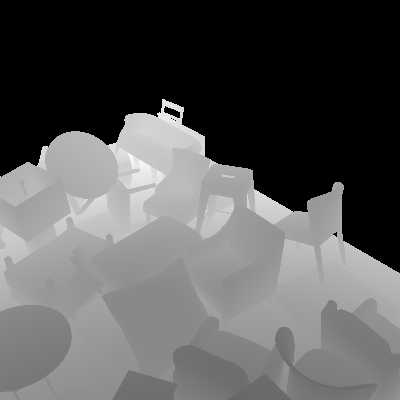}
\\

\rotatebox[origin=lt]{90}{\small \ \ \ \ \ \ \ \ \ \ \ \ \ \ SVLF} &
\includegraphics[trim=0 0 0 -5, width=0.4\columnwidth]{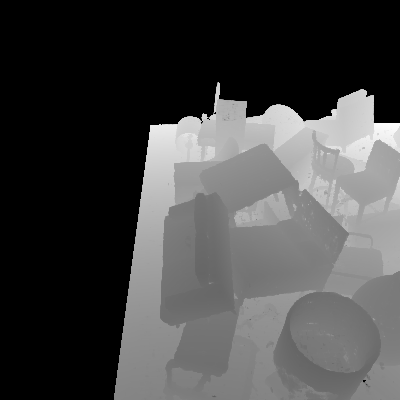}
&

\includegraphics[trim=0 0 0 -5, width=0.4\columnwidth]{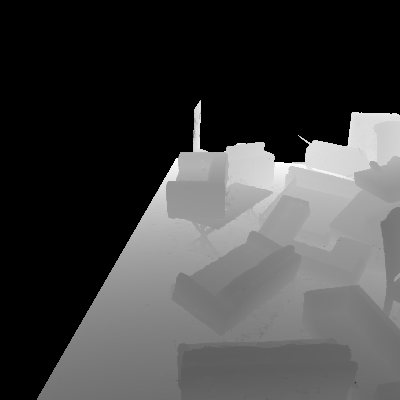}
&

\includegraphics[trim=0 0 0 -5, width=0.4\columnwidth]{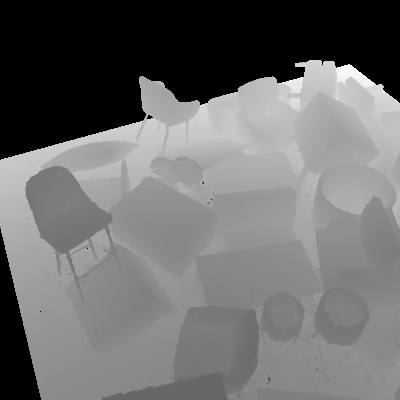}
&

\includegraphics[trim=0 0 0 -5, width=0.4\columnwidth]{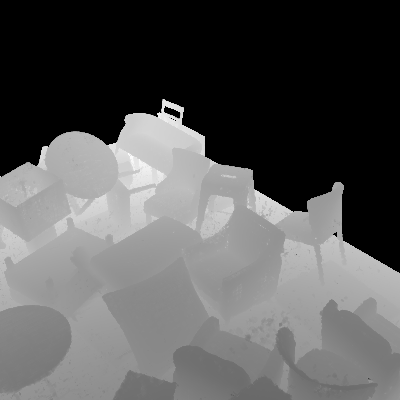}
\\

\rotatebox[origin=lt]{90}{\small \ \ \ \ \ \ \ \ \ \ Ins.-NGP} &
\includegraphics[trim=0 0 0 -5, width=0.4\columnwidth]{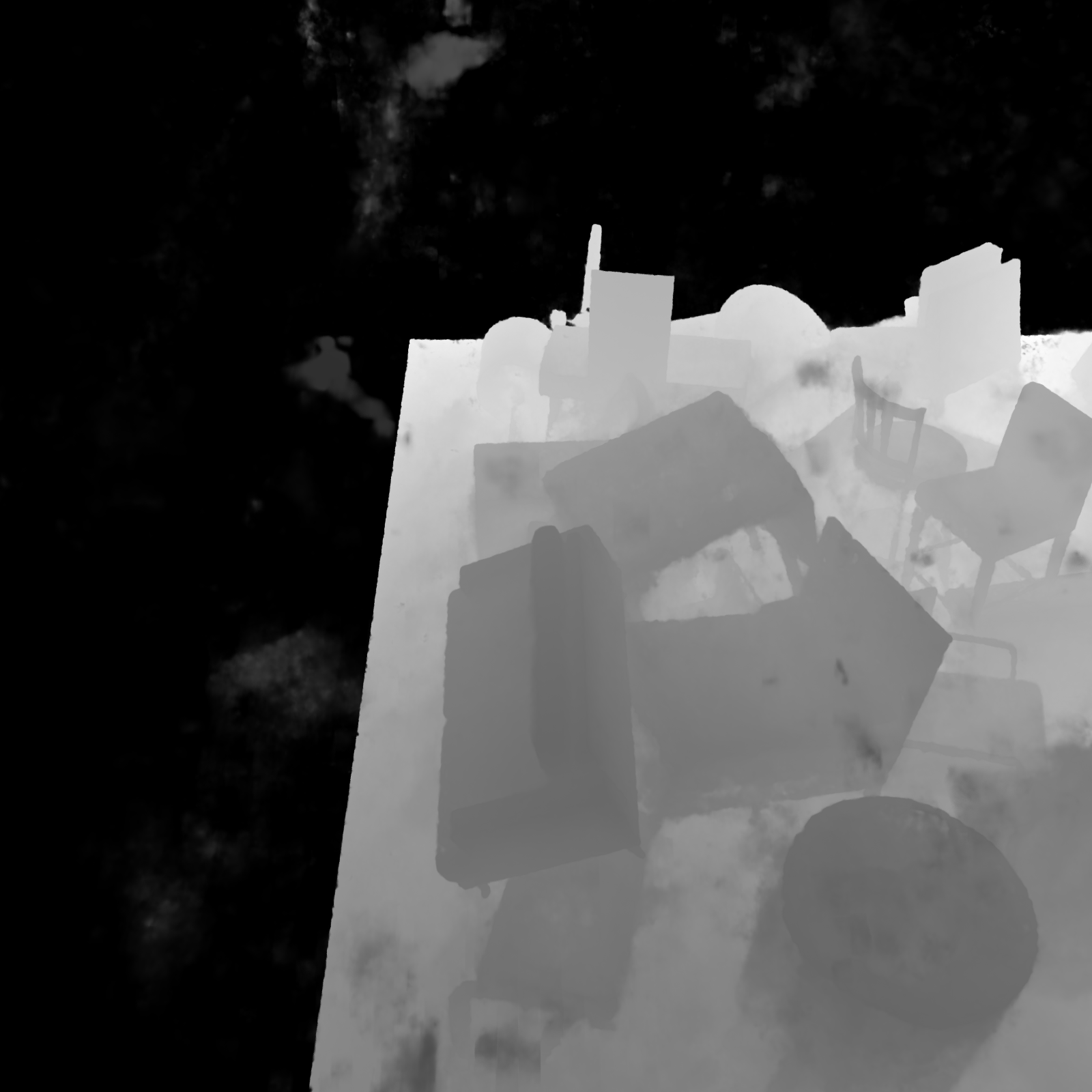}
&

\includegraphics[trim=0 0 0 -5, width=0.4\columnwidth]{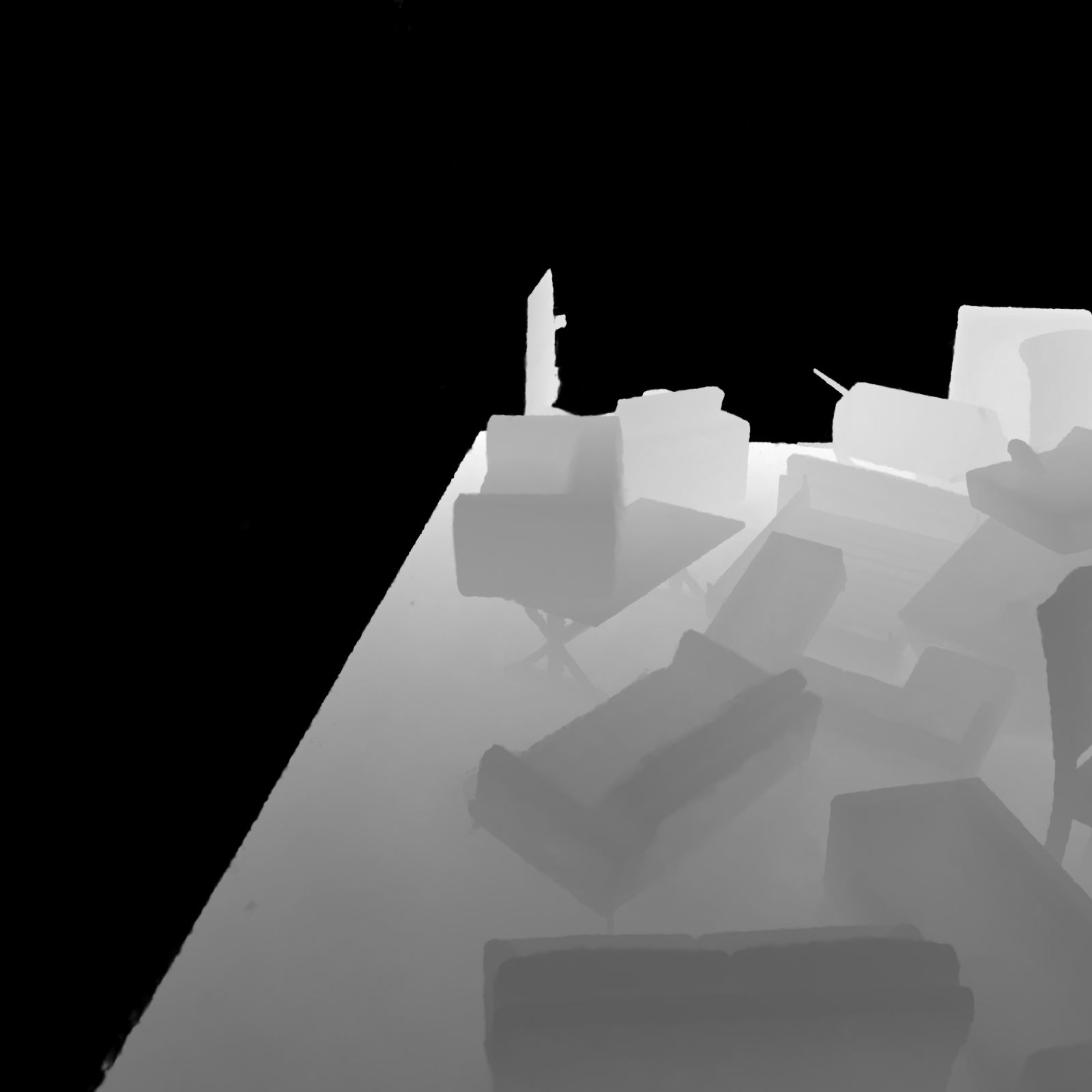}
&

\includegraphics[trim=0 0 0 -5, width=0.4\columnwidth]{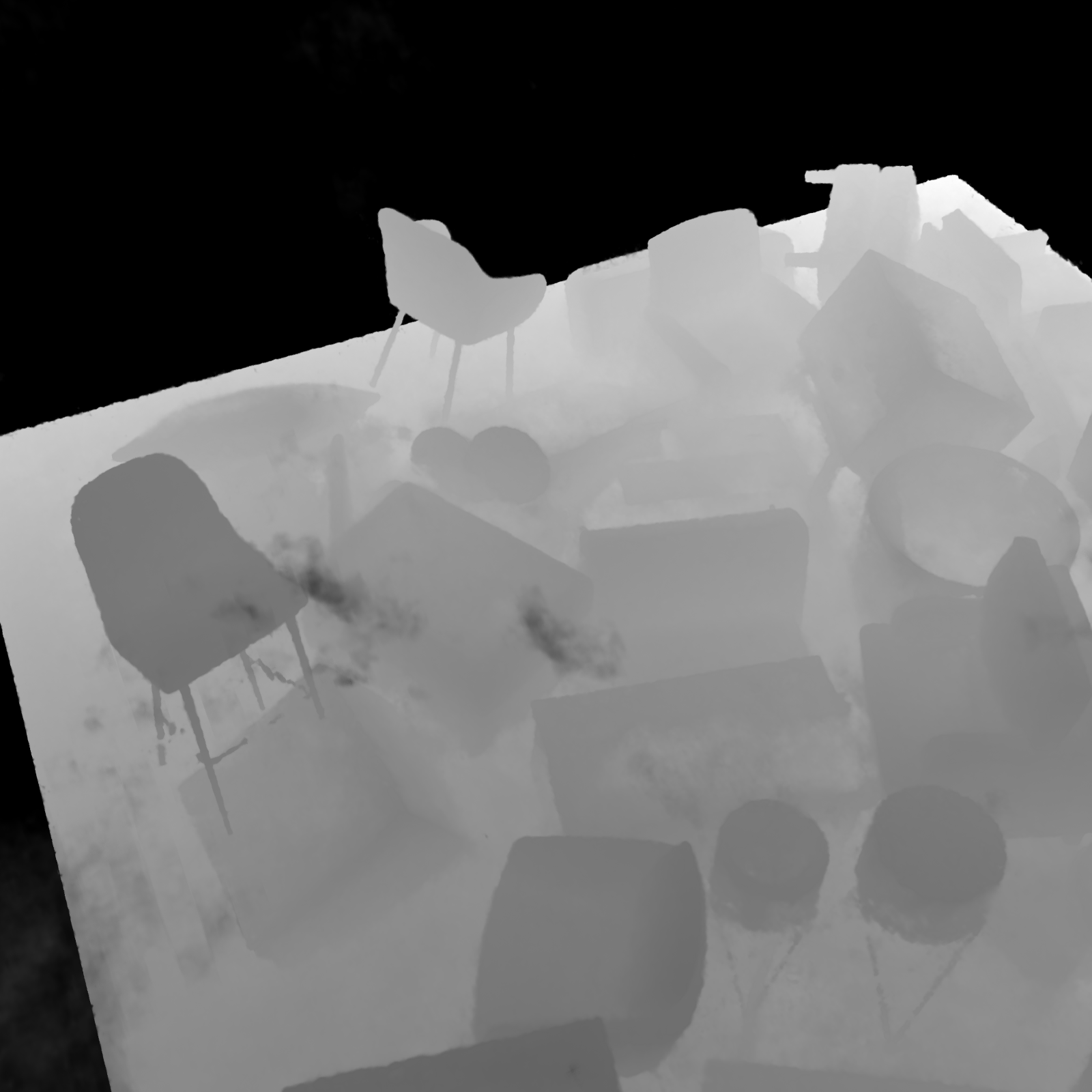}
&

\includegraphics[trim=0 0 0 -5, width=0.4\columnwidth]{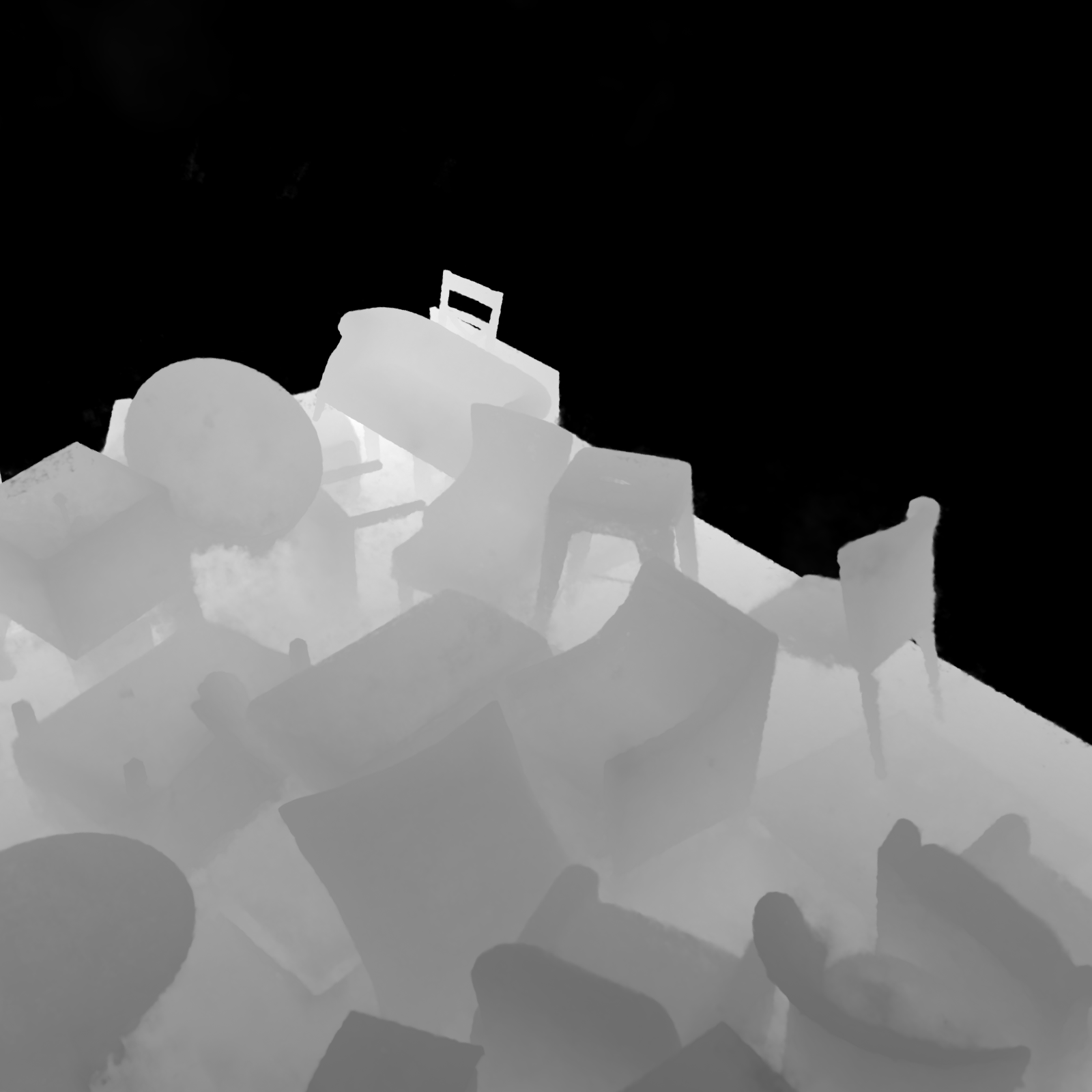}
\\

\rotatebox[origin=lt]{90}{\small \ \ \ \ \ \ \ \ \ mip-NeRF} &
\includegraphics[trim=0 0 0 -5, width=0.4\columnwidth]{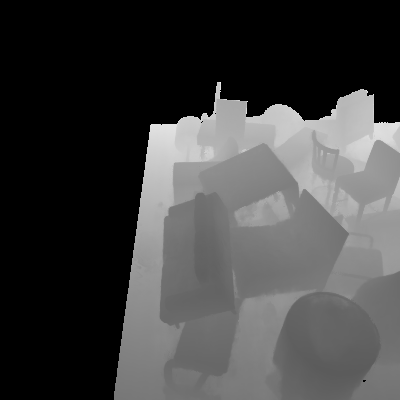}
&

\includegraphics[trim=0 0 0 -5, width=0.4\columnwidth]{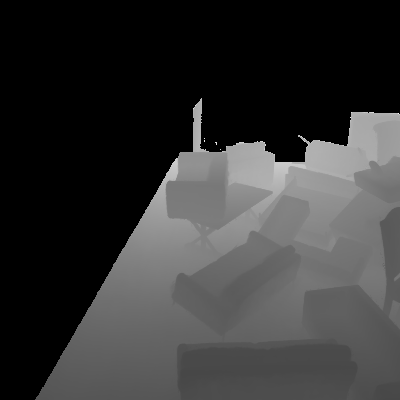}
&

\includegraphics[trim=0 0 0 -5, width=0.4\columnwidth]{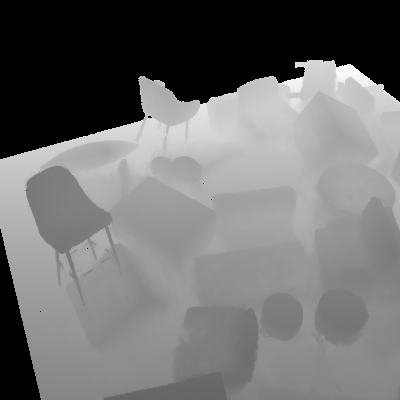}
&

\includegraphics[trim=0 0 0 -5, width=0.4\columnwidth]{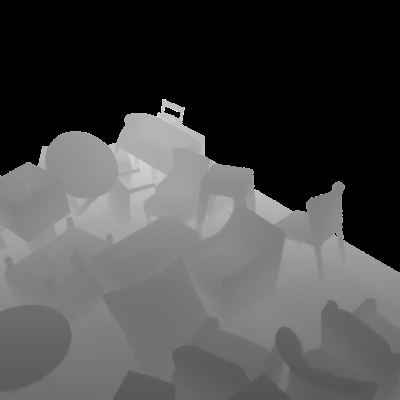}
\\

\rotatebox[origin=lt]{90}{\small \ \ \ \ \ \ \ \ \ \ \ \ \ \ NeRF} &
\includegraphics[trim=0 0 0 -5, width=0.4\columnwidth]{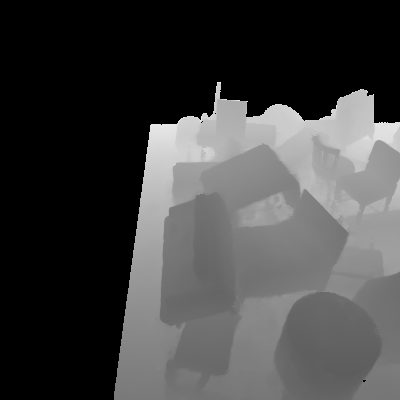}
&

\includegraphics[trim=0 0 0 -5, width=0.4\columnwidth]{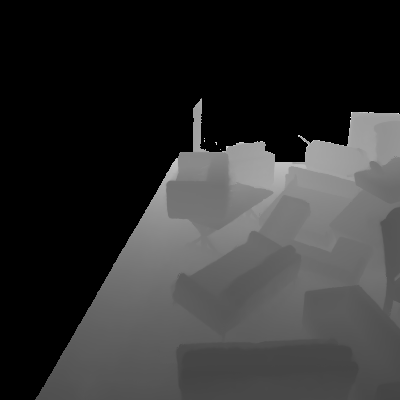}
&

\includegraphics[trim=0 0 0 -5, width=0.4\columnwidth]{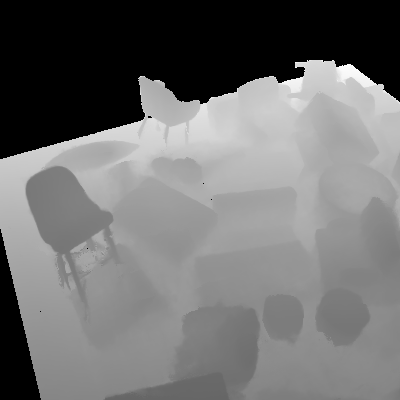}
&

\includegraphics[trim=0 0 0 -5, width=0.4\columnwidth]{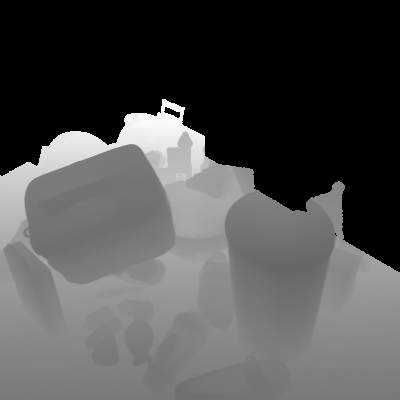}
\\

\end{tabular}
\caption{Depth qualitative results on the Amazon Berkeley environment (figure best seen in zoom).}
\label{fig:more_qual_results_4_depth}
\end{figure*}

We show more qualitative results on each of our four environments in Figures~\ref{fig:more_qual_results_1},~\ref{fig:more_qual_results_2},~\ref{fig:more_qual_results_3},~\ref{fig:more_qual_results_4},
and the corresponding depth maps in Figures~\ref{fig:more_qual_results_1_depth},~\ref{fig:more_qual_results_2_depth},~\ref{fig:more_qual_results_3_depth},~\ref{fig:more_qual_results_4_depth}.
SVLF trains with depth supervision, and thus constructs better depth maps compared to NeRF and mip-NeRF.
Instant-NGP gives the best results but suffers from floating artifacts which can also be seen in the recovered depth maps.
All baselines perform reasonably well on the Google Scanned Objects and the Bricks environments.
However, the performance of both NeRF~\cite{mildenhall2020nerf} and SVLF drops on the ABC environment due its challenging camera poses and lighting conditions.
And eventually, all baselines struggle on our most challenging environment (Amazon Berkeley).
The multi-scale training of mip-NeRF~\cite{barron2021mip} makes it more robust to random camera poses compared to the other baselines, especially on the Amazon Berkeley environment.
Instant-NGP suffers from floating artifacts especially with random camera poses (\eg, ABC and Amazon Berkeley environments).
But overall, Instant-NGP recovers fine details, renders sharp results, and achieves the best performance on our dataset.

\begin{figure*}
     \centering
     \begin{tabular}{cccc}

       \rotatebox[origin=lt]{90}{\Large \ \ \ \ \ \ \ \ \ \ \ \ \ \ \ \ \ \ \ SVLF} &
	     \includegraphics[width=0.31\textwidth]{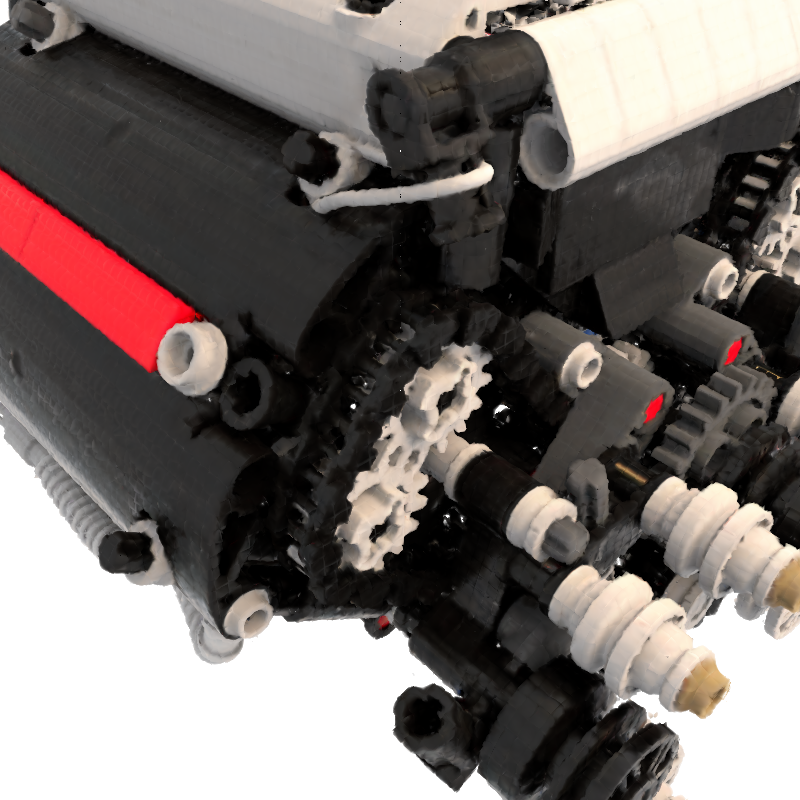} &
	     \includegraphics[width=0.31\textwidth]{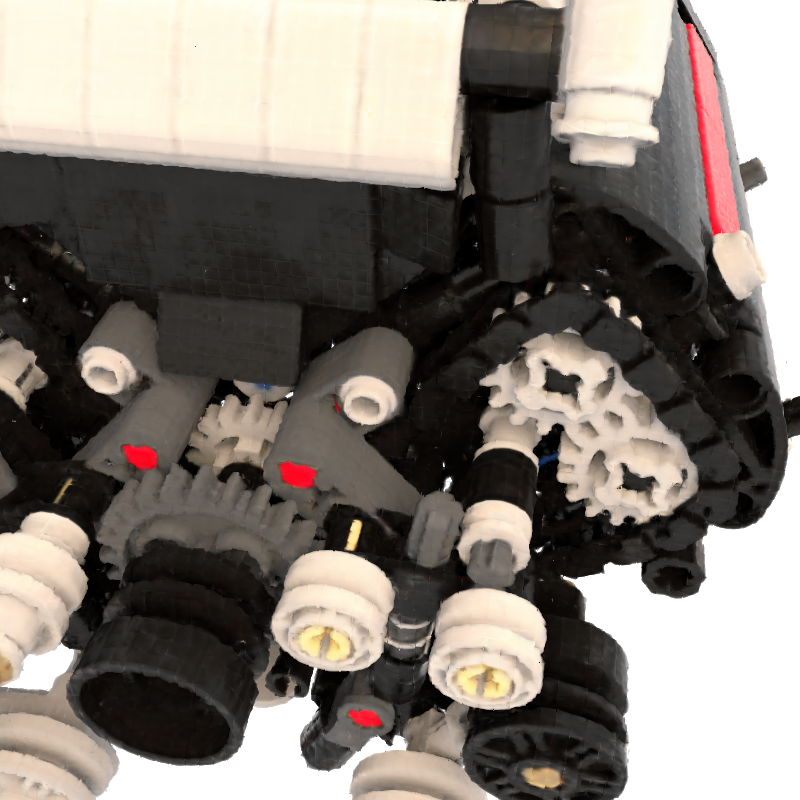} &
	     \includegraphics[width=0.31\textwidth]{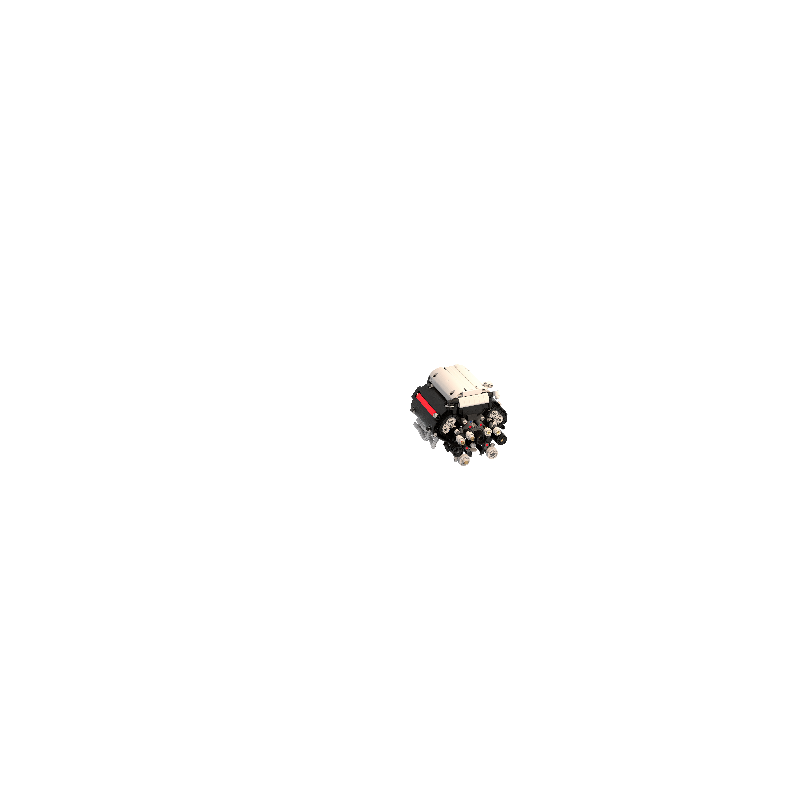}
       \\
       \rotatebox[origin=lt]{90}{\Large \ \ \ \ \ \ \ \ \ \ \ \ \ \ mip-NeRF} &
	     \includegraphics[width=0.31\textwidth]{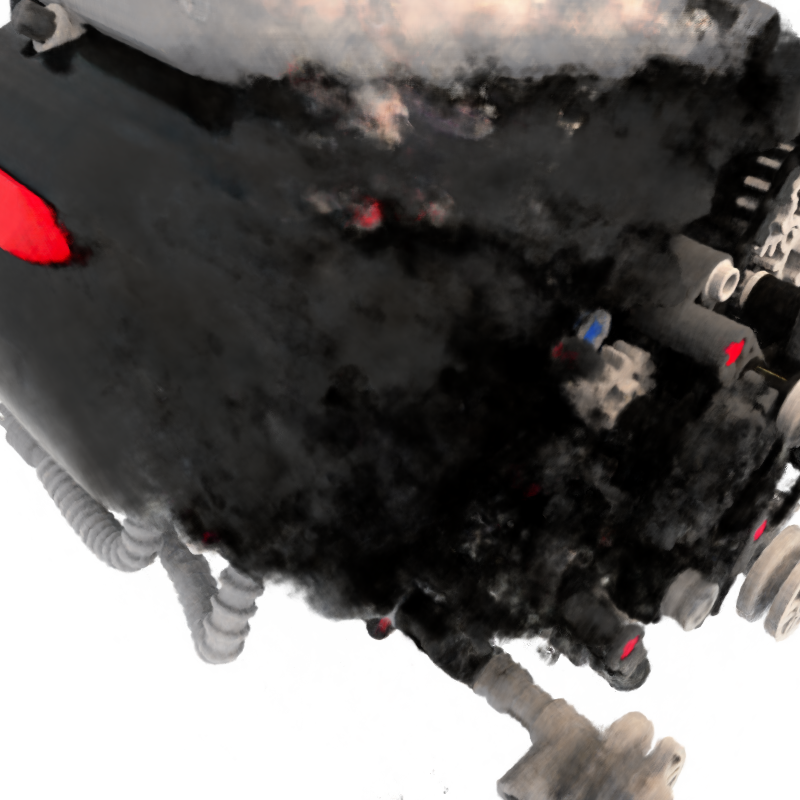} &
	     \includegraphics[width=0.31\textwidth]{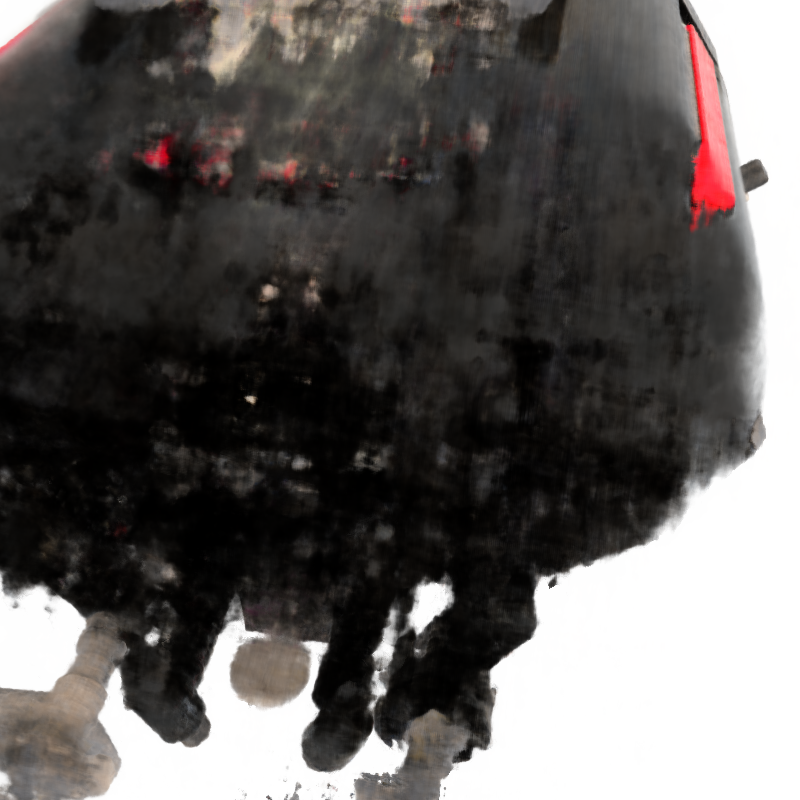} &
	     \includegraphics[width=0.31\textwidth]{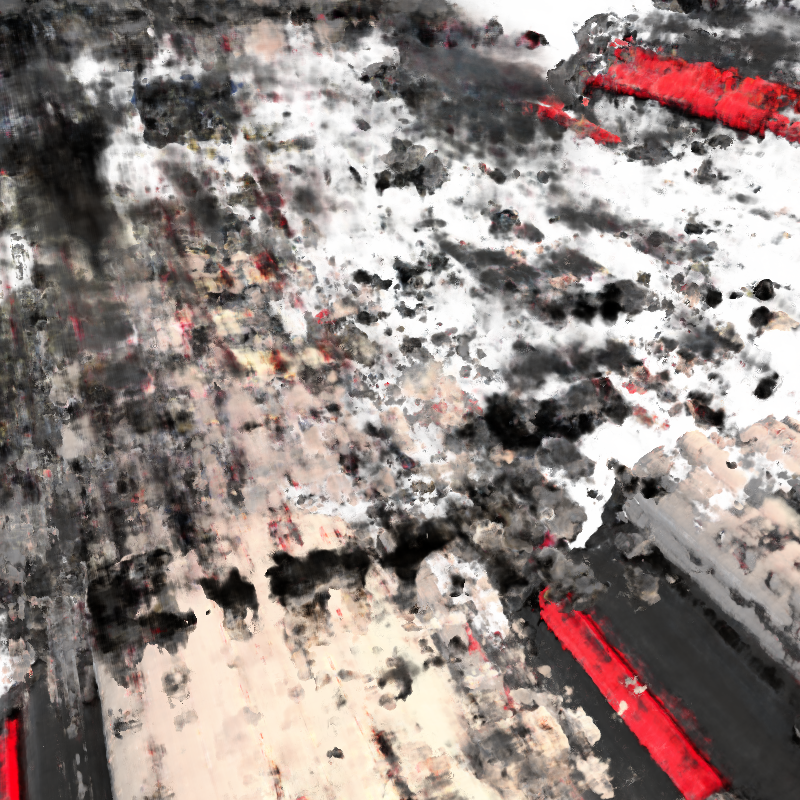}
       \\
       \rotatebox[origin=lt]{90}{\Large \ \ \ \ \ \ \ \ \ \ \ \ \ \ \ \ \ \ \ NeRF} &
	     \includegraphics[width=0.31\textwidth]{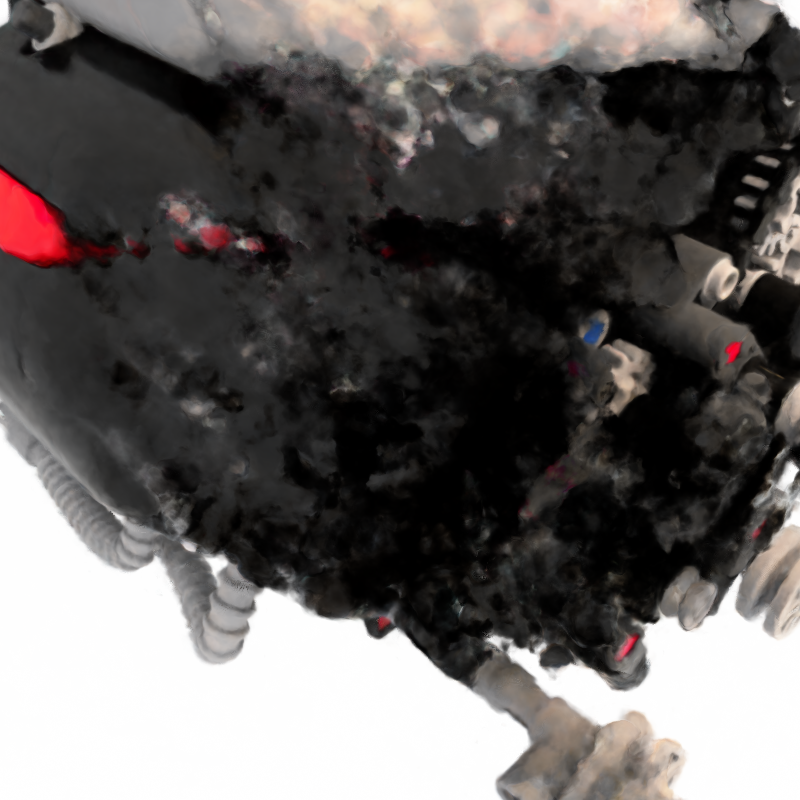} &
	     \includegraphics[width=0.31\textwidth]{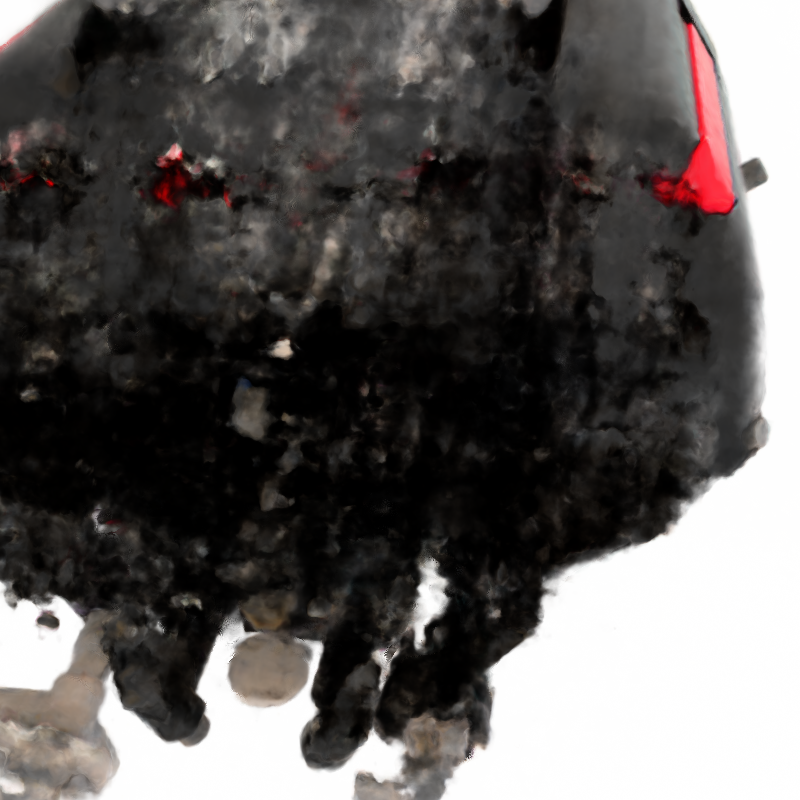} &
	     \includegraphics[width=0.31\textwidth]{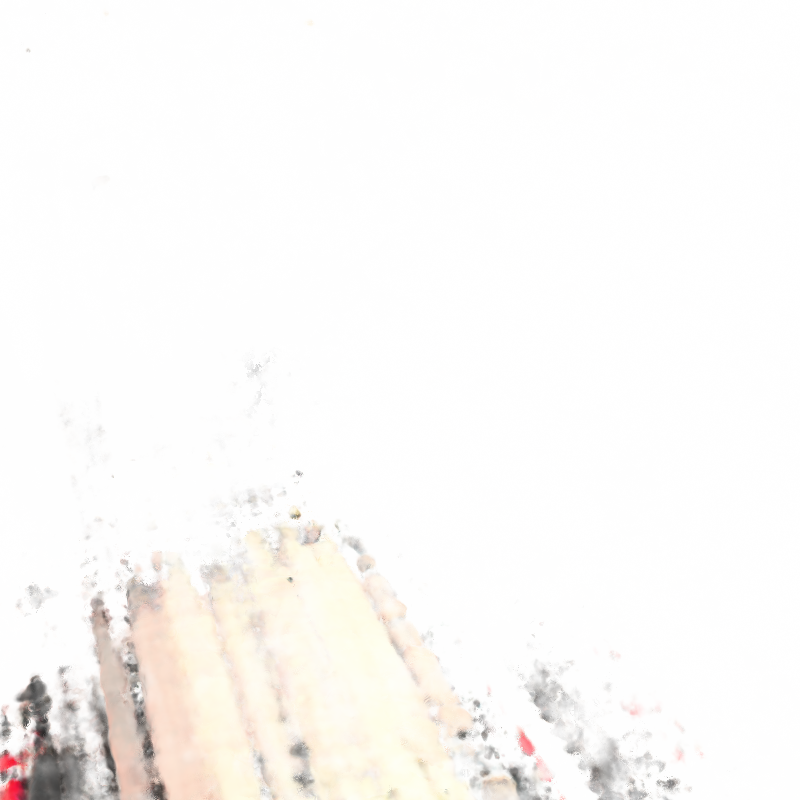}
     \end{tabular}
\vspace{-0.2cm}
\caption{The render planes parameterization of the baselines can lead to sub-optimal results when the camera moves freely in the space.}
\label{fig:render_planes}
\vspace{-0.4cm}
\end{figure*}

The voxel-based formulation of SVLF allows for a better representation of the scene that accommodates free-moving cameras.
This is because SVLF focuses on rays intersecting voxels, rather than where rays originate or end.
In contrast, the near and far render planes parametrization of NeRF and mip-NeRF poses a limitation.
On one hand, setting near and far planes to tightly bound the scene leads to better point sampling along the ray during training,
but allows for less flexibility for moving the camera at render time.
On the other hand, setting the near and far plane to loosely enclose the scene allows for more flexible camera movement at inference time,
but leads to poor training quality as NeRF and mip-NeRF are sensitive to how densely points are sampled along the ray.
Figure~\ref{fig:render_planes} highlights this limitation
by showing example results for when the camera is placed at very close (first two rows) or far (third row) distance from the scene.

\subsection{SVLF limitations}

In this section we discuss some of the limitations of the proposed SVLF approach.

\begin{figure*}
  \centering
  \begin{tabular}{ccc|cc|cc}
    \rotatebox{90}{\small Ground truth} &
    \includegraphics[width=0.14\linewidth]{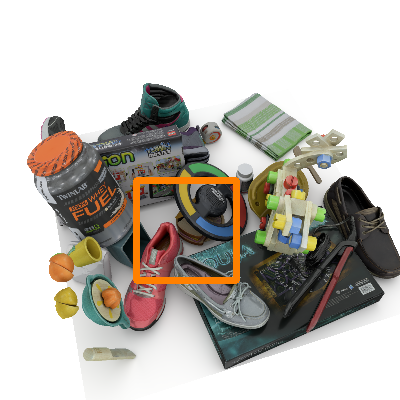} &
    \includegraphics[width=0.14\linewidth]{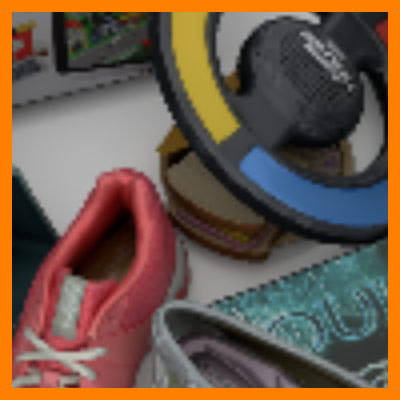} &
    \includegraphics[width=0.14\linewidth]{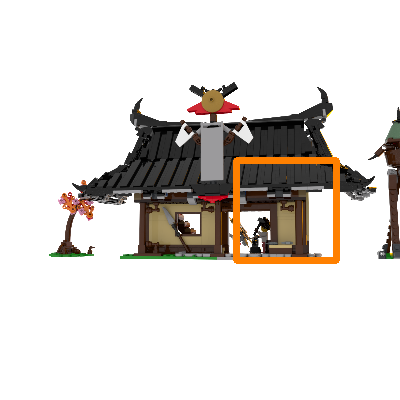} &
    \includegraphics[width=0.14\linewidth]{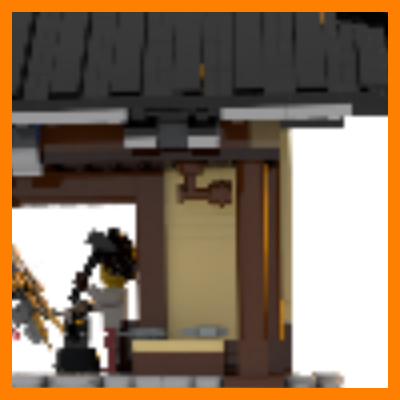} &
    \includegraphics[width=0.14\linewidth]{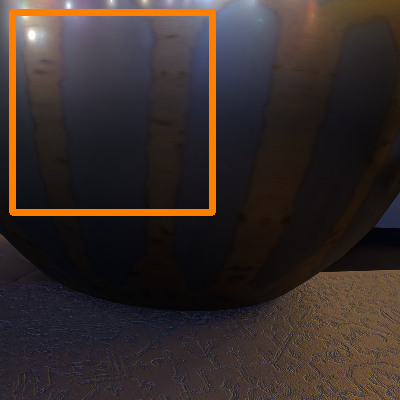} &
    \includegraphics[width=0.14\linewidth]{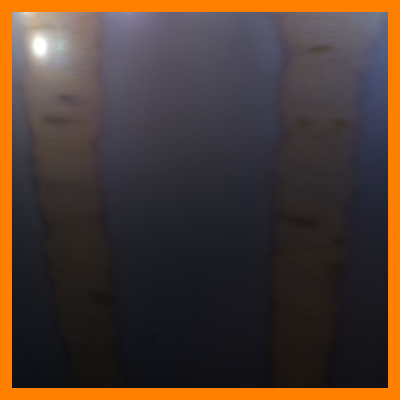}
    \\
    \rotatebox{90}{\small  \ \ \ \ \ \ SVLF} &
    \includegraphics[width=0.14\linewidth]{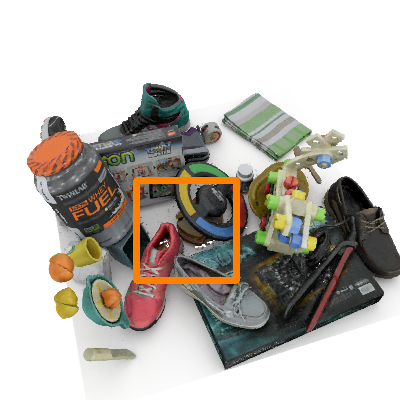} &
    \includegraphics[width=0.14\linewidth]{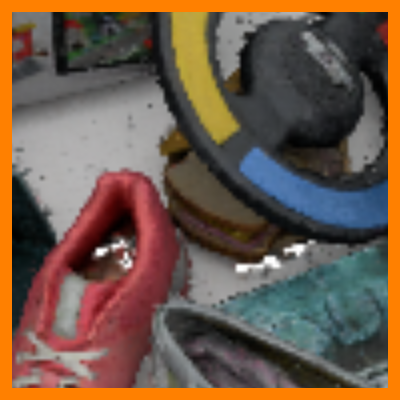} &
    \includegraphics[width=0.14\linewidth]{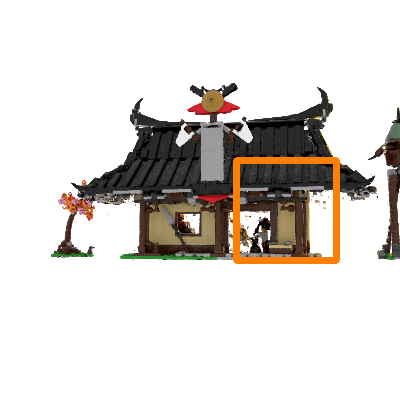} &
    \includegraphics[width=0.14\linewidth]{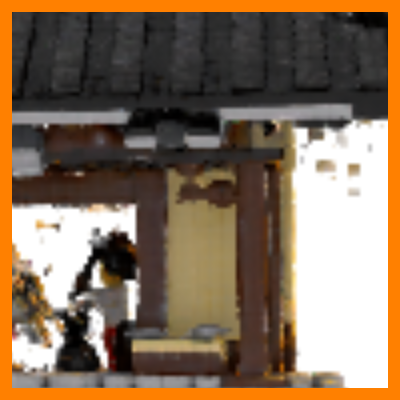} &
    \includegraphics[width=0.14\linewidth]{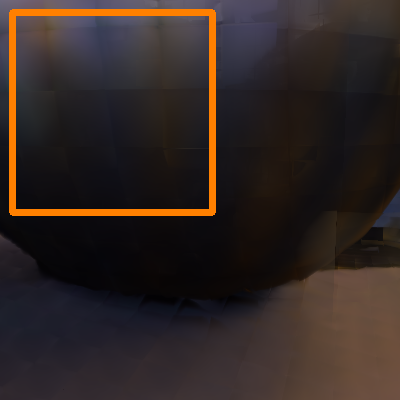} &
    \includegraphics[width=0.14\linewidth]{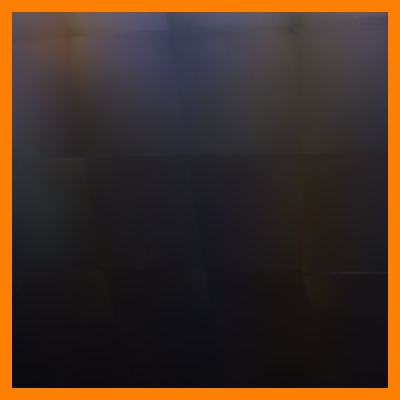}
    \\

    &
    \multicolumn{2}{c}{(a)} &
    \multicolumn{2}{c}{(b)} &
    \multicolumn{2}{c}{(c)}
    \\

  \end{tabular}
  \caption{Example limitations of SVLF.
           (a) and (b) show occasional floater/hole artifacts due to sub-optimal training of the optical thickness $\tau$.
  				 (c) shows seams between voxel boundaries when the camera is too close to the scene, and a lack of specular highlights (figure best seen in zoom).
          }
  \label{fig:limitations}
\end{figure*}

\paragraph{Voxel resolution.}
SVLF uses a sparse octree to represent a voxel grid of size $128^3$.
While this voxel resolution allows for high quality results over the test set of our scenes,
it can produce voxel boundary artifacts if the camera zooms in too close to the scene.
Specifically, the seams between voxels become more visible as the camera zooms in (see Figure~\ref{fig:limitations}).

\paragraph{Sub-optimal training of optical thickness.}
SVLF learns voxel-based functions.
Although using a shared decoder as well as the tri-linear interpolation between voxel features provide a degree of continuity between adjacent voxels,
voxel features could still suffer from a degree of independence.
This especially happens to the optical thickness features, $\zdensity$, where the tri-linear interpolations happens at the intersection points, $(x_1, x_2)$, between the ray and the voxel, rather than at the estimated surface point $\hat{x}_s$.
This could lead to a lack of global context between voxels along the ray, thus resulting in errors in the learned optical thickness $\tau$.
Figure~\ref{fig:limitations} show example failures where incorrect predictions of the optical thickness $\tau$ lead to either floaters or holes in the rendered image.
Note the holes inside and beside the red shoe in Figure~\ref{fig:limitations}-(a),
and random floaters in Figure~\ref{fig:limitations}-(a), (b).

\paragraph{Translucent surfaces.}
SVLF evaluates the ray color within a voxel at a single point location (the estimated surface hit $\hat{x}_s$).
Therefore, we assume solid surfaces, and bootstrap the training using surface rendering (Sec.~{\color{red} 4.4} of the main paper).
While SVLF utilizes a volumetric training stage to learn non-binary optical thicknesses/densities, the support for translucent surfaces remains limited 
due to sampling a single point within each voxel.

\paragraph{View-dependent effects.}
We observe that SVLF does not capture view dependent effects very well (\eg, specular highlight on the vase in Figure~\ref{fig:limitations}-column 3.)
We hypothesize this is due to the absence of an explicit view direction input to our network.
Adding the view direction as an extra input besides the estimated surface hit $\hat{x}_s$ could help mitigate the problem.

%% file: main.bbl
\begin{thebibliography}{10}\itemsep=-1pt

\bibitem{barron2021mip}
Jonathan~T. Barron, Ben Mildenhall, Matthew Tancik, Peter Hedman, Ricardo
  Martin-Brualla, and Pratul~P. Srinivasan.
\newblock Mip-{NeRF}: A multiscale representation for anti-aliasing neural
  radiance fields.
\newblock {\em arXiv preprint arXiv:2103.13415}, 2021.

\bibitem{buehler2001unstructured}
Chris Buehler, Michael Bosse, Leonard McMillan, Steven Gortler, and Michael
  Cohen.
\newblock Unstructured {Lumigraph} rendering.
\newblock In {\em Proceedings of the 28th Annual Conference on Computer
  Graphics and Interactive Techniques}, pages 425--432, 2001.

\bibitem{burley2015extending}
Brent Burley.
\newblock Extending the {Disney BRDF} to a {BSDF} with integrated subsurface
  scattering.
\newblock {\em Physically Based Shading in Theory and Practice SIGGRAPH
  Course}, 2015.

\bibitem{chanmonteiro2020pi-GAN}
Eric Chan, Marco Monteiro, Petr Kellnhofer, Jiajun Wu, and Gordon Wetzstein.
\newblock pi-{GAN}: Periodic implicit generative adversarial networks for
  {3D}-aware image synthesis.
\newblock In {\em arXiv}, 2020.

\bibitem{chaurasia2013depth}
Gaurav Chaurasia, Sylvain Duchene, Olga Sorkine-Hornung, and George Drettakis.
\newblock Depth synthesis and local warps for plausible image-based navigation.
\newblock {\em ACM Transactions on Graphics (TOG)}, 32(3):1--12, 2013.

\bibitem{chaurasia2011silhouette}
Gaurav Chaurasia, Olga Sorkine, and George Drettakis.
\newblock Silhouette-aware warping for image-based rendering.
\newblock {\em Computer Graphics Forum}, 30(4):1223--1232, 2011.

\bibitem{chen2021mvsnerf}
Anpei Chen, Zexiang Xu, Fuqiang Zhao, Xiaoshuai Zhang, Fanbo Xiang, Jingyi Yu,
  and Hao Su.
\newblock {MVSNeRF}: Fast generalizable radiance field reconstruction from
  multi-view stereo, 2021.

\bibitem{chen2019learning}
Zhiqin Chen and Hao Zhang.
\newblock Learning implicit fields for generative shape modeling.
\newblock In {\em Proceedings of the IEEE/CVF Conference on Computer Vision and
  Pattern Recognition}, pages 5939--5948, 2019.

\bibitem{choi2019extreme}
Inchang Choi, Orazio Gallo, Alejandro Troccoli, Min~H. Kim, and Jan Kautz.
\newblock Extreme view synthesis.
\newblock In {\em Proceedings of the IEEE/CVF International Conference on
  Computer Vision}, pages 7781--7790, 2019.

\bibitem{collins2021abo}
Jasmine Collins, Shubham Goel, Achleshwar Luthra, Leon Xu, Kenan Deng, Xi
  Zhang, Tomas~F Yago~Vicente, Himanshu Arora, Thomas Dideriksen, Matthieu
  Guillaumin, and Jitendra Malik.
\newblock Abo: Dataset and benchmarks for real-world 3d object understanding.
\newblock {\em arXiv preprint arXiv:2110.06199}, 2021.

\bibitem{coumans2019:pybullet}
Erwin Coumans and Yunfei Bai.
\newblock {PyBullet}, a {P}ython module for physics simulation for games,
  robotics and machine learning.
\newblock \url{http://pybullet.org}, 2016--2019.

\bibitem{unity2020perception}
Adam Crespi, Cesar Romero, Srinivas Annambhotla, Jonathan Hogins, and Alex
  Thaman.
\newblock Unity perception, 2020.
\newblock \url{https://blogs.unity3d.com/2020/06/10/}.

\bibitem{debevec1996modeling}
Paul Debevec, Camillo Taylor, and Jitendra Malik.
\newblock Modeling and rendering architecture from photographs: A hybrid
  geometry-and image-based approach.
\newblock In {\em Proceedings of the 23rd Annual Conference on Computer
  Graphics and Interactive Techniques}, pages 11--20, 1996.

\bibitem{dellaert2020neural}
Frank Dellaert and Lin Yen-Chen.
\newblock Neural volume rendering: {NeRF} and beyond.
\newblock {\em arXiv preprint arXiv:2101.05204}, 2020.

\bibitem{denninger2019blenderproc}
Maximilian Denninger, Martin Sundermeyer, Dominik Winkelbauer, Youssef Zidan,
  Dmitry Olefir, Mohamad Elbadrawy, Ahsan Lodhi, and Harinandan Katam.
\newblock {BlenderProc}.
\newblock {\em arXiv preprint arXiv:1911.01911}, 2019.

\bibitem{dosovitskiy2015iccv:flychairs}
Alexey Dosovitskiy, Philipp Fischer, Eddy Ilg, Philip H{\"a}usser, Canaer
  Haz{\i}rba{\c{s}}, Vladimir Golkov, Patrick van~der Smagt, Daniel Cremers,
  and Thomas Brox.
\newblock {FlowNet}: {L}earning optical flow with convolutional networks.
\newblock In {\em ICCV}, 2015.

\bibitem{du2021nerflow}
Yilun Du, Yinan Zhang, Hong-Xing Yu, Joshua~B. Tenenbaum, and Jiajun Wu.
\newblock Neural radiance flow for {4D} view synthesis and video processing.
\newblock In {\em Proceedings of the IEEE/CVF International Conference on
  Computer Vision}, 2021.

\bibitem{flynn2019deepview}
John Flynn, Michael Broxton, Paul Debevec, Matthew DuVall, Graham Fyffe, Ryan
  Overbeck, Noah Snavely, and Richard Tucker.
\newblock Deepview: View synthesis with learned gradient descent.
\newblock In {\em Proceedings of the IEEE/CVF Conference on Computer Vision and
  Pattern Recognition}, pages 2367--2376, 2019.

\bibitem{Gafni_2021_CVPR}
Guy Gafni, Justus Thies, Michael Zollh{\"o}fer, and Matthias Nie{\ss}ner.
\newblock Dynamic neural radiance fields for monocular {4D} facial avatar
  reconstruction.
\newblock In {\em Proceedings of the IEEE/CVF Conference on Computer Vision and
  Pattern Recognition (CVPR)}, pages 8649--8658, June 2021.

\bibitem{gaidon2016cvpr:vkitti}
Adrien Gaidon, Qiao Wang, Yohann Cabon, and Eleonora Vig.
\newblock Virtual worlds as proxy for multi-object tracking analysis.
\newblock In {\em CVPR}, 2016.

\bibitem{Gao-portraitnerf}
Chen Gao, Yichang Shih, Wei-Sheng Lai, Chia-Kai Liang, and Jia-Bin Huang.
\newblock Portrait neural radiance fields from a single image.
\newblock {\em arXiv preprint arXiv:2012.05903}, 2020.

\bibitem{GoogleScanned}
GoogleResearch.
\newblock {Google Scanned Objects}.
\newblock In {\em Open Robotics}, 2021.

\bibitem{gortler1996lumigraph}
Steven Gortler, Radek Grzeszczuk, Richard Szeliski, and Michael Cohen.
\newblock The {Lumigraph}.
\newblock In {\em SIGGRAPH’96: Proceedings of the 23rd Annual Conference on
  Computer Graphics and Interactive Techniques}. ACM Press, New York, NY, USA,
  1996.

\bibitem{Gu_2017_CVPR}
Jinwei Gu, Xiaodong Yang, Shalini De~Mello, and Jan Kautz.
\newblock Dynamic facial analysis: From {B}ayesian filtering to recurrent
  neural network.
\newblock In {\em Proceedings of the IEEE Conference on Computer Vision and
  Pattern Recognition (CVPR)}, July 2017.

\bibitem{handa2015arx:sn}
Ankur Handa, Viorica P\u{a}tr\u{a}ucean, Vijay Badrinarayanan, Simon Stent, and
  Roberto Cipolla.
\newblock {SceneNet}: Understanding real world indoor scenes with synthetic
  data.
\newblock In {\em arXiv 1511.07041}, 2015.

\bibitem{robotpose_etfa2019_cheind}
Christoph Heindl, Sebastian Zambal, and Josef Scharinger.
\newblock Learning to predict robot keypoints using artificially generated
  images.
\newblock In {\em 24th IEEE International Conference on Emerging Technologies
  and Factory Automation (ETFA)}, pages 1536--1539, 2019.

\bibitem{henzler2020learning}
Philipp Henzler, Niloy Mitra, and Tobias Ritschel.
\newblock Learning a neural {3D} texture space from {2D} exemplars.
\newblock In {\em Proceedings of the IEEE/CVF Conference on Computer Vision and
  Pattern Recognition}, pages 8356--8364, 2020.

\bibitem{henzler2021unsupervised}
Philipp Henzler, Jeremy Reizenstein, Patrick Labatut, Roman Shapovalov, Tobias
  Ritschel, Andrea Vedaldi, and David Novotny.
\newblock Unsupervised learning of {3D} object categories from videos in the
  wild.
\newblock In {\em Proceedings of the IEEE/CVF Conference on Computer Vision and
  Pattern Recognition}, pages 4700--4709, 2021.

\bibitem{IchnowskiAvigal2021DexNeRF}
Jeffrey Ichnowski, Yahav Avigal, Justin Kerr, and Ken Goldberg.
\newblock {Dex-NeRF}: Using a neural radiance field to grasp transparent
  objects.
\newblock In {\em Conference on Robot Learning (CoRL)}, 2020.

\bibitem{iraci2013blender}
Bernardo Iraci.
\newblock {\em Blender {Cycles}: Lighting and Rendering Cookbook}.
\newblock Packt Publishing Ltd, 2013.

\bibitem{Kaolin}
Krishna~Murthy Jatavallabhula, Edward Smith, Jean-Francois Lafleche,
  Clement~Fuji Tsang, Artem Rozantsev, Wenzheng Chen, Tommy Xiang, Rev
  Lebaredian, and Sanja Fidler.
\newblock Kaolin: A {PyTorch} library for accelerating {3D} deep learning
  research.
\newblock {\em arXiv:1911.05063}, 2019.

\bibitem{jensen2014large}
Rasmus Jensen, Anders Dahl, George Vogiatzis, Engin Tola, and Henrik Aan{\ae}s.
\newblock Large scale multi-view stereopsis evaluation.
\newblock In {\em Proceedings of the IEEE Conference on Computer Vision and
  Pattern Recognition}, pages 406--413, 2014.

\bibitem{kar2017learning}
Abhishek Kar, Christian H{\"a}ne, and Jitendra Malik.
\newblock Learning a multi-view stereo machine.
\newblock {\em NeurIPS}, 2017.

\bibitem{adamOptimizer}
Diederik~P. Kingma and Jimmy Ba.
\newblock Adam: {A} method for stochastic optimization.
\newblock {\em CoRR}, abs/1412.6980, 2014.

\bibitem{Knapitsch2017}
Arno Knapitsch, Jaesik Park, Qian-Yi Zhou, and Vladlen Koltun.
\newblock Tanks and temples: Benchmarking large-scale scene reconstruction.
\newblock {\em ACM Transactions on Graphics}, 36(4), 2017.

\bibitem{Koch_2019_CVPR}
Sebastian Koch, Albert Matveev, Zhongshi Jiang, Francis Williams, Alexey
  Artemov, Evgeny Burnaev, Marc Alexa, Denis Zorin, and Daniele Panozzo.
\newblock {ABC}: A big {CAD} model dataset for geometric deep learning.
\newblock In {\em The IEEE Conference on Computer Vision and Pattern
  Recognition (CVPR)}, June 2019.

\bibitem{kolve2017ai2}
Eric Kolve, Roozbeh Mottaghi, Winson Han, Eli VanderBilt, Luca Weihs, Alvaro
  Herrasti, Daniel Gordon, Yuke Zhu, Abhinav Gupta, and Ali Farhadi.
\newblock {AI2-THOR}: {A}n interactive {3D} environment for visual {AI}.
\newblock {\em arXiv preprint arXiv:1712.05474}, 2017.

\bibitem{labbe2020}
Yann Labb{\'{e}}, Justin Carpentier, Mathieu Aubry, and Josef Sivic.
\newblock {CosyPose}: Consistent multi-view multi-object {6D} pose estimation.
\newblock In {\em Proceedings of the European Conference on Computer Vision
  (ECCV)}, 2020.

\bibitem{li2021neural}
Zhengqi Li, Simon Niklaus, Noah Snavely, and Oliver Wang.
\newblock Neural scene flow fields for space-time view synthesis of dynamic
  scenes.
\newblock In {\em Proceedings of the IEEE/CVF Conference on Computer Vision and
  Pattern Recognition}, pages 6498--6508, 2021.

\bibitem{li2020crowdsampling}
Zhengqi Li, Wenqi Xian, Abe Davis, and Noah Snavely.
\newblock Crowdsampling the plenoptic function.
\newblock In {\em European Conference on Computer Vision}, pages 178--196.
  Springer, 2020.

\bibitem{liu2020neural}
Lingjie Liu, Jiatao Gu, Kyaw~Zaw Lin, Tat-Seng Chua, and Christian Theobalt.
\newblock Neural sparse voxel fields.
\newblock {\em NeurIPS}, 2020.

\bibitem{lombardi2019neural}
Stephen Lombardi, Tomas Simon, Jason Saragih, Gabriel Schwartz, Andreas
  Lehrmann, and Yaser Sheikh.
\newblock Neural volumes: Learning dynamic renderable volumes from images.
\newblock {\em ACM Trans. Graph.}, 38(4):65:1--65:14, July 2019.

\bibitem{majercik2018ray}
Alexander Majercik, Cyril Crassin, Peter Shirley, and Morgan McGuire.
\newblock A ray-box intersection algorithm and efficient dynamic voxel
  rendering.
\newblock {\em Journal of Computer Graphics Techniques Vol}, 7(3):66--81, 2018.

\bibitem{martinbrualla2020nerfw}
Ricardo Martin-Brualla, Noha Radwan, Mehdi S.~M. Sajjadi, Jonathan~T. Barron,
  Alexey Dosovitskiy, and Daniel Duckworth.
\newblock {NeRF} in the wild: Neural radiance fields for unconstrained photo
  collections.
\newblock In {\em CVPR}, 2021.

\bibitem{mayer2016cvpr:flythings}
Nikolaus Mayer, Eddy Ilg, Philip Hausser, Philipp Fischer, Daniel Cremers,
  Alexey Dosovitskiy, and Thomas Brox.
\newblock A large dataset to train convolutional networks for disparity,
  optical flow, and scene flow estimation.
\newblock In {\em CVPR}, 2016.

\bibitem{merrill2017cub}
Duane Merrill.
\newblock {CUB}: A library of warp-wide, block-wide, and device-wide {GPU}
  parallel primitives, 2017.
\newblock \url{https://nvlabs.github.io/cub/}.

\bibitem{Occupancy_Networks}
Lars Mescheder, Michael Oechsle, Michael Niemeyer, Sebastian Nowozin, and
  Andreas Geiger.
\newblock Occupancy networks: Learning {3D} reconstruction in function space.
\newblock In {\em Proceedings IEEE Conf. on Computer Vision and Pattern
  Recognition (CVPR)}, 2019.

\bibitem{mildenhall2019llff}
Ben Mildenhall, Pratul~P. Srinivasan, Rodrigo Ortiz-Cayon, Nima~Khademi
  Kalantari, Ravi Ramamoorthi, Ren Ng, and Abhishek Kar.
\newblock Local light field fusion: Practical view synthesis with prescriptive
  sampling guidelines.
\newblock {\em ACM Transactions on Graphics (TOG)}, 2019.

\bibitem{mildenhall2020nerf}
Ben Mildenhall, Pratul~P. Srinivasan, Matthew Tancik, Jonathan~T. Barron, Ravi
  Ramamoorthi, and Ren Ng.
\newblock {NeRF}: Representing scenes as neural radiance fields for view
  synthesis.
\newblock {\em arXiv preprint arXiv:2003.08934}, 2020.

\bibitem{morrical2021nvisii}
Nathan Morrical, Jonathan Tremblay, Yunzhi Lin, Stephen Tyree, Stan Birchfield,
  Valerio Pascucci, and Ingo Wald.
\newblock {NViSII}: A scriptable tool for photorealistic image generation.
\newblock In {\em ICLR Workshop on Synthetic Data Generation}, May 2021.

\bibitem{mueller2022instant}
Thomas M\"uller, Alex Evans, Christoph Schied, and Alexander Keller.
\newblock Instant neural graphics primitives with a multiresolution hash
  encoding.
\newblock {\em ACM Transactions on Graphics (SIGGRAPH)}, July 2022.

\bibitem{neff2021donerf}
Thomas Neff, Pascal Stadlbauer, Mathias Parger, Andreas Kurz, Joerg~H. Mueller,
  Chakravarty R.~Alla Chaitanya, Anton~S. Kaplanyan, and Markus Steinberger.
\newblock {DONeRF: Towards Real-Time Rendering of Compact Neural Radiance
  Fields using Depth Oracle Networks}.
\newblock {\em Computer Graphics Forum}, 40(4), 2021.

\bibitem{novak2018monte}
Jan Nov{\'a}k, Iliyan Georgiev, Johannes Hanika, and Wojciech Jarosz.
\newblock Monte {Carlo} methods for volumetric light transport simulation.
\newblock {\em Computer Graphics Forum}, 37(2):551--576, 2018.

\bibitem{Oechsle2021ICCV}
Michael Oechsle, Songyou Peng, and Andreas Geiger.
\newblock {UNISURF}: Unifying neural implicit surfaces and radiance fields for
  multi-view reconstruction.
\newblock In {\em International Conference on Computer Vision (ICCV)}, 2021.

\bibitem{Park_2019_CVPR}
Jeong~Joon Park, Peter Florence, Julian Straub, Richard Newcombe, and Steven
  Lovegrove.
\newblock {DeepSDF}: Learning continuous signed distance functions for shape
  representation.
\newblock In {\em The IEEE Conference on Computer Vision and Pattern
  Recognition (CVPR)}, June 2019.

\bibitem{park2021nerfies}
Keunhong Park, Utkarsh Sinha, Jonathan~T. Barron, Sofien Bouaziz, Dan~B.
  Goldman, Steven~M. Seitz, and Ricardo Martin-Brualla.
\newblock Nerfies: Deformable neural radiance fields.
\newblock {\em ICCV}, 2021.

\bibitem{optix}
Steven Parker, James Bigler, Andreas Dietrich, Heiko Friedrich, Jared Hoberock,
  David Luebke, David McAllister, Morgan McGuire, Keith Morley, and Austin
  Robison.
\newblock {OptiX}: {A} {General} {Purpose} {Ray} {Tracing} {Engine}.
\newblock {\em ACM Transactions on Graphics (Proceedings of ACM SIGGRAPH)},
  2010.

\bibitem{penner2017soft}
Eric Penner and Li Zhang.
\newblock Soft {3D} reconstruction for view synthesis.
\newblock {\em ACM Transactions on Graphics (TOG)}, 36(6):1--11, 2017.

\bibitem{pumarola2020d}
Albert Pumarola, Enric Corona, Gerard Pons-Moll, and Francesc Moreno-Noguer.
\newblock {D-NeRF: Neural Radiance Fields for Dynamic Scenes}.
\newblock In {\em Proceedings of the IEEE/CVF Conference on Computer Vision and
  Pattern Recognition}, 2020.

\bibitem{rebain2021derf}
Daniel Rebain, Wei Jiang, Soroosh Yazdani, Ke Li, Kwang~Moo Yi, and Andrea
  Tagliasacchi.
\newblock {DeRF}: Decomposed radiance fields.
\newblock In {\em Proceedings of the IEEE/CVF Conference on Computer Vision and
  Pattern Recognition}, pages 14153--14161, 2021.

\bibitem{reizenstein2021common}
Jeremy Reizenstein, Roman Shapovalov, Philipp Henzler, Luca Sbordone, Patrick
  Labatut, and David Novotny.
\newblock Common objects in {3D}: Large-scale learning and evaluation of
  real-life {3D} category reconstruction.
\newblock In {\em Proceedings of the IEEE/CVF International Conference on
  Computer Vision}, pages 10901--10911, 2021.

\bibitem{Richter_2016_ECCV}
Stephan~R. Richter, Vibhav Vineet, Stefan Roth, and Vladlen Koltun.
\newblock Playing for data: {G}round truth from computer games.
\newblock In {\em European Conference on Computer Vision (ECCV)}, pages
  102--118, 2016.

\bibitem{riegler2020free}
Gernot Riegler and Vladlen Koltun.
\newblock Free view synthesis.
\newblock In {\em European Conference on Computer Vision}, pages 623--640.
  Springer, 2020.

\bibitem{ron2020expohd}
Roey Ron and Gil Elbaz.
\newblock {EXPO-HD}: Exact object perception using high distraction synthetic
  data.
\newblock In {\em arXiv:2007.14354}, 2020.

\bibitem{ros2016cvpr:syn}
German Ros, Laura Sellart, Joanna Materzynska, David Vazquez, and Antonio
  Lopez.
\newblock The {SYNTHIA} dataset: {A} large collection of synthetic images for
  semantic segmentation of urban scenes.
\newblock In {\em CVPR}, 2016.

\bibitem{sajjan2019cleargrasp}
Shreeyak~S. Sajjan, Matthew Moore, Mike Pan, Ganesh Nagaraja, Johnny Lee, Andy
  Zeng, and Shuran Song.
\newblock {ClearGrasp}: {3D} shape estimation of transparent objects for
  manipulation.
\newblock {\em arXiv preprint arXiv:1910.02550}, 2019.

\bibitem{Schwarz2020NEURIPS}
Katja Schwarz, Yiyi Liao, Michael Niemeyer, and Andreas Geiger.
\newblock {GRAF}: Generative radiance fields for {3D}-aware image synthesis.
\newblock In {\em Advances in Neural Information Processing Systems (NeurIPS)},
  2020.

\bibitem{sinha2009piecewise}
Sudipta Sinha, Drew Steedly, and Rick Szeliski.
\newblock Piecewise planar stereo for image-based rendering.
\newblock In {\em 2009 International Conference on Computer Vision}, 2009.

\bibitem{sitzmann2019srns}
Vincent Sitzmann, Michael Zollh{\"o}fer, and Gordon Wetzstein.
\newblock Scene representation networks: Continuous {3D}-structure-aware neural
  scene representations.
\newblock In {\em Advances in Neural Information Processing Systems}, 2019.

\bibitem{srinivasan2019pushing}
Pratul~P. Srinivasan, Richard Tucker, Jonathan~T. Barron, Ravi Ramamoorthi, Ren
  Ng, and Noah Snavely.
\newblock Pushing the boundaries of view extrapolation with multiplane images.
\newblock In {\em Proceedings of the IEEE/CVF Conference on Computer Vision and
  Pattern Recognition}, pages 175--184, 2019.

\bibitem{takikawa2021nglod}
Towaki Takikawa, Joey Litalien, Kangxue Yin, Karsten Kreis, Charles Loop, Derek
  Nowrouzezahrai, Alec Jacobson, Morgan McGuire, and Sanja Fidler.
\newblock Neural geometric level of detail: Real-time rendering with implicit
  {3D} shapes.
\newblock In {\em Proceedings of the IEEE/CVF Conference on Computer Vision and
  Pattern Recognition (CVPR)}, 2021.

\bibitem{tancik2021learned}
Matthew Tancik, Ben Mildenhall, Terrance Wang, Divi Schmidt, Pratul~P.
  Srinivasan, Jonathan~T. Barron, and Ren Ng.
\newblock Learned initializations for optimizing coordinate-based neural
  representations.
\newblock In {\em Proceedings of the IEEE/CVF Conference on Computer Vision and
  Pattern Recognition}, pages 2846--2855, 2021.

\bibitem{tewari2021advances}
Ayush Tewari, Justus Thies, Ohad Fried, Vincent Sitzmann, Kalyan Sunkavalli,
  Ricardo Martin-Brualla, Tomas Simon, Jason Saragih, Matthias Nie{\ss}ner,
  Rohit~K. Pandey, Sean Fanello, Gordon Wetzstein, Jun-Yan Zhu, Christian
  Theobalt, Maneesh Agrawala, Eli Shechtman, Dan~B. Goldman, and Michael
  Zollh\"ofer.
\newblock Advances in neural rendering.
\newblock In {\em ACM SIGGRAPH 2021 Courses}, pages 1--320, 2021.

\bibitem{to2018ndds}
Thang To, Jonathan Tremblay, Duncan McKay, Yukie Yamaguchi, Kirby Leung, Adrian
  Balanon, Jia Cheng, and Stan Birchfield.
\newblock {NDDS}: {NVIDIA} deep learning dataset synthesizer, 2018.
\newblock \url{https://github.com/NVIDIA/Dataset_Synthesizer}.

\bibitem{tremblay2018wad:car}
Jonathan Tremblay, Aayush Prakash, David Acuna, Mark Brophy, Varun Jampani, Cem
  Anil, Thang To, Eric Cameracci, Shaad Boochoon, and Stan Birchfield.
\newblock Training deep networks with synthetic data: Bridging the reality gap
  by domain randomization.
\newblock In {\em CVPR Workshop on Autonomous Driving (WAD)}, 2018.

\bibitem{tremblay2018arx:fat}
Jonathan Tremblay, Thang To, and Stan Birchfield.
\newblock Falling things: {A} synthetic dataset for {3D} object detection and
  pose estimation.
\newblock In {\em CVPR Workshop on Real World Challenges and New Benchmarks for
  Deep Learning in Robotic Vision}, 2018.

\bibitem{tremblay:corl2018}
Jonathan Tremblay, Thang To, Balakumar Sundaralingam, Yu Xiang, Dieter Fox, and
  Stan Birchfield.
\newblock Deep object pose estimation for semantic robotic grasping of
  household objects.
\newblock In {\em CoRL}, 2018.

\bibitem{grf2020}
Alex Trevithick and Bo Yang.
\newblock {GRF}: Learning a general radiance field for {3D} scene
  representation and rendering.
\newblock In {\em arXiv:2010.04595}, 2020.

\bibitem{tulsiani2017multi}
Shubham Tulsiani, Tinghui Zhou, Alexei~A. Efros, and Jitendra Malik.
\newblock Multi-view supervision for single-view reconstruction via
  differentiable ray consistency.
\newblock In {\em Proceedings of the IEEE conference on computer vision and
  pattern recognition}, pages 2626--2634, 2017.

\bibitem{wang2021neus}
Peng Wang, Lingjie Liu, Yuan Liu, Christian Theobalt, Taku Komura, and Wenping
  Wang.
\newblock Neus: Learning neural implicit surfaces by volume rendering for
  multi-view reconstruction.
\newblock {\em NeurIPS}, 2021.

\bibitem{wang2021ibrnet}
Qianqian Wang, Zhicheng Wang, Kyle Genova, Pratul~P. Srinivasan, Howard Zhou,
  Jonathan~T. Barron, Ricardo Martin-Brualla, Noah Snavely, and Thomas
  Funkhouser.
\newblock Ibrnet: Learning multi-view image-based rendering.
\newblock In {\em CVPR}, 2021.

\bibitem{wood2001iccv:fakeit}
Erroll Wood, Tadas Baltru\v{s}aitis, Charlie Hewitt, Sebastian Dziadzio,
  Matthew Johnson, Virginia Estellers, Thomas~J. Cashman, and Jamie Shotton.
\newblock Fake it till you make it: Face analysis in the wild using synthetic
  data alone.
\newblock In {\em ICCV}, 2021.

\bibitem{wood2015_iccv}
Erroll Wood, Tadas Baltru\v{s}aitis, Xucong Zhang, Yusuke Sugano, Peter
  Robinson, and Andreas Bulling.
\newblock Rendering of eyes for eye-shape registration and gaze estimation.
\newblock In {\em Proc. of the IEEE International Conference on Computer Vision
  (ICCV)}, 2015.

\bibitem{xian2021space}
Wenqi Xian, Jia-Bin Huang, Johannes Kopf, and Changil Kim.
\newblock Space-time neural irradiance fields for free-viewpoint video.
\newblock In {\em Proceedings of the IEEE/CVF Conference on Computer Vision and
  Pattern Recognition (CVPR)}, pages 9421--9431, 2021.

\bibitem{xiang2020sapien}
Fanbo Xiang, Yuzhe Qin, Kaichun Mo, Yikuan Xia, Hao Zhu, Fangchen Liu, Minghua
  Liu, Hanxiao Jiang, Yifu Yuan, He Wang, et~al.
\newblock {SAPIEN}: A simulated part-based interactive environment.
\newblock In {\em Proceedings of the IEEE/CVF Conference on Computer Vision and
  Pattern Recognition}, pages 11097--11107, 2020.

\bibitem{yang2021robi}
Jun Yang, Yizhou Gao, Dong Li, and Steven~L Waslander.
\newblock Robi: A multi-view dataset for reflective objects in robotic
  bin-picking.
\newblock {\em arXiv preprint arXiv:2105.04112}, 2021.

\bibitem{yao2020blendedmvs}
Yao Yao, Zixin Luo, Shiwei Li, Jingyang Zhang, Yufan Ren, Lei Zhou, Tian Fang,
  and Long Quan.
\newblock {BlendedMVS}: A large-scale dataset for generalized multi-view stereo
  networks.
\newblock {\em Computer Vision and Pattern Recognition (CVPR)}, 2020.

\bibitem{yu2021plenoctrees}
Alex Yu, Ruilong Li, Matthew Tancik, Hao Li, Ren Ng, and Angjoo Kanazawa.
\newblock {PlenOctrees} for real-time rendering of neural radiance fields.
\newblock In {\em ICCV}, 2021.

\bibitem{yu2020pixelnerf}
Alex Yu, Vickie Ye, Matthew Tancik, and Angjoo Kanazawa.
\newblock {pixelNeRF}: Neural radiance fields from one or few images, 2020.

\bibitem{kaizhang2020}
Kai Zhang, Gernot Riegler, Noah Snavely, and Vladlen Koltun.
\newblock {NeRF++}: Analyzing and improving neural radiance fields.
\newblock {\em arXiv:2010.07492}, 2020.

\bibitem{zhang2018perceptual}
Richard Zhang, Phillip Isola, Alexei~A. Efros, Eli Shechtman, and Oliver Wang.
\newblock The unreasonable effectiveness of deep features as a perceptual
  metric.
\newblock In {\em CVPR}, 2018.

\bibitem{zhang2016arx:unst}
Yi Zhang, Weichao Qiu, Qi Chen, Xaolin Hu, and Alan Yuille.
\newblock {UnrealStereo}: {A} synthetic dataset for analyzing stereo vision.
\newblock In {\em arXiv:1612.04647}, 2016.

\bibitem{zhou2018stereo}
Tinghui Zhou, Richard Tucker, John Flynn, Graham Fyffe, and Noah Snavely.
\newblock Stereo magnification: Learning view synthesis using multiplane
  images.
\newblock {\em ACM Trans. Graph. (Proc. SIGGRAPH)}, 37, 2018.

\end{thebibliography}
